\pdfoutput=1

\documentclass[10pt,twocolumn]{article}          % twocolumn

\usepackage{geometry}
\usepackage{authblk}

\usepackage[section]{placeins}

\usepackage[skip=2pt,font=footnotesize]{caption}

\usepackage{color}
\usepackage{times}
\usepackage{epsfig}
\usepackage{amsmath}
\usepackage{amssymb}
\usepackage{booktabs}            %para tablas
\usepackage{multirow}
\usepackage[pagebackref=true,breaklinks=true,letterpaper=true,colorlinks,bookmarks=false,breaklinks]{hyperref} % TEST Test for long list of citations
\usepackage{nicefrac}

\usepackage{footnote} % TO PUT A FOOTNOTE ON A TABLE

\usepackage{bm}

\def\bx{\mathbf{x}}

\def\bS{\mathbf{S}}
\def\bu{\mathbf{u}}

\def\bI{\mathbf{I}}

\def\3d{\textsc{3d}}
\def\2d{\textsc{2d}}
\def\bomega{{\bm \omega}}
\def\balpha{{\bm \alpha}}

\def\bS{{\bf S}}

\def\bI{{\bf I}}

\def\bu{{\bf u}}

\def\bx{{\bf x}}

\def\bw{{\bf w}}

\def\ft2{{\bf c}}
\def\sft3{{\bf d}}

\def\bomega{{\bm \omega}}
\def\balpha{{\bm \alpha}}

\def\im34width{6cm}

\DeclareMathOperator*{\argmin}{arg\,min}

\newcommand{\nada}[1]{}

\def\dmin{{\delta_\text{min}}}
\def\nspo{{n_\text{spo}}}
\def\smin{{\sigma_\text{min}}}

\newcommand{\DeltaN}{nL}

\title{An analysis of the factors affecting keypoint stability in scale-space}

\author{Ives Rey-Otero$^{\dagger}$ \qquad Jean-Michel Morel$^{\dagger}$  \qquad Mauricio Delbracio $^{\star}$}
\affil{$^{\dagger}$CMLA, ENS-Cachan, France  \qquad $^{\star}$ECE, Duke University, USA}

\begin{document}
%\linenumbers
\date{}
\maketitle

%\mdC{In the end, do you think we should add the reference that Weickert pointed out in the report?}

\begin{abstract}
The most popular image matching algorithm SIFT,  introduced by D. Lowe a decade
ago, has proven to be sufficiently scale invariant to be used in numerous
applications.
In practice, however, scale invariance may be weakened by various sources of
error inherent to the SIFT implementation affecting the stability and accuracy
of keypoint detection.
The density of the sampling of the Gaussian scale-space and the level of blur
in the input image are two of these sources.  This article presents a numerical
analysis of their impact on the extracted keypoints stability.
Such an analysis has both methodological and practical implications, on how to
compare feature detectors and on how to improve SIFT.
We show that even with a significantly oversampled scale-space numerical
errors prevent from achieving perfect stability.
Usual strategies to filter out unstable detections are shown to be inefficient.
We also prove that the effect of the error in the assumption on the initial
blur is asymmetric and that the method is strongly degraded in presence of
aliasing or without a correct assumption on the camera blur.
\end{abstract}

%\begin{keywords}
%SIFT, invariance, scale-space, sampling, aliasing
%\end{keywords}

%\tableofcontents

\section{Introduction}

SIFT \cite{Lowe1999,Lowe2004} is a popular image matching method extensively
used in image processing and computer vision applications.  SIFT relies on the
extraction of keypoints and the computation of local invariant feature
descriptors.
The scale invariance property is crucial. The matching of SIFT features
is used in various applications such as image stitching
\cite{brown2007automatic}, \3d reconstruction \cite{riggi2006fundamental} and
camera calibration~\cite{strecha2008benchmarking}.

%%  POSITION DU PROBLEME
SIFT was proved to be theoretically scale invariant~\cite{morel2011sift}.
Indeed, SIFT keypoints are covariant, being the extrema of the image Gaussian
scale-space \cite{weickert1999linear,lindeberg1993scale}.
In practice, however, the computation of the SIFT keypoints is affected in many
ways, which in turn limits the scale invariance.  

The literature on SIFT focuses on variants, alternatives and accelerations
\cite{brown2007automatic,tuytelaars2008local,bay2006surf,Miko2005,forstner2009detecting,sifer,ancuti2007sift,pele2008linear,rabin2009statistical,ke2004pca,calonder2010brief,rublee2011orb,tola2008fast,tola2010daisy,vedaldi2010vlfeat,leutenegger2011brisk,agrawal2008censure,winder2007learning,winder2009picking,chen2010wld,grabner2006fast,liu2008sift,moreno2009improving,brown2005multi,dickscheid2011coding,sadek2012affine}.
A majority of them use the scale-space keypoints as defined in the SIFT method.
The huge amount of citations of SIFT indicates that it has
become a standard and a reference in many applications.
In contrast, there are
almost no articles discussing the scale-space settings in the SIFT method and trying to compare SIFT with
itself.  By this comparison we mean the question of comparing the scale invariance  claim in SIFT with its empirical invariance, and the influence of the SIFT
scale-space and keypoint detection parameters on its own performance.
On this strict subject D. Lowe's paper
\cite{Lowe2004} remains the principal reference, and it seems that very few of
its claims on the parameter choices of the method have undergone a serious
scrutiny.
This paper intends to fill in the gap for the main claim of the SIFT
method, namely the scale invariance of its keypoint detector, and incidentally on its translation
invariance. This is investigated by means of a
strict image simulation framework allowing us to control the main image and
scale-space sampling parameters: initial blur, scale and space sampling rates and noise
level.
We show that even in a particularly favorable scenario, many of the detected SIFT keypoints
are instable.
We prove that the scale-space sampling has an influence on the scale invariance and
that finely sampling the Gaussian scale-space improves the detection of scale-space extrema.
We quantify how the empirical invariance is affected by image aliasing and other errors due to wrong assumptions on the input image blur level.

Also, we verify the importance of the quadratic interpolation proposed in SIFT for refining the precision of the localized extrema. This is a fundamental step for the overall algorithm stability by filtering out unstable discrete extrema. On the other hand, we  show that the contrast threshold proposed in SIFT is ineffective to remove the unstable detections.

%<F8>A preliminary version of this work was published in~\cite{rey2014analysis}.
%
Some of the conclusions of this paper were announced in~\cite{rey2014analysis}.
The present article incorporates a more thorough rigorous analysis of the
scale-space extrema and their stability.
We reach this by separating the mathematical definition of the scale-space from
the numerical implementation.
We also add an analysis of the difference of Gaussians (DoG) scale-space
operator and a discussion on how fine the scale-space should be sampled to
fulfill the SIFT invariance claim.

The remainder of the paper is organized as follows.
Section~\ref{sec:sift} presents the SIFT algorithm and details how to implement
the Gaussian scale-space for the requirements of the present work.
%the
%Gaussian scale-space is implemented.
%
Section~\ref{sec:theoretical} exposes the SIFT theoretical scale invariance. With
that aim in view, we explicit the camera model consistent with the SIFT method.
Section~\ref{sec:simulating:digital:camera} details how input images are
simulated to be rigorously consistent with SIFT camera model. 
Section~\ref{sec:analysis:discrete:scale:space} explores the extraction of SIFT
keypoints at each stage of the algorithm focusing on the impact of
the scale-space sampling on detections.
%
%%Section~\ref{sec:analysis:discrete:scale:space} provides an empirical analysis
%%of the scale-space sampling.
%
%
Section~\ref{sec:deviation} looks at the impact of image aliasing and of errors
in the estimation of camera blur.
Section~\ref{sec:conclusion} is the conclusion.

\section{The SIFT method and its exact implementation}\label{sec:sift}

In this section we briefly review the SIFT method and fix the adjustments that
are required to make it ideally precise. This ideal SIFT  will be used in the
next sections to explore the limits of the SIFT method to detect scale-space
extrema.

\subsection{SIFT overview}
\label{sec:siftoverview}

SIFT derives from scale invariance properties of the Gaussian scale-space
\cite{weickert1999linear,lindeberg1993scale}.  The Gaussian scale-space of an
initial image $u$ is the \3d function
$$
v:(\sigma,\bx) \mapsto G_\sigma u(\bx),
$$
where $G_\sigma u(\bx)$ denotes the convolution of $u(\bx)$ with a Gaussian
kernel of standard deviation $\sigma>0$ (the scale).
In this framework, the Gaussian kernel acts as an approximation of the optical
blur introduced in the camera (represented by its point spread function).
Among other important properties~\cite{lindeberg1993scale}, the Gaussian
approximation is convenient because it satisfies the semi-group property
\begin{align}
G_{\sigma}  G_{\gamma} u (\bx)  = G_{\sqrt{ \sigma ^2 + \gamma ^2} }u(\bx).
\label{eq:semigroup}
\end{align}
In particular, this permits to simulate distant snapshots from closer ones.
Thus, the scale-space can be seen as a stack of images, each one corresponding
to a different zoom factor.  Matching two images with  SIFT consists in
matching keypoints extracted from these two stacks.

SIFT keypoints are defined as the \3d extrema of the difference of Gaussians
(DoG) scale-space.  Let $v$ be the Gaussian scale-space and $\kappa>1$, the DoG is the \3d
function
$$
w: (\sigma,\bx) \mapsto v( \kappa \sigma,\bx)-v(\sigma,\bx).
$$
When $\kappa \to 1$, the DoG operator acts as an approximation of  the normalized Laplacian of
the scale-space~\cite{Lowe2004,lindeberg1993scale},
$$
 v( \kappa \sigma,\bx)-v(\sigma,\bx) \approx (\kappa-1)\sigma^2 \Delta v(\sigma, \bx).
$$

Continuous \textsc{3d} extrema of the digital DoG are calculated in two successive 
steps. First, the DoG scale-space is scanned for localizing discrete extrema. This is done by comparing 
each voxel to its 26 neighbors. 
Since the location of the discrete extrema is constrained to the scale-space sampling grid,
SIFT refines the position and scale of each candidate keypoint using a local interpolation model. 
Given a detected discrete extremum  $(\sigma, \bx)$ of the digital DoG space, we denote by $\bomega_{\sigma,\bx}(\balpha)$  the quadratic function at sample point $(\sigma,\bx)$ given by
\begin{equation}
 \bomega_{\sigma,\bx}(\balpha) = \bw_{\sigma,\bx} +  \balpha^T   g_{\sigma,\bx} + \frac{1}{2} \balpha^T   H_{\sigma,\bx}  \balpha,
 \label{eq:3d:quadratic:model}
\end{equation}
where  $\balpha = (\alpha_1,\alpha_2,\alpha_3) \in
[-\nicefrac12,\nicefrac12]^3$; $ g_{\sigma,\bx}$ and $ H_{\sigma,\bx}$ denote
the \textsc{3d} gradient and Hessian at $(\sigma,\bx)$ computed with a finite
difference scheme. This quadratic function can be interpreted as an
approximation of the second order Taylor expansion of the underlying
continuous function (where its derivatives are approximated by finite
differences).

To refine the position of a discrete extremum $(\sigma_0,\bx_0)$  SIFT proceeds as follows.
\begin{enumerate}
\item Initialize $(\sigma,\bx) = (\sigma_0,\bx_0)$.

\item Find the extrema of $\bomega_{\sigma,\bx}$ by solving $\nabla \bomega_{\sigma,\bx} (\balpha) =  0$. This yields
$\balpha^\ast  = - \left( H_{\sigma,\bx} \right)^{-1} \,   g_{\sigma,\bx}$ and a refined DoG value $\bomega_{\sigma,\bx}({\balpha}^\ast)$.
The corresponding keypoint coordinates are updated accordingly.

\item If $\|\balpha^\ast \|_\infty  <  M_\text{offset} = 0.6$ the extremum is accepted. Otherwise, go back to Step~1 and recompute the quadratic model at the closest point in the scale-space discrete grid.

\end{enumerate}
This process is repeated up to $N_\text{interp}$ times  (in SIFT, $N_\text{interp}=5$) or until the interpolation is validated. If after five iterations the result is not yet validated, the candidate keypoint is discarded.  

Low contrast detections are filtered out by discarding keypoints with a
small DoG value. Keypoints lying on edges are also discarded since their
location is not precise due to their intrinsic translation invariant nature.

A reference keypoint orientation is computed based on the dominant gradient
orientation in the keypoint surrounding.  This orientation along with the
keypoint coordinates are used to extract a covariant patch.  Finally, the
gradient orientation distribution in this patch is encoded into a $128$ elements
feature, the so-called SIFT descriptor. We shall not discuss further the
constitution of the descriptor and refer to the abundant
literature~\cite{mikolajczyk2005performance,moreno2009improving,van2010evaluating,ke2004pca,sadek2012affine,calonder2010brief}.
For a detailed description of the SIFT method we refer the reader
to~\cite{reyotero2014anatomy}.

\subsection{The Gaussian scale-space and its implementation}

Let us assume that the input image has Gaussian blur level $c$.
The construction of the digital scale-space begins with the computation of a
\emph{seed} image.  For that purpose, the input image is oversampled by a factor
$\nicefrac{1}{\dmin}$ and filtered by a Gaussian kernel $G_{\sqrt{\smin^2 -
c^2}}$  to reach the minimal level of blur $\smin$ and inter-pixel
distance $\dmin$.
The scale-space set is split into subsets where images share a common
inter-pixel distance. Since in the original SIFT algorithm the sampling rate is
iteratively decreased by a factor of two, these subsets are called
\emph{octaves}. We shall denote by $n_\text{spo}$ the number of scales per octave.

The subsequent images are computed iteratively from the \emph{seed} image using
the semi-group property~\eqref{eq:semigroup} to simulate the blurs following a
geometric progression 
$$\sigma_s = \smin 2^{s/\nspo},\quad s=1,\ldots,\nspo\!-\!1.$$ 
The digital Gaussian scale-space architecture is unequivocally defined by four
parameters: the number of octaves $n_\text{oct}$, the minimal blur level
$\sigma_\text{min}$ in the scale-space, the number of scales per octave
$n_\text{spo}$ and the initial oversampling factor $\delta_\text{min}$.
The standard values proposed in SIFT~\cite{Lowe1999} are $\nspo=3$, $\dmin=\nicefrac{1}{2}$ and $\sigma_\text{min}=0.8$.
By increasing $\nspo$ the scale dimension can be sampled arbitrarily finely. In
the same way by considering a small $\dmin$ value, the 2\textsc{d} spatial position can
be sampled finely.

From this digital Gaussian scale-space the difference of Gaussian scale-space (DoG) is computed. 
A DoG image at scale $\sigma$ is computed by
subtracting from the image with blur level $\kappa\sigma$ the image with blur level $\sigma$
(with $\kappa>1$).
Originally, the DoG scale-space is computed as a simple difference between two
successive scales of the Gaussian scale-space so that $\kappa = 2^{1/\nspo}$.
In the present work, we have modified this definition by unlinking the parameters $\kappa$ and $\nspo$. This will allow us to
better analyze the implications of the mathematical definition of the DoG operator (given by the $\kappa$-value) and the
algorithmic implementation (given by the sampling parameter $\nspo$).

\vspace{1em}

\noindent \textbf{The Gaussian convolution implementation.}
%\label{sub:section:implementation:gaussian:convolution}
The architecture of the Gaussian scale-space requires for the Gaussian
convolution to be implemented so it satisfies the semi-group
property~\eqref{eq:semigroup}.
In SIFT, the Gaussian convolution is implemented as a discrete convolution with
a sampled truncated Gaussian kernel.
Such an implementation satisfies the semi-group property for the SIFT default
parameters ($\nspo=3$), but it fails for larger values of $\nspo$, as the
level of blur to be added approaches zero. 

\begin{figure}[t]
\begin{center}
\begin{minipage}[b]{.45\linewidth}
  \centering
  \centerline{\includegraphics[width=\textwidth]{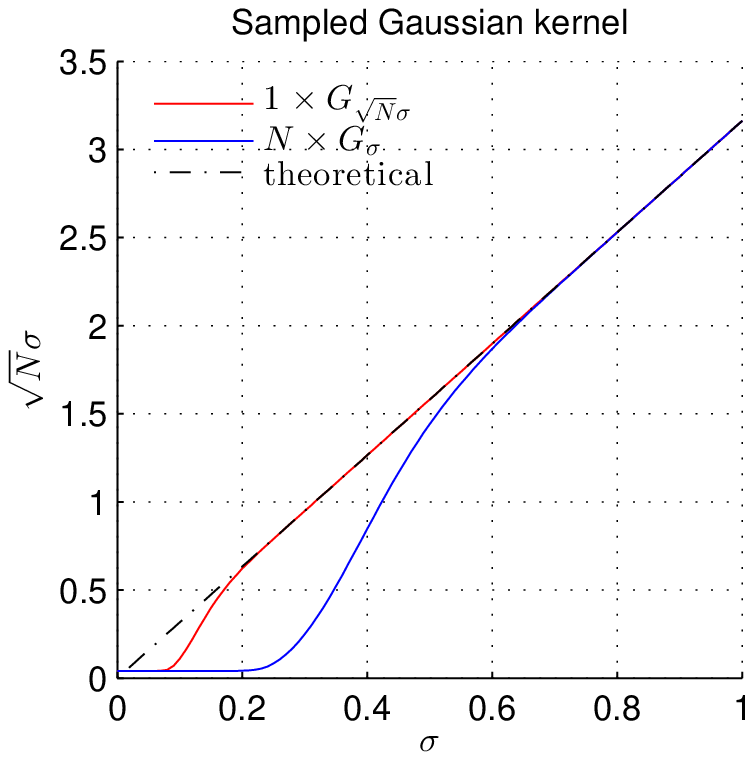}}

  {\bf (a)}
\end{minipage}
%\hfill
\hspace{1em}
\begin{minipage}[b]{.45\linewidth}
  \centering
  \centerline{\includegraphics[width=\textwidth]{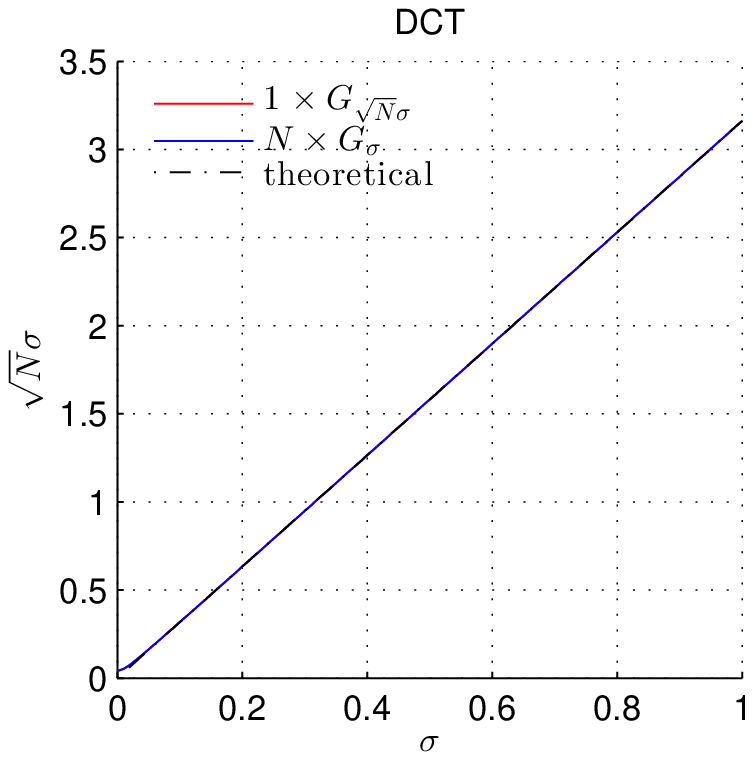}}

  {\bf (b)}
\end{minipage}
\end{center}
\caption{
%  \ivC{To perform a thorough analysis of the influence of scale-space sampling, we need
%  The DCT interpolation.}
  %
  Analysis of the Gaussian convolution implementation through the semi-group property.
An image having a Gaussian blob of standard deviation $c=1.1$ was filtered by (i) a Gaussian convolution of parameter $\sqrt{N}\sigma$, and (ii)  by applying  $N=10$  iterations of a Gaussian convolution of parameter $\sigma$ for different
  values of $\sigma$. Then the blur levels of the filtered images were estimated and compared to the theoretical expected value.
  {\bf (a)}~Discrete convolution with sampled Gaussian kernel. For low
  values of $\sigma$, the estimated blur deviates from the theoretical value
  $\sqrt{N}\sigma$. This is due to image aliasing when sampling the Gaussian kernel.
  {\bf (b)}~The DCT convolution fully satisfies the semi-group property.
}
\label{fig:semigroup:methods}
\end{figure}

To illustrate and quantify how the discrete Gaussian convolution fails to
satisfy the semi-group property, we carried out the following experiment.
A sampled Gaussian function of standard deviation $c=1.0$ was
filtered $N=10$ times using a discrete Gaussian filter of standard deviation $\sigma$.
If the Gaussian semigroup property were valid, then, applying $N$ times a
Gaussian filter of parameter $\sigma$ should produce the same result as
filtering only once with a Gaussian function of parameters $\sqrt{N}\sigma$.
We fitted a Gaussian function to the filtered image by least squares.
We compared the estimated standard deviation  to the theoretical expected
value $\sqrt{\sigma^2_\text{in} + N\sigma^2}$
(Figure~\ref{fig:semigroup:methods}~{\bf (a)}).
For low values of $\sigma$ (i.e., $\sigma<0.7$), the estimated blur deviates from the theoretical
value $\sqrt{N}\sigma$ indicating that the method fails to satisfy the semi-group property.
%
%This is a direct consequence image aliasing when sampling
%a Gaussian kernel with low standard deviation~\cite{reyoteroComputUnpub}.
%
%This is a direct consequence of image aliasing produced by excessive undersampling~\cite{reyoteroComputUnpub}.
%
This is a direct consequence of image aliasing produced by excessive undersampling of the Gaussian kernel~\cite{reyoteroComputUnpub}.

To avoid this undesired phenomenon in our experiments that will consider strong
scale oversampling,  we replaced the discrete convolution by a Fourier-domain
based convolution using the Discrete Cosine Transform (DCT).  This can be
interpreted as the continuous convolution between the DCT interpolation the
discrete input image and the Gaussian kernel.
The implementation details can be found in~\cite{reyoteroComputUnpub}.

Figure~\ref{fig:semigroup:methods}{\bf (b)} shows that the Fourier-based convolution
satisfies the semi-group property even for low values of $\sigma$.
%This is why we opted to use the DCT Gaussian filtering.

%\subsection{Other modifications of the SIFT implementation}
\subsection{Building an ideal SIFT for parameter exploration}
\label{sub:section:modification:implementation}

% DEF : TO HEM : turn under and sew the edge of (a piece of cloth or clothing).

% % Since our goal was  to explore extrema detection, we implemented an ideal SIFT
% % where not only the convolution is exact, but also the extrema filters were
% % hemmed.
% % %%%\ivC{hemmed ?}
%
Since our goal was  to explore extrema detection, we implemented an ideal SIFT
where not only the convolution is exact, but also the extrema filters were
turned off.
The implementation of SIFT used in the present work differs from the original one
on two aspects (besides the replacement of the discrete convolution by the Fourier-based one).
%
%The threshold and the computation of the difference of Gaussian (DoG) scale-space.
%
First, SIFT proposes two filters to discard unreliable keypoints.
The first one eliminates poorly contrasted extrema (those with low DoG value)
and the second one discards extrema laying on edges (using a threshold on the
local Hessian spectrum).
These filters were deactivated to gain a full control of all detected extrema
and to isolate the impact of each of them in terms of keypoints stability.
This choice will be {\it a posteriori} justified, as we demonstrate in
Section~\ref{sub:section:can:unstable:be:detected} that the DoG contrast
threshold is inefficient.

Secondly, we decided to implement the DoG operator in such a way that the same
mathematical definition is kept (i.e., using the same $\kappa$-value) regardless of the scale
sampling rate $\nspo$.
SIFT approximates the normalized Laplacian $\sigma^2 \Delta $ by the difference
of Gaussian operator.
Different DoG definitions lead to different extrema.
Consider for instance an image with a Gaussian blob of standard deviation
$\sigma_\text{blob}$ as input. The normalized Laplacian will have an extremum at
the center of the Gaussian blob, and scale $\sigma_\text{detect} =
\sigma_\text{blob}$.
On the other hand, the DoG scale-space of parameter $\kappa$ yields an extremum
at scale
$ \sigma_\text{detect} = \nicefrac{ \sigma_\text{blob}}{\sqrt{\kappa}}$.
Consequently, the range of scales simulated in the scale-space is affected by
the parameter $\kappa$. 

For the requirements of the present work, and to investigate thoroughly how the
operator definition affects extrema extraction, the considered DoG scale-space
implementation allows us to set $\kappa$ and $\nspo$ independently.

\vspace{.5em}

\noindent \textbf{Implementation details.}
The input image was oversampled by a factor $1/\dmin$ to reach the $\dmin$ sampling rate. 
This was done by using a cubic B-spline interpolation of order 3.
From this interpolated image all images in the scale-space were computed using a
combination of DCT Gaussian convolution and subsampling.
For each scale $\sigma$ simulated in an octave, the algorithm computes two
images, the first one corresponding to scale $\sigma$ and the second one corresponding to scale
$\kappa\sigma$ (both being directly computed from the input image).
Although we lost the benefit of a low computational cost, this gave us
flexibility and allowed us to investigate the influence of the operator definition
regardless of the scale-space sampling rate.

\section{The theoretical scale invariance}\label{sec:theoretical}

In this section we give the correct proof that SIFT is scale invariant and
stress the fact that this proof also indicates that knowing exactly the initial
camera blur is crucial for the method's consistency.

\subsection{The camera model}

In the SIFT framework, the camera point spread function is modeled by a
Gaussian kernel $G_c$ and all digital images are frontal snapshots of an ideal
planar object described by the infinite resolution image $u_0$.
In the underlying SIFT invariance model, the camera is allowed to rotate around
its optical axis, to take some distance, or to translate while keeping the same
optical axis direction. All digital images can therefore be expressed as
\begin{align}
\label{imageformationmodel}
\bu=:\bS_1 G_c  H \mathcal T R  u_0,
\end{align}
where
$\bS_1$ denotes the sampling operator, $H$ an arbitrary homothety,
$\mathbf{\mathcal T}$ an arbitrary translation and $R$ an arbitrary rotation.

\subsection{The SIFT method is theoretically invariant to zoom outs}

It is not difficult to prove that SIFT is consistent with the camera model.
Nevertheless, the proof in \cite{morel2011sift} is inexact, as pointed out in
\cite{sadek2012some}. Let $\bu_\lambda$ and $\bu_\mu$ denote two digital
snapshots of the scene $u_0$.
More precisely,
\begin{equation}
\bu_\lambda = \bS_1 G_c H_\lambda u_0
\text{ \,\,  and \, \,  }
\bu_\mu = \bS_1 G_c H_\mu u_0.
    \label{}
\end{equation}

Assuming that the images are well sampled, namely that $\bS_1$ is invertible by
Shannon interpolation, and taking advantage of the semi-group property
\eqref{eq:semigroup}, the respective scale-spaces are
\begin{eqnarray}
    v_\lambda(\sigma,\bx) &=& G_{\sqrt{\sigma^2 - c^2}} \bI_1 \bS_1 G_c H_\lambda u_0 (\bx) = G_\sigma H_\lambda u_0(\bx) \\
%                          &=& G_\sigma H_\lambda u_0(\bx), \\
    v_\mu(\sigma,\bx) &=& G_\sigma H_\mu u_0(\bx),
\end{eqnarray}
where $\bI_1$ denotes the Shannon interpolation operator.
These formulae imply that both scale-spaces only differ by a reparameterization.
Indeed, if $v_0$ denotes the Gaussian scale-space of the infinite resolution
image $u_0$ (i.e., $v_0(\sigma, \bx) = G_\sigma u_0(\sigma, \bx)$) we have
\begin{align}
    v_\lambda(\sigma, \bx) &= H_\lambda ( G_{\lambda \sigma} u_0(\bx) ) = v_0( \lambda \sigma, \lambda \bx), \\
    v_\mu(\sigma, \bx) &= v_0( \mu \sigma, \mu \bx),
\end{align}
thanks to a commutation relation between homothety and convolution.

By a similar argument, the two respective DoG functions are related to the DoG function $w_0$ derived from $u_0$.
For a ratio $\kappa > 1$ we have
\begin{align}
    w_\lambda(\sigma, \bx) &= v_\lambda(\kappa \sigma, \bx) - v_\lambda(\sigma, \bx) \\
                           &= v_0( \kappa \lambda \sigma, \lambda \bx) - v_0( \lambda \sigma, \lambda \bx) \\
                           &= w_0( \lambda \sigma, \lambda \bx) % \\
\end{align}
and similarly $w_\mu(\sigma, \bx) = w_0( \mu \sigma, \mu \bx)$.

Consider an extremum point $(\sigma_0,\bx_0)$ of the DoG scale-space $w_0$.
Then if $\sigma_0\geq \max(\lambda c, \mu c),$  this extremum corresponds to 
extrema $(\sigma_1,\bx_1)$ and $(\sigma_2,\bx_2)$ in $w_\lambda$ and $w_\mu$
respectively, satisfying
$
\sigma_0=\lambda\sigma_1=\mu\sigma_2.
$
This equivalence of extrema between the two scale-space guarantees that the
SIFT descriptors are identical.

Note that this same relation links the two normalized Laplacians applied on $v_\lambda$
and $v_\mu$, denoted respectively $\DeltaN_\lambda$ and $\DeltaN_\mu$, both related to 
the normalized Laplacian of $v_0$ denoted $\DeltaN_0$.
We have
\begin{align}
    \DeltaN_\lambda(\sigma, \bx) &= \sigma^2 \Delta v_\lambda(\sigma, \bx) \\
                          %  &= \sigma^2 \Delta \left( v_0( \lambda \sigma, \lambda \bx) \right) \\
                            &= (\lambda\sigma)^2 \Delta v_0( \lambda \sigma, \lambda \bx)  \\
                            &= \DeltaN_0(\lambda \sigma, \lambda \bx) \\
    \DeltaN_\mu(\sigma, \bx) &= \DeltaN_0( \mu \sigma, \mu \bx) % \\
\end{align}
Therefore, considering extrema of the normalized Laplacian as keypoints will also lead to 
SIFT descriptors that are identical.

\subsection{Knowing the camera blur is crucial for scale invariance}
\label{sub:section:unknown:blur}

The knowledge of the camera blur is crucial to ensure the theoretical
invariance to zoom-outs~\cite{sadek2012some}.
Indeed, DoG scale-spaces computed with a wrong camera blur have in general
unrelated extrema. 
Starting again from the two digital snapshots $\bu_\lambda$ and $\bu_\mu$, but
assuming a wrong blur $c'$ instead of the correct blur $c$, the respective
Gaussian scale-spaces are:
\begin{align}
    v_\lambda(\sigma,\bx) &= G_{\sqrt{\sigma^2 - c'^2}} \bI_1 \bS_1 G_c H_\lambda u_0 (\bx) \\
                          &= G_{\sqrt{\sigma^2 - c'^2 + c^2  }} H_\lambda u_0(\bx) \\
                          &= v_0( \lambda \sqrt{\sigma^2-c'^2+c^2}, \lambda \bx)
    \label{eq:invar:1}
\end{align}
and
\begin{align}
    v_\mu(\sigma,\bx)     &= v_0( \mu \sqrt{\sigma^2-c'^2+c^2}, \mu \bx).
    \label{eq:invar:2}
\end{align}
We see that, because of the wrong blur assumption, the scale-space function
$v_0$ is shrunken or dilated along scale.
%\ivC{shrunk - verifier}
% Which should you use, shrunk or shrunken? The adjective shrunken means
% smaller than before. The verb shrink means to become smaller. Shrunk, not
% shrunken, is the past participle form of shrink.
%
%
The corresponding DoG scale-spaces are:
\begin{align}
\begin{split}
w_\lambda(\sigma, \bx) &= v_0( \lambda \sqrt{ \kappa^2 \sigma^2 -c'^2 +c^2}, \lambda \bx)\\
                      & - v_0( \lambda \sqrt{ \sigma^2 -c'^2 +c^2}, \lambda \bx),
\end{split}
    \\
\begin{split}
w_\mu(\sigma, \bx) & = v_0( \mu \sqrt{ \kappa^2 \sigma^2 -c'^2 +c^2}, \mu \bx)\\
                   & - v_0( \mu \sqrt{ \sigma^2 -c'^2 +c^2}, \mu \bx).
\end{split}
\end{align}
None of these are linear reparameterizations of the DoG function $w_0$ anymore.
They yield therefore unrelated extrema.
Such bias is maximal with detections at finer scales and with large zoom
factors.

\section{Simulating the digital camera}
\label{sec:simulating:digital:camera}

\begin{figure}[t]
\begin{center}
\begin{minipage}[b]{0.32\linewidth}
  \centering
  \centerline{\includegraphics[width=\textwidth]{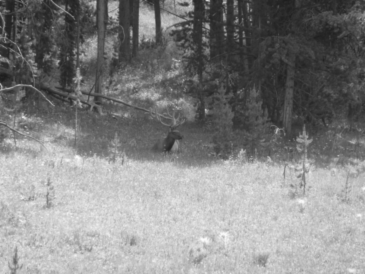}}
\end{minipage}
\hfill
\begin{minipage}[b]{0.32\linewidth}
  \centering
  \centerline{\includegraphics[width=\textwidth]{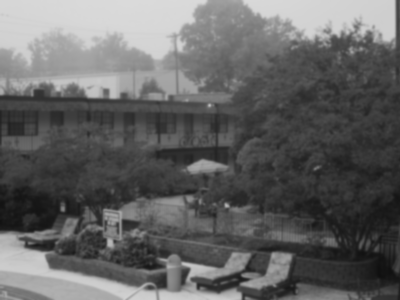}}
\end{minipage}
\hfill
\begin{minipage}[b]{0.32\linewidth}
  \centering
  \centerline{\includegraphics[width=\textwidth]{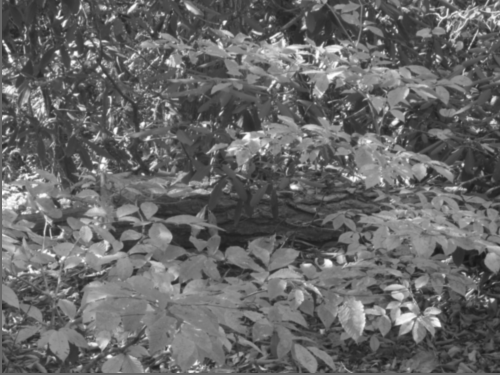}}
\end{minipage}

\end{center}
\caption{
 Examples of simulated images consistent with SIFT's image camera model. The
 respective blur levels are $c = 0.5$,  $c = 1.0$ and $c = 0.6$.}
\label{fig:synth:images}
\end{figure}

Controlling the image formation process permits us to measure how invariant SIFT is
in different scenarios.
Such a control was achieved by simulating images that are consistent with the SIFT
camera model.
Images at different zoom levels were simulated from a large reference real digital image
$u_\text{ref}$ through Gaussian convolution and subsampling.
To simulate a camera having a Gaussian blur level $c$, a Gaussian convolution of standard
deviation $c S$, with $S>10$ was first applied to the reference image. The convolved image was then
subsampled by a factor $S$.
Assuming that the reference image has an intrinsic Gaussian blur level of
$c_\text{ref} \ll c S$, the resulting Gaussian blur level is $\sqrt{c^2 +
(c_\text{ref}/S)^2 } \approx c$. 
We estimated the blur level introduced by a digital reflex camera by fitting a Gaussian function to the 
estimated camera point-spread-function (following~\cite{ipol.2012.admm-nppsf}).  The
obtained Gaussian blur levels varied from $c=0.35\text{--}0.95$, depending on
the aperture of the lens (blur level increases with aperture size).
Different zoomed-out and translated versions were simulated  by
adjusting the scale parameter $S$ and by translating the sampling grid.
Thanks to the large subsampling factor, the generated images are noiseless. In
addition, the images were stored with 32 bit precision to mitigate quantization
effects.  Figure~\ref{fig:synth:images} shows some examples of simulated images used in the
experiments.

It might be objected that our simulations are highly unrealistic as the images
to be compared by SIFT  in a real scenario are not perfectly sampled or noiseless.
Nevertheless, with an ever growing image resolution, more and more images will
be compared by SIFT in large octaves, and therefore after a large subsampling,
so that these properties can become realistic in practice.
Furthermore, even if applying SIFT to the originals and regardless of initial
noise and blur, the images at large scales also become anyway perfect so that
the accuracy and repeatability issues under such favorable conditions are
relevant.

\section{Empirical analysis of the digital scale-space sampling}\label{sec:analysis:discrete:scale:space}

The SIFT method aims at locating accurately the extrema of the DoG
scale-space.
Ideally, one would like to detect and locate all extrema from the
underlying continuous DoG scale-space.  However, in practice, we do not have
access to the continuous scale-space but to its discrete counterpart.
In theory, as $\delta_\text{min} \to 0$ and $\nspo \to \infty$ the
discrete scale-space better approximates the continuous scale-space therefore
allowing to extract reliably all continuous extrema.
This section investigates what happens when the sampling rates increase and how
sampling affects the successive steps of the rudimentary procedure for
detecting \3d scale-space discrete extrema, namely the extraction of discrete
extrema, their quadratic interpolation and their filtering based on their DoG
response.

To focus on the influence of the scale-space sampling, the study was carried out
in the most favorable conditions: noiseless and aliasing-free input images  ($c = 1.1$ and  $S=10$).
In all experiments we set $\kappa=2^{1/3}$ to separate the mathematical definition of the DoG analysis 
operator from the scale-space discretization.

\subsection{Number of detections}
\label{sub:section:detection:number}

To evaluate how the scale-space sampling rates affects the number of detections 
we generated different scale-space discretizations by varying the parameters $(\dmin, \nspo)$, 
and extracted the \3d discrete extrema for each one of them. 

\begin{figure}[h]
\begin{center}
\begin{minipage}[m]{.25\linewidth}
  \centering
  \centerline{\includegraphics[width=\textwidth]{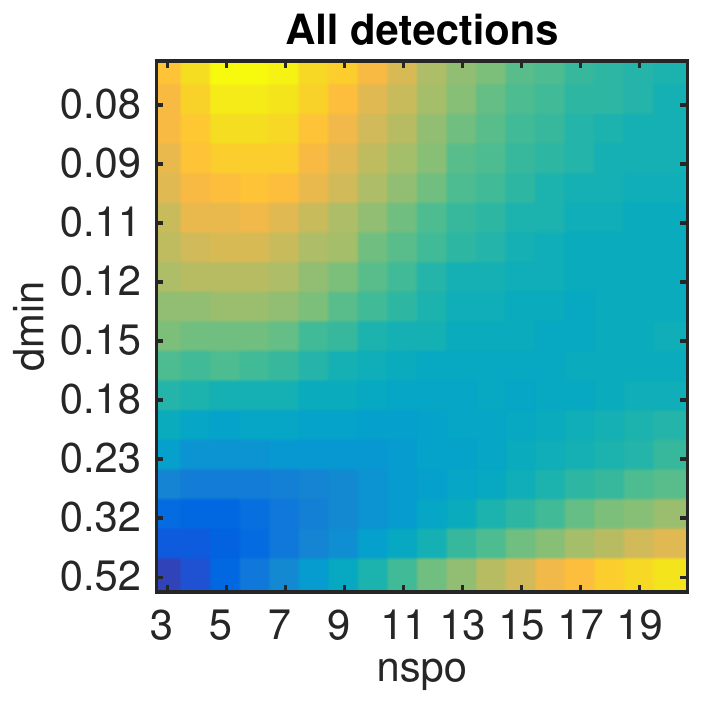}}
  {\bf (a)}
\end{minipage}
\begin{minipage}[m]{.25\linewidth}
  \centering
  \centerline{\includegraphics[width=\textwidth]{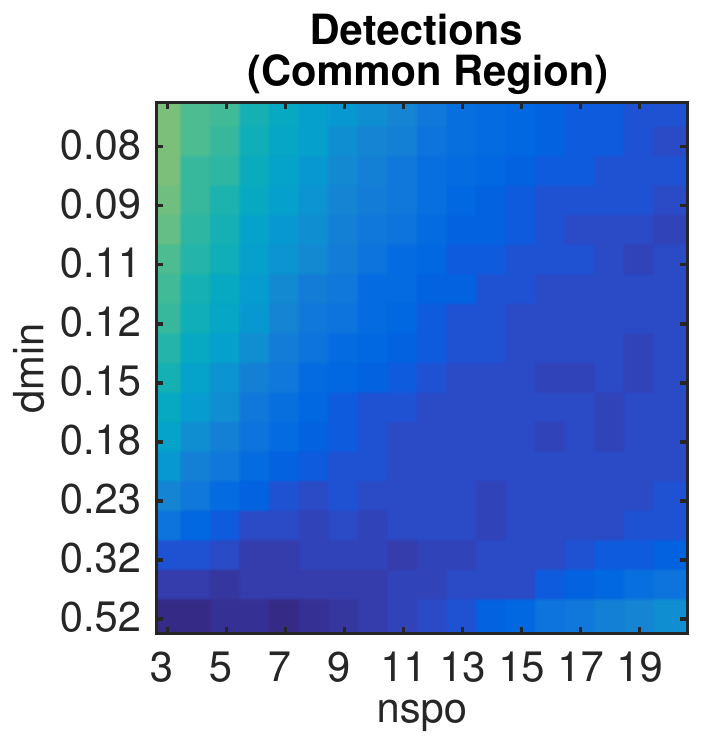}}
  {\bf (b)}
\end{minipage}
\begin{minipage}[m]{.25\linewidth}
  \centering
  \centerline{\includegraphics[width=\textwidth]{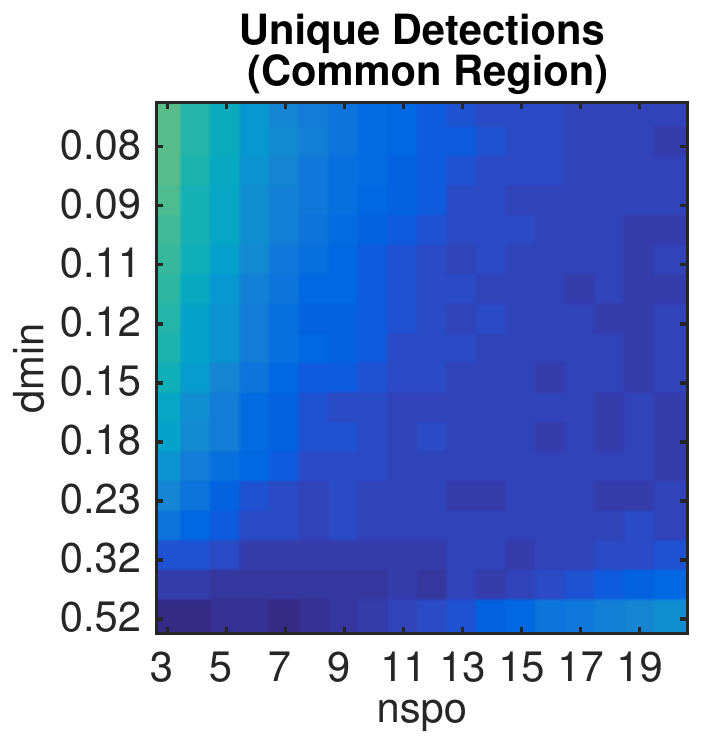}}
  {\bf (c)}
\end{minipage}
\begin{minipage}[m]{.07\linewidth}
  \centering
  \includegraphics[width=\textwidth]{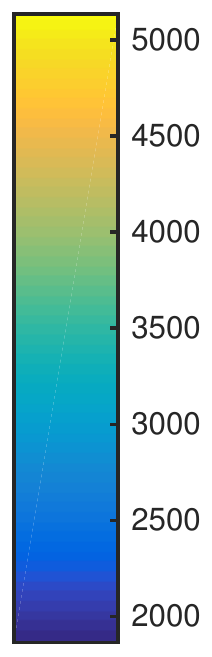}\vspace{2.3em}

\end{minipage}

\vspace{1em}

\begin{minipage}[b]{0.4\linewidth}
    \centering
    \includegraphics[width=.8\textwidth]{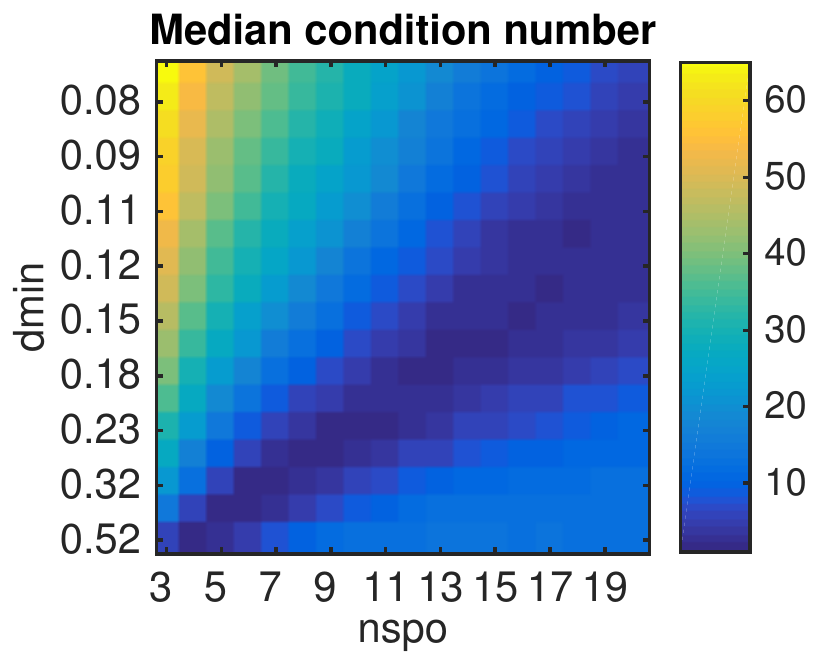}
    
      {\bf (d)}
      \vspace{0.4em}
      
\end{minipage}
\begin{minipage}[b]{0.40\linewidth}
    \centering
    \includegraphics[width=.85\textwidth]{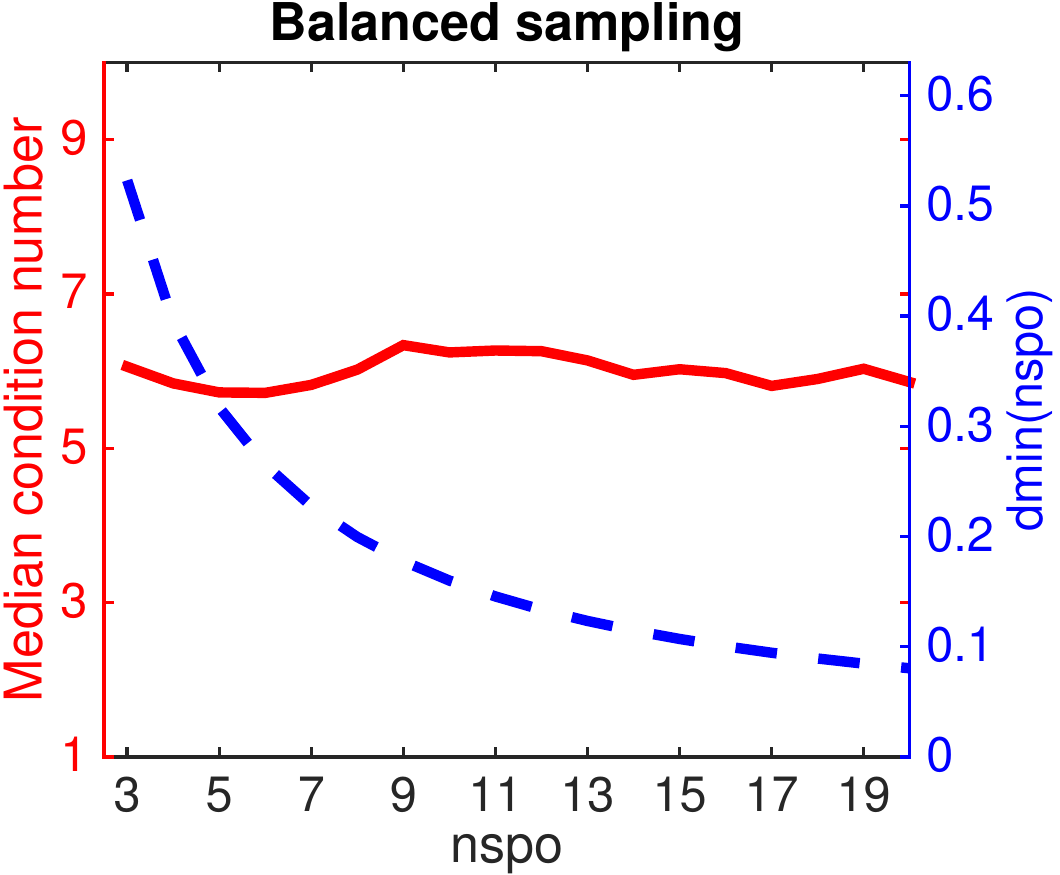}
    
      {\bf (e)}
     \vspace{.5em}
\end{minipage}

\end{center}
\caption{%
    Influence of the scale-space sampling rate $(\nspo,\dmin)$ on the number of detected DoG extrema.
    %
    %From the image \texttt{plants} (camera blur $c = 1.1$, $S=4$), we computed different discretizations of the  DoG scale-space.
    %
    %Different DoG scale-space samplings are computed.
    %
    {\bf (a)} Number of \3d DoG discrete extrema.
    Unbalanced discretizations can produce twice as many detections as the
    default scale-space sampling used in SIFT
    ($\nspo=3,\dmin=\nicefrac{1}{2}$). This gap is reduced after compensating
    for a boundary effect by discarding \3d discrete extrema with detected
    scale below $\smin 2^{1/3}$ {\bf (b)}, and after removing duplicate
    detections {\bf (c)}.
    Unbalanced discretizations may lead to inaccurate local models for the
    extrema refinement proposed in SIFT.
    {\bf(d)} Median of the condition numbers of DoG \3d Hessians used for
    extrema interpolations.
    Unbalanced sampling grids (shown in the top-right or bottom-left parts of
    this graph) produce extrema with significantly poor Hessian condition
    number. This leads to unstable extrema interpolations.    
    {\bf(e)} Balanced sampling rates (those satisfying~\eqref{eq:optSampling},
    shown in the dotted blue line) lead to extrema having well conditioned
    Hessian matrices (red line). 
}
\label{fig:discrete:detection:numbers}
\end{figure}

Figure~\ref{fig:discrete:detection:numbers}~{\bf (a)} shows the number of detected
extrema for the different scale-space samplings.
At first sight, it seems that some digital scale-space samplings produce many more
keypoints than the SIFT default sampling ($\dmin=\nicefrac12, \nspo=3$).
However, this increase in detections happens for discretizations that are significantly unbalanced in space and
in scale. By unbalance we mean that the scale and the space dimensions are sampled with very different
sampling rates.

\vspace{.5em}
\noindent \textbf{Boundary effect.} 
To do a fair comparison of the different  \emph{discrete} detected extrema when
changing the scale-space sampling rates, we have to consider that depending on
the scale-space sampling, some extrema close to the lower scale boundary
are not detected.
Indeed, due to the scale discretization there are no detected keypoints with
scale below \linebreak $\smin 2^{1/2\nspo}$.
To compensate for this dead range, which is a function of $\nspo$, we restricted
the analysis to a common scale range independent of $\nspo$.
This was achieved by discarding all extrema with scale below  $\smin 2^{1/3}$.
%
%\footnote{To avoid issues due to the coarse scale discretization, we used the
%keypoint scale value after refinement \eqref{eq:3d:quadratic:model}.
%The interpolation of the detected discrete extrema is only used for this
%filtering
%purpose, its location remains to be the one of the discrete extrema.}
%
To avoid issues due to the coarse scale discretization, we used the
keypoint scale obtained after refinement \eqref{eq:3d:quadratic:model}.
Figure~\ref{fig:discrete:detection:numbers}~{\bf (b)} shows, for all
scale-space tested configurations, the number of detections in the common scale
region.
The number of detected extrema lying in the common region is much more similar
for all the scale-space samplings.

\vspace{.7em}

\noindent \textbf{Duplicate detections.} 
%Given the set of detected extrema, there can be some detections that are actually the same.
We will say that detections $(\sigma_0,\bx_0)$ and $(\sigma_1,\bx_1)$ are the same, if:
\begin{align}
  ||\bx_0 - \bx_1||_\infty \le  \epsilon \quad \text{and} \quad   R^{-1}  \le  \sigma_1 / \sigma_0 \le R,
\label{eq:tolDupDet}
\end{align}
where $\epsilon$ and $R$ are the spatial tolerance and  scale relative
tolerance values respective.

Clearly, there is a compromise between saying that two detections are not the
same and allowing some displacement due to numerical errors.
Currently, we are not tackling the problem of precision
%(how accurate a keypoint can be localized)
but the problem of not mixing two different detections.
With that aim, it seems reasonable that the tolerance values are set  in order
to avoid that one detection be mistaken for another.
%
%Therefore, since the number of detections is almost constant, for different
%scale-space samplings
%
We opted to set tolerance values to $\epsilon=1.0$ and
$R=2^{\nicefrac{1}{2}}$ independently of the scale-space sampling.
 
Let $\mathcal{D}$ be the set of detected DoG extrema.
 We call  duplicates of $(\bx_0,\sigma_0) \in \mathcal{D}$ 
 the subset of detected extrema $\text{D}(\bx_0,\sigma_0) \subset \mathcal{D}$ that satisfy \eqref{eq:tolDupDet}.
Given the set of all detected keypoints $\mathcal{D}$, we say that $\mathcal{U}$ is a representative set of unique detections if
$$
\mathcal{U} = \argmin |U|  \quad \text{s.t.} \quad U \subset \mathcal{D} \,\, \text{and} \,\, \cup_{(\bx,\sigma) \in U } D(\bx,\sigma) = \mathcal{D},
$$
where the number of keypoints in the set $U$ is denoted by $|U|$.
Figure~\ref{fig:discrete:detection:numbers}~{\bf (c)} shows the number of unique detections in the
common scale region. The number of unique detections is similar to the number
of detections (Figure \ref{fig:discrete:detection:numbers}~{\bf (b)}).
This indicates that in general duplicate detections are negligible.

\vspace{.5em}

\noindent\textbf{Balancing the scale and space DoG sampling.}

The SIFT algorithm proposes to refine the position of a discrete extremum using a quadratic interpolation.
Having an unbalance sampling in scale and space may lead to an unreliable interpolation due to the very different discretization.
As we presented in Section~2, the refinement of a keypoint is done by solving a linear system (from~\eqref{eq:3d:quadratic:model}).
The sensitiveness to numerical errors can be measured by the linear system's condition number (i.e., the condition number of the Hessian at the extrema to be refined).
Figure~\ref{fig:discrete:detection:numbers}~{\bf (d)} shows the median of the condition
number for the sets of detected extrema associated with different scale-space samplings.
It shows that using a balanced sampling rate improves the overall stability of the extrema interpolation.

By balanced sampling we mean that the distance separating 
adjacent samples in the scale dimension is  similar to the distance
separating adjacent samples in space.
For a DoG scale-space with parameter $\kappa$, the distance between the first two 
simulated scales is
$$
\Delta \sigma = \kappa \smin (2^{1/\nspo} - 1).
$$
Thus, to equally sample the Gaussian kernel 
$$G(\bx,\sigma) = \frac{1}{2\pi\sigma^2} e^{-||\bx||^2/2\sigma^2}$$
 in scale and space, the spatial inter pixel distance should be
\begin{align}
\dmin = \sqrt{2} \Delta \sigma = \sqrt{2} \kappa \smin (2^{1/\nspo} - 1).
\label{eq:optSampling}
\end{align}

This relation between both sampling rates is plotted in
Figure~\ref{fig:discrete:detection:numbers}~{\bf (e)} along with the
median condition numbers on this set of balanced sampling rates.
The condition number is mostly constant for balanced samplings.

%\FloatBarrier

\subsection{Stability of DoG extrema to scale-space sampling }
\label{sub:sec:increasing:sampling:rates}

\begin{figure}[h!]
\begin{center}

\begin{minipage}[b]{.47\linewidth}
  \centering
  \centerline{\includegraphics[width=.85\textwidth]{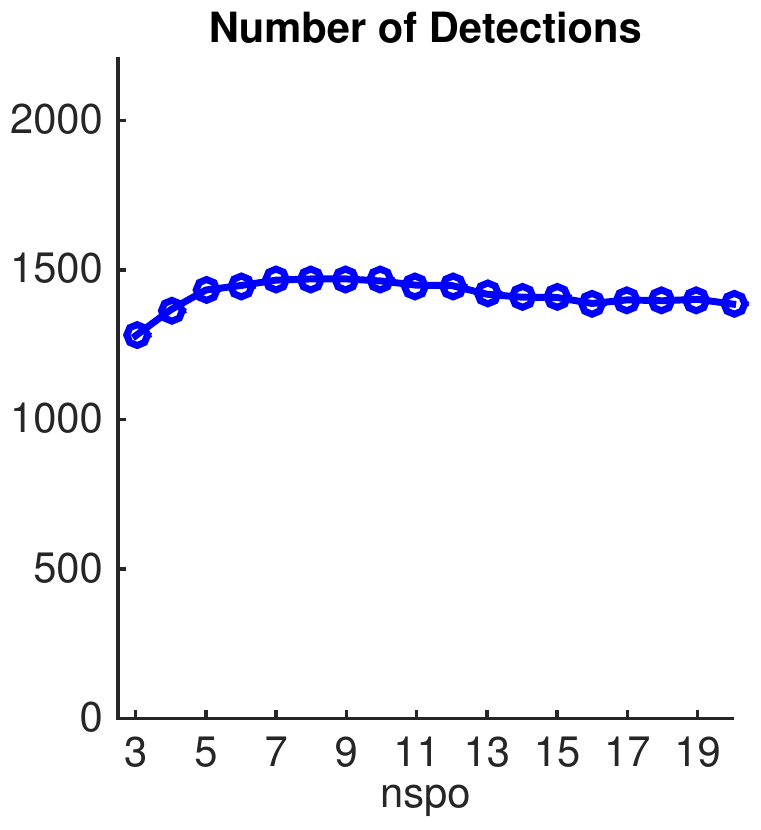}}
  {\bf (a)}
\end{minipage}
\begin{minipage}[b]{.47\linewidth}
  \centering
  \centerline{\includegraphics[width=.85\textwidth]{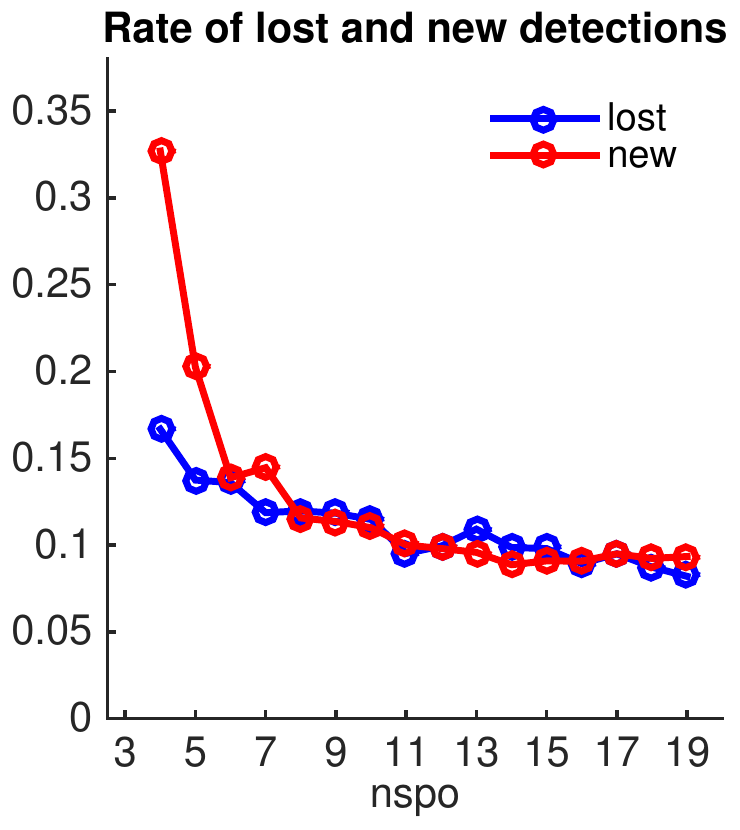}}
%%%  {\bf \ivC{(d)}}
  {\bf (b)}
\end{minipage}

        \begin{minipage}[b]{0.97\linewidth}
            \centering
            \includegraphics[width=\textwidth]{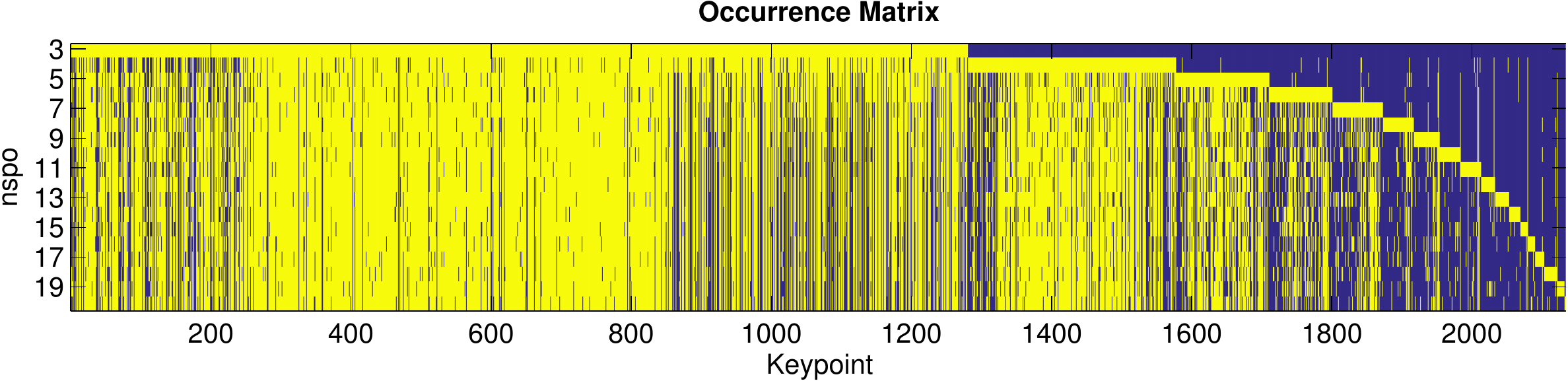}
            {\bf (c)}
        \end{minipage}
        
        \vspace{.5em}
        
        \begin{minipage}[b]{0.97\linewidth}
            \centering
            \includegraphics[width=\textwidth]{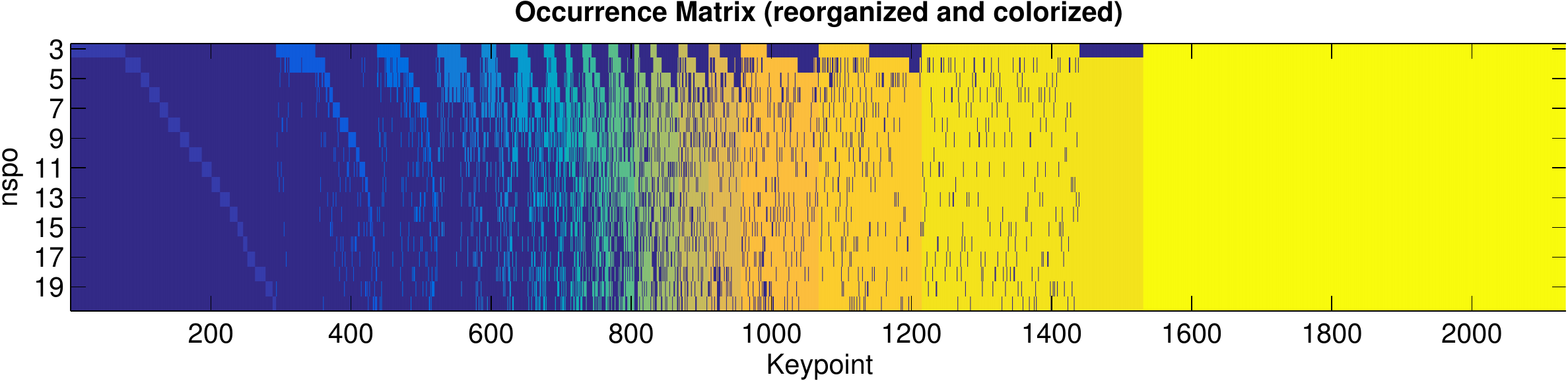}
            {\bf (d)}
        \end{minipage}

\end{center}
\caption{
    Influence of sampling density on keypoint stability.
    %
    %From the \texttt{plants} image (camera blur $c = 1.1$, $S=4$), we computed
    %a set of increasingly dense and balanced scale-spaces $(\nspo,\dmin)$. The
    %scale-space samplings are indexed by the $\nspo$ value, and the $\dmin$ is
    %given by \eqref{eq:optSampling}. 
    %
    A set of increasingly dense and balanced scale-spaces is computed.
    The scale-space samplings are indexed by the $\nspo$ value, and $\dmin$
    is given by \eqref{eq:optSampling}. 
    {\bf (a)} The number of detections is roughly constant for different sampling rates.
    {\bf (b)} The rates of lost extrema (detected in the current sampling but
    not in the immediately finer sampling) and of new extrema (detected in the
    current sampling but not in the immediately coarser sampling) decrease with
    the sampling rate $\nspo$ and stabilize around $10\%$
    of the total number of detections.
   % 
%    {\bf (b)} The rates of lost extrema (detected in the immediately coarser
    %    sampling but not in the current) and of new extrema (detected in the
    %    current sampling but not in the immediately coarser sampling) decrease
    %    with the sampling rate $\nspo$ and stabilize around $10\%$
%    of the total number of detections.
    %
    %
    {\bf (c)} The occurrence matrix. Each row in this matrix corresponds to
    one of the simulated samplings ($\nspo$), while each column indicates
    if a keypoint was detected in that particular sampling.
    {\bf (d)} For better visualization, the columns are colored and
    reorganized in increasing order of stability (yellow: always detected,
    blue: detected only once).
    %
    %Almost $20\%$ of the detections appear no matter the scale-space sampling rate.
    Almost $20\%$ of the detections appear for all scale-space sampling rates.
    %
    %  \ivC{ FIXED - xtick corrected avec interpolation - using tolerances
    %       $dx=1, Rds = 2^{1/2}$ with $L_\infty$ distance.}
    %
    }
\label{fig:discrete:detection:how:stable:NRR}
\end{figure}

To evaluate if all \3d discrete extrema are equally stable to an increase of the
DoG sampling rate, we simulated a set of increasingly dense balanced
scale-spaces.
We set the  minimal scale-space blur level to $\smin=1.1$.
We simulated increasingly dense scale-space samplings $(\nspo,\dmin)_i$, for
$i=1,\ldots,n$ with $\nspo = 3,\ldots,19$ and the balanced spatial
sampling rate $\dmin := \dmin(\nspo)$ given by~\eqref{eq:optSampling} ($i=1$ being the coarsest one and $i=n$ the finest one).
Figure~\ref{fig:discrete:detection:how:stable:NRR}~{\bf (a)} shows that the
number of detections is approximately constant for different balanced sampling rates.

Let $\mathcal{D}_i$ for $i=1\ldots,n$ be the sets of detected 3D extrema for the discretizations described above.
Given a detected extremum $(\bx_0,\sigma_0) \in \mathcal{D}_i$, we say the
extremum is detected in $\mathcal{D}_j$ if there exists  $(\bx,\sigma) \in \mathcal{D}_j$
such that they are the same detection according to the precision conditions~\eqref{eq:tolDupDet}.
We  say that a detected extremum $(\bx_0,\sigma_0) \in \mathcal{D}_i$ is new if
it was not detected in $\mathcal{D}_{i-1}$.
Given the sampling $i$, the rate of new extrema is computed as the proportion of
new detected keypoints among the total number of detections.
In the same way, we  define the rate of lost extrema as the proportion of those present in the
(coarser) sampling $i$ and not present in the (finer) sampling $i+1$. 
Figure~\ref{fig:discrete:detection:how:stable:NRR}~{\bf (b)} shows the rate of
new and lost detections as a function of the sampling rate. 
The new detection rate decreases with the sampling rate and stabilizes to a
minimal rate of $10\%$ of the total number of detections for $\nspo \geq 14$.
The same observations apply to the rate of lost extrema.

This surprising result means that despite sampling the scale-space very finely,
\3d discrete extrema keep appearing and disappearing when changing the
sampling.
%
%Various sources of errors such as the quantization errors prevent from reliably 
%extracting the \3d discrete extrema.
%

To illustrate how discrete extrema appear and disappear as scale-space sampling
rates changes, we decided to investigate the stability of each single detected
extremum.
The set of all unique detected extrema was formed by gathering the extrema
detected on all the simulated scale-spaces
$\mathcal{D}_\text{all}=\cup_{i=1,\ldots,n} \mathcal{D}_i$
and then by extracting a unique set of detections $\mathcal{U}_\text{all}$.
For each detected extremum $(\bx,\sigma) \in \mathcal{U}_\text{all}$, we
checked for its presence in each of the $\mathcal{D}_i$  detection sets.
This was done by using the same definition as in~\eqref{eq:tolDupDet}.
The results are summarized in the \emph{occurrence} matrix shown in
Figure~\ref{fig:discrete:detection:how:stable:NRR} {\bf(c)}. Each simulated
discretization is indexed by the $\nspo$ value.
Each entry in this matrix indicates if a keypoint in $\mathcal{U}_\text{all}$
(column) was found in the scale-space with a given discretization
$i=1,\ldots,n$ (where $i$ is the row index in the matrix).

%We define the \emph{stability} of a discretization as the ratio between the number of
%(unique) detections at that discretization divided by the number of unique
%detections in all the simulated discretizations.
%
% %We define the \emph{stability} of a unique keypoint as the ratio between the
% %number of discretizations it is detected in divided by the total number of
% %discretizations.
%
We define the \emph{stability} of a unique keypoint as the proportion of
discretizations where it is detected.
Figure~\ref{fig:discrete:detection:how:stable:NRR} {\bf (d)} shows the
normalized \emph{occurence} matrix, where each entry in the occurence matrix is
multiplied by the \emph{stability} value (therefore each column has the
same color).
Also, keypoints (columns) were reorganized from less to more stable (left
to right).

The normalized occurrence matrix confirms that a  majority of the keypoints are
stable as they appear on at least 80\% of the discretizations, and that some
keypoints tend to appear and disappear repeatedly as sampling rates increase.
It also shows that the proportion of unstable keypoints (e.g., those appearing
less than 20\% of the times) is low overall but is significantly larger for coarse
discretizations than in denser ones.

\subsection{Can unstable (intermittent) detections be detected?}
\label{sub:section:can:unstable:be:detected}
To increase its overall detection stability, SIFT discards non-contrasted
extrema based on their absolute DoG value.
However, many other features, computed from the values of the extremum and its
neighbors, could be used as well.
The DoG value, the Laplacian of the DoG, the DoG Hessian condition number and
the minimal absolute value of the difference  between the extremum and its
adjacent samples are some of them.

\begin{figure}[h!]
\begin{center}
    \begin{minipage}[b]{0.45\linewidth}
        \centering
        \includegraphics[width=.95\textwidth]{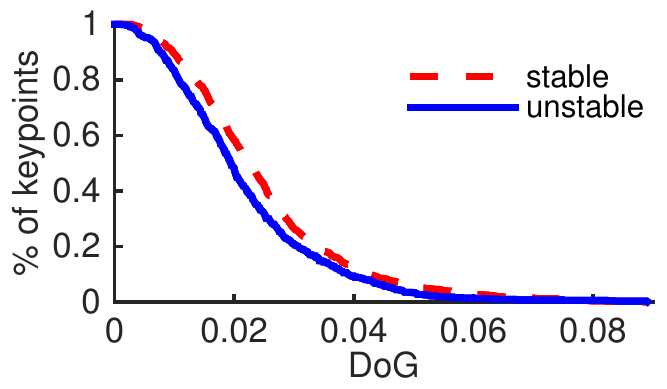}
        {\bf (a)}
    \end{minipage}
    \begin{minipage}[b]{0.45\linewidth}
        \centering
        \includegraphics[width=.95\textwidth]{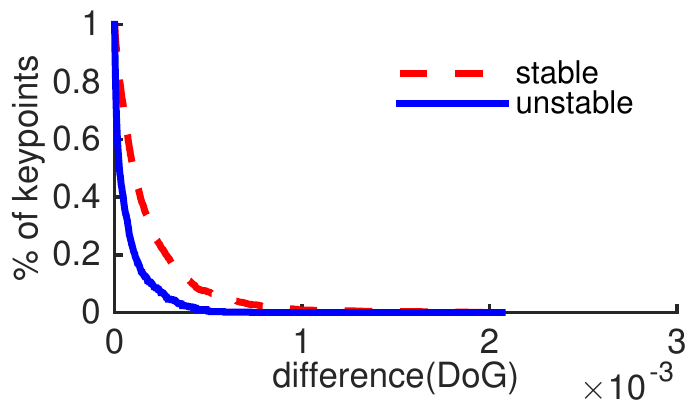}
        {\bf (b)}
    \end{minipage}

    \begin{minipage}[b]{0.45\linewidth}
        \centering
        \includegraphics[width=.95\textwidth]{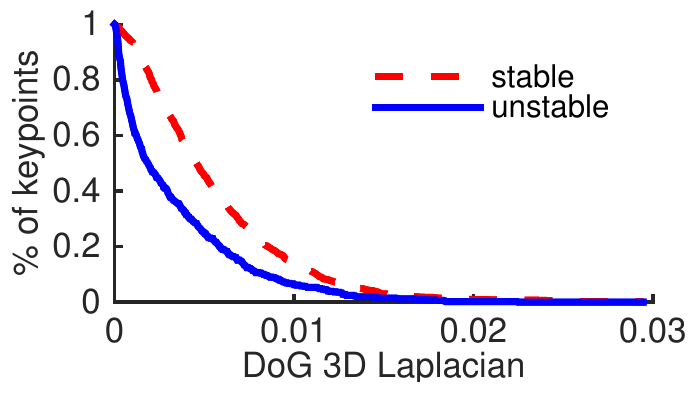}
        {\bf (c)}
    \end{minipage}
    \begin{minipage}[b]{0.45\linewidth}
        \centering
        \includegraphics[width=.95\textwidth]{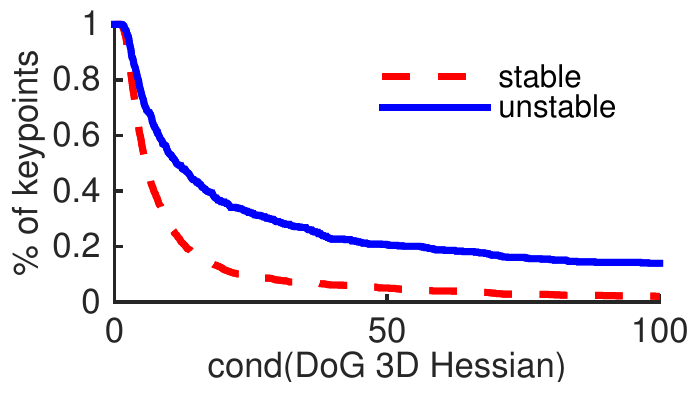}
        {\bf (d)}
    \end{minipage}
    
\vspace{.5em}
   
   \begin{minipage}[b]{0.6\linewidth}
        \centering
        \includegraphics[width=.95\textwidth]{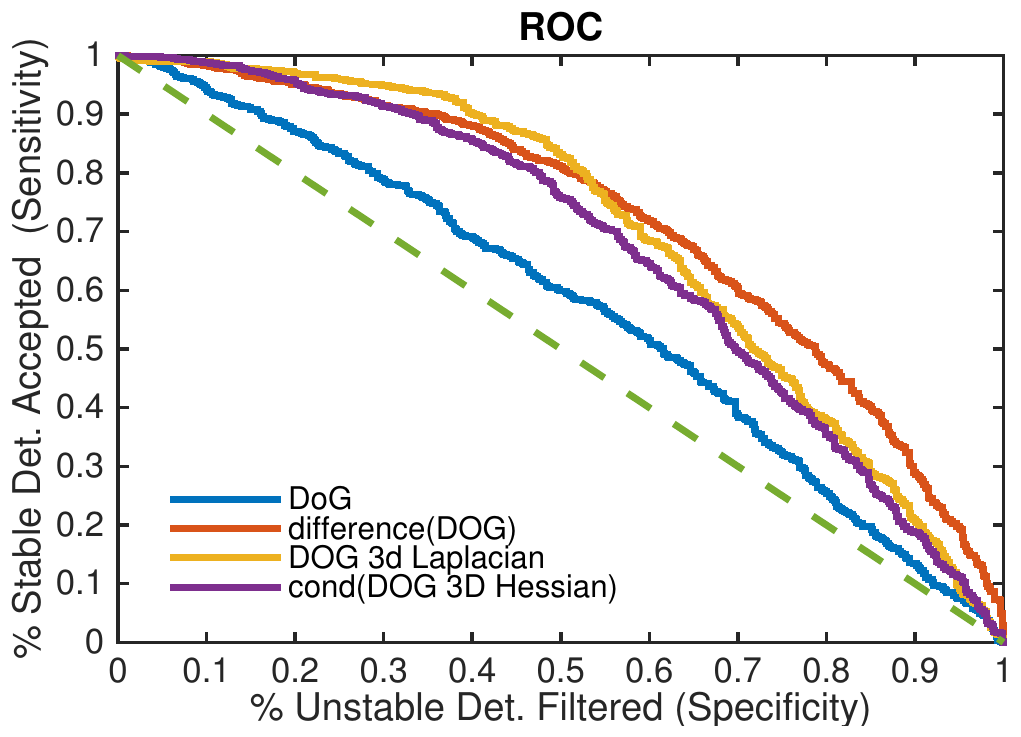}
        {\bf (e)}
    \end{minipage}

\end{center}
\caption{
    Attempts at filtering keypoints that are unstable to changes in the scale-space
    sampling. 
    Increasing thresholds are applied respectively to the set of stable and
    unstable detections. The considered features are:
    {\bf (a)} the extremum DoG value,
    {\bf (b)} the difference of extremum DoG value and the adjacent samples in
    the scale-space,
    {\bf (c)} the DoG \3d Laplacian value at the extremum,
    {\bf (d)} the condition number of the DoG \3d Hessian at the extremum.
    None of the tested features separates convincingly the unstable from the
    stable detections.
    This is confirmed by the ROC curves, illustrating the performance of each
    feature, shown in {\bf (e)}.
    A point in a ROC curve indicates the proportion of non-filtered stable
    keypoints (good detections -- sensitivity) as a function of the filtered
    unstable ones (good removals -- specificity) for a particular threshold
    value. A perfect feature should produce a ROC that is always equal to $1$.
    %
    %According to this, the worst feature for eliminating unstable keypoints
    %whatn changing the scales space sampling is the DoG value.
    %
    According to this experiment, the worst feature for eliminating keypoints unstable to
    changes in the scale-space sampling is the DoG value.
   }
\label{fig:discrete:simple:thresholds}
\end{figure}

To find out if any of these simple features is good at predicting if a discrete
extremum is stable (to different sampling rates), we proceeded as follows.
Given the set of unique detections $\mathcal{U}_\text{all}$ computed by
gathering all detections from the different scale-spaces with different
sampling rates,
%Subsection~\ref{sub:sec:increasing:sampling:rates}, 
we considered two subsets of unique keypoints: one subset of \emph{stable}
unique extrema (with occurrence rate above $80\%$) and one subset of
\emph{unstable} unique extrema (occurrence rate below $20\%$).
Figure~\ref{fig:discrete:simple:thresholds} {\bf (a--d)} shows the proportion
of extrema in both stable/unstable sets respectively, that have a feature value below
a certain threshold.
The considered features are: (a) the DoG value, (b) the Laplacian of the DoG,
(c) the DoG Hessian condition number and (d) the minimal absolute value of the
difference  between the extremum and its adjacent samples.

This figure demonstrates that none of these features manages to faithfully
separate the stable from the unstable ones.
This is confirmed by the ROC curve shown in
Figure~\ref{fig:discrete:simple:thresholds}~{\bf (e)} (see figure caption for
details).
Noticeably,  the keypoint feature giving the lowest discrimination performance
is the DoG value used by  SIFT.

%\FloatBarrier
\subsection{Visualizing unstable (intermittent) detections}

\begin{figure}[htp]
\begin{center}
    \begin{minipage}[b]{0.19\linewidth}
        \centering
        \includegraphics[width=1\textwidth]{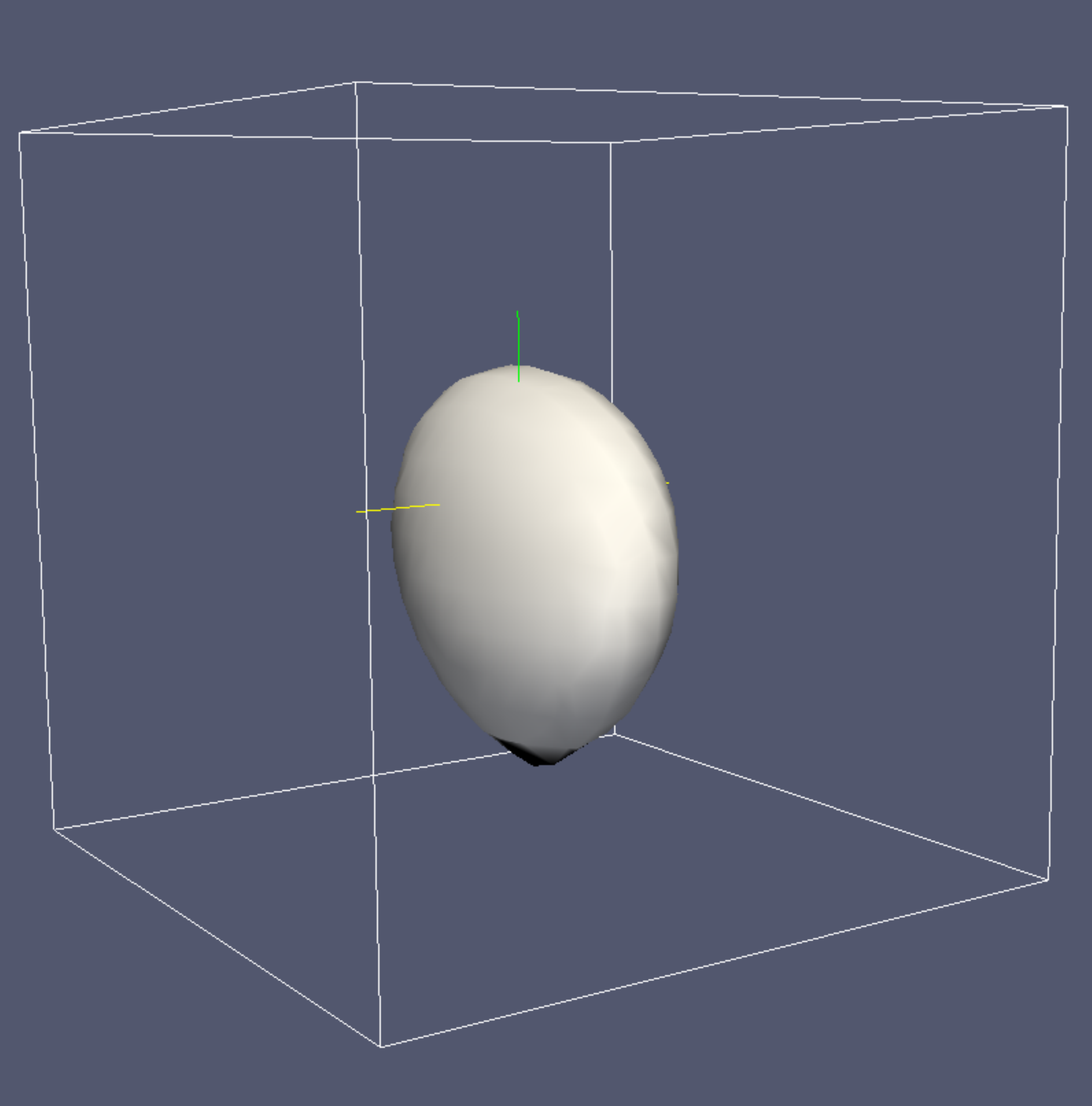}
        %{\bf (a)}
    \end{minipage}
    \begin{minipage}[b]{0.19\linewidth}
        \centering
        \includegraphics[width=1\textwidth]{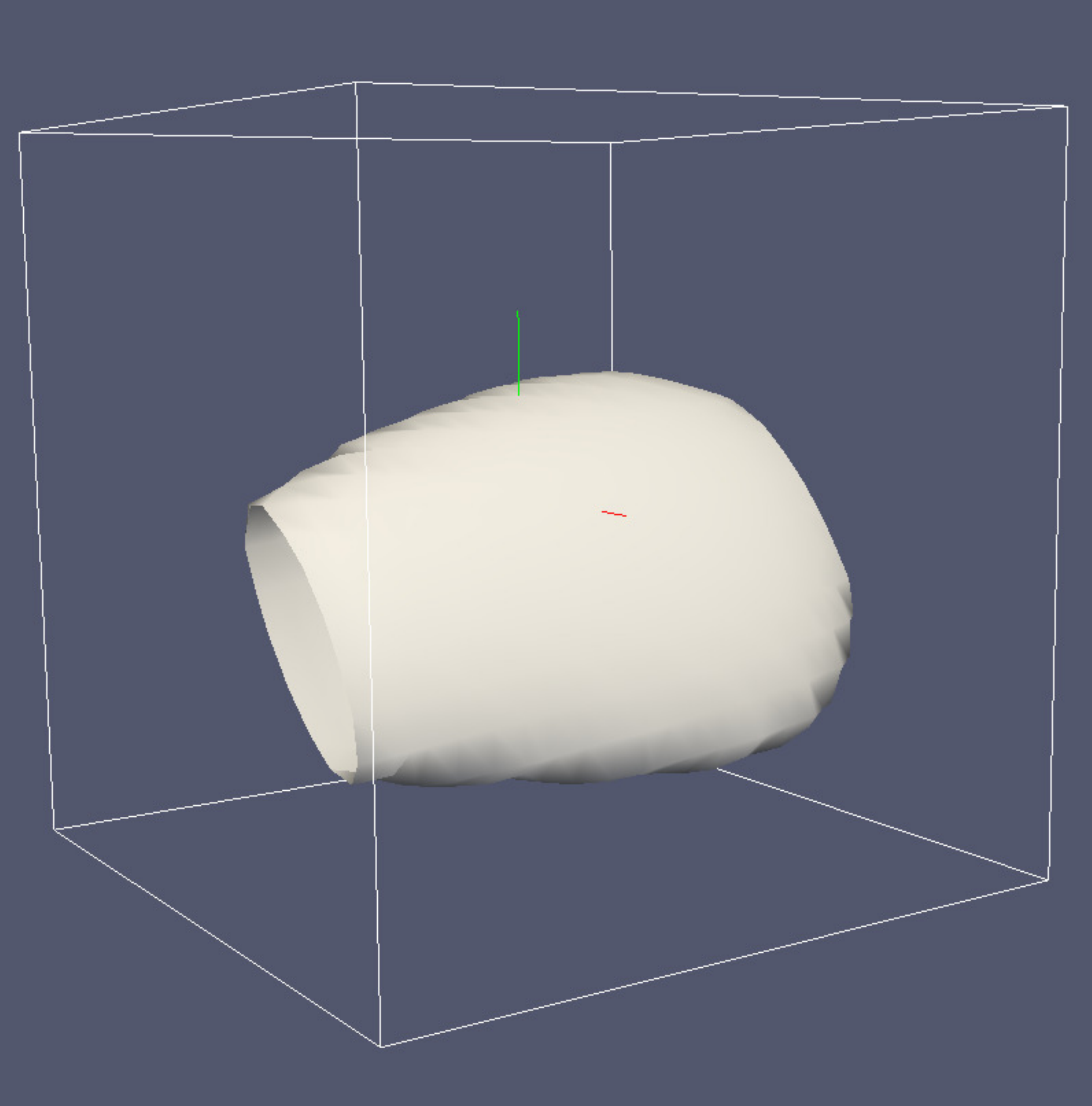}
        %{\bf (a)}
    \end{minipage}
    \begin{minipage}[b]{0.19\linewidth}
        \centering
        \includegraphics[width=1\textwidth]{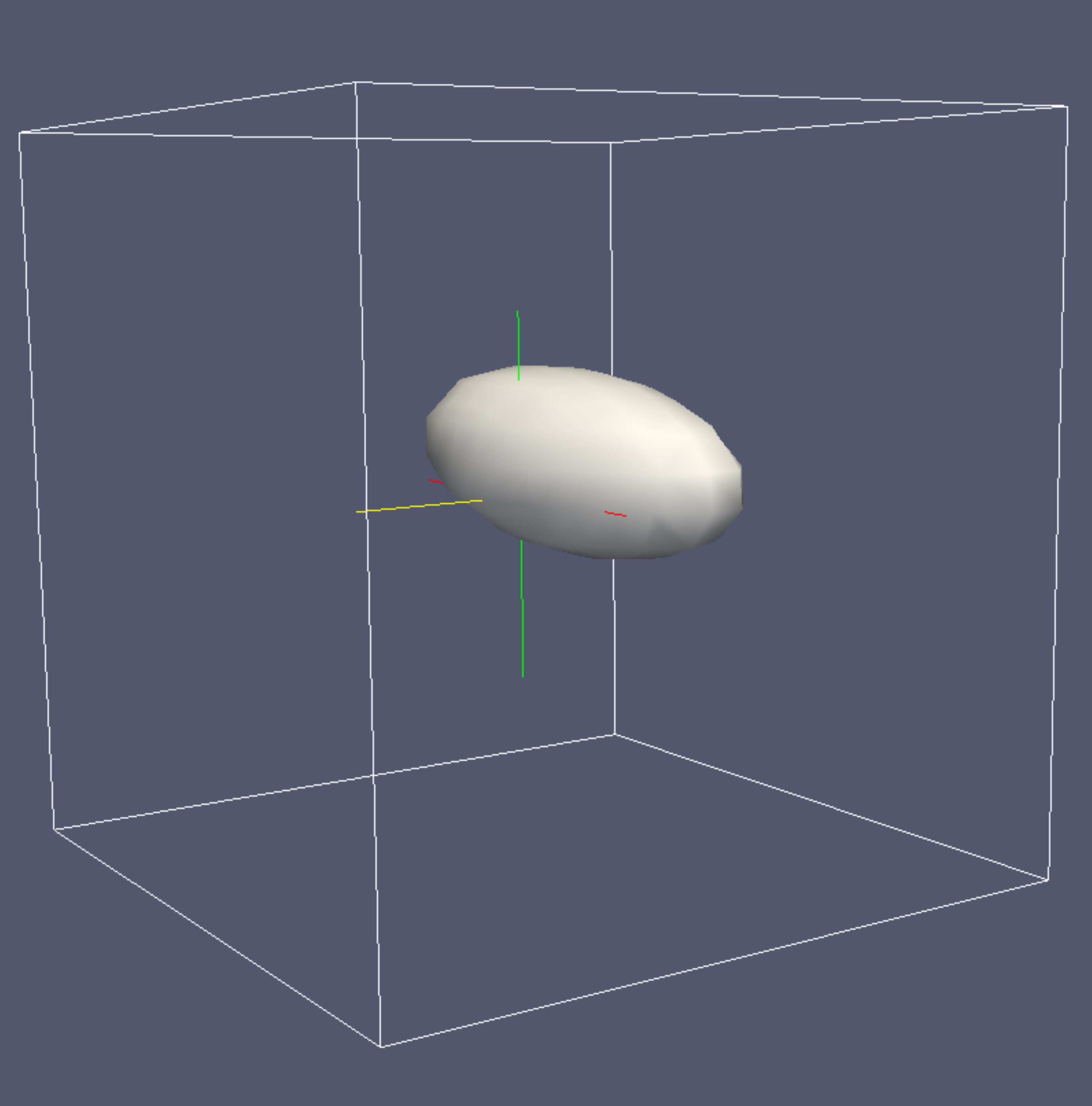}
        %{\bf (a)}
    \end{minipage}
    \begin{minipage}[b]{0.19\linewidth}
        \centering
        \includegraphics[width=1\textwidth]{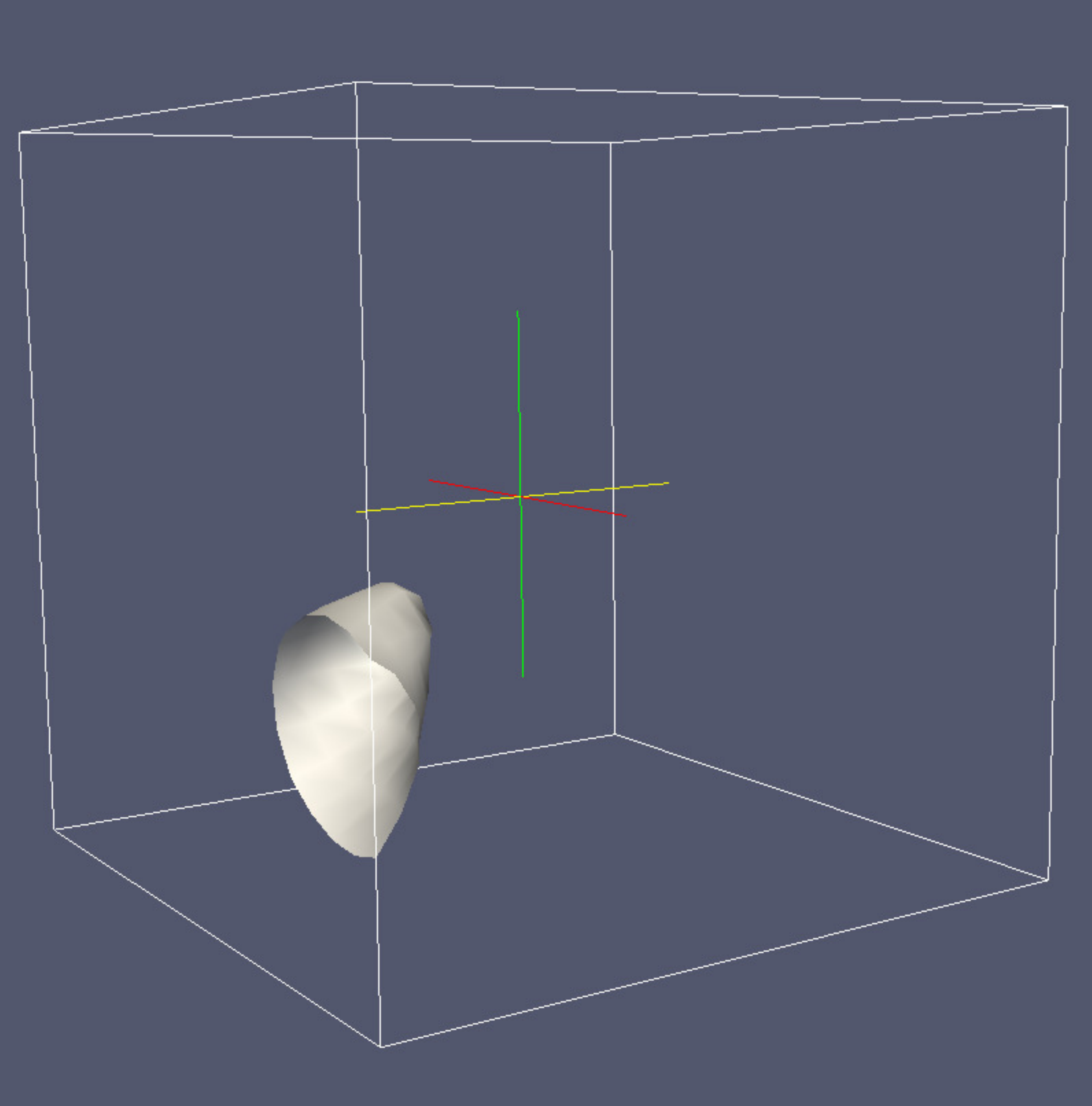}
        %{\bf (a)}
    \end{minipage}
    \begin{minipage}[b]{0.19\linewidth}
        \centering
        \includegraphics[width=1\textwidth]{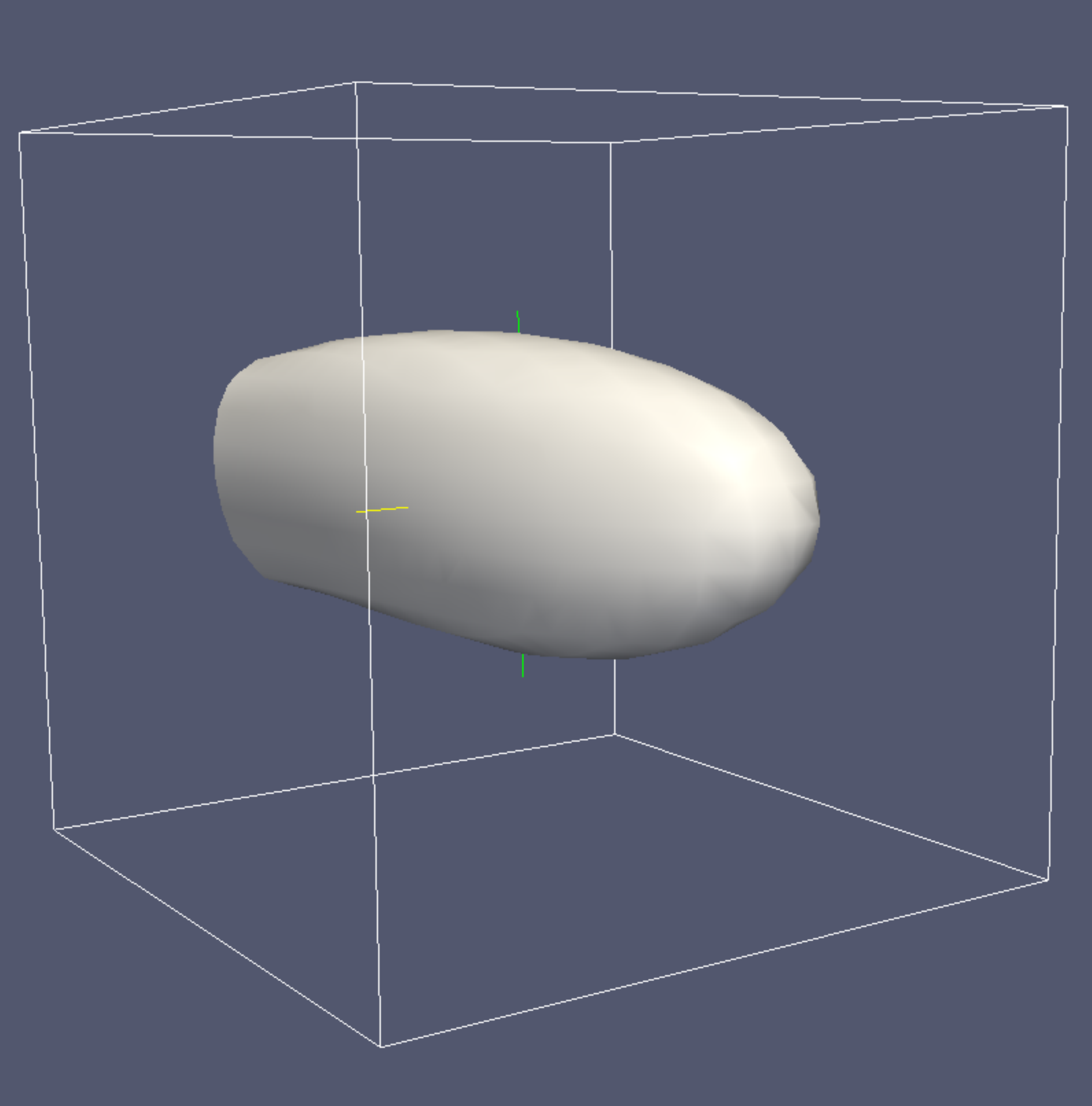}
        %{\bf (a)}
    \end{minipage}
    \begin{minipage}[b]{0.19\linewidth}
        \centering
        \includegraphics[width=1\textwidth]{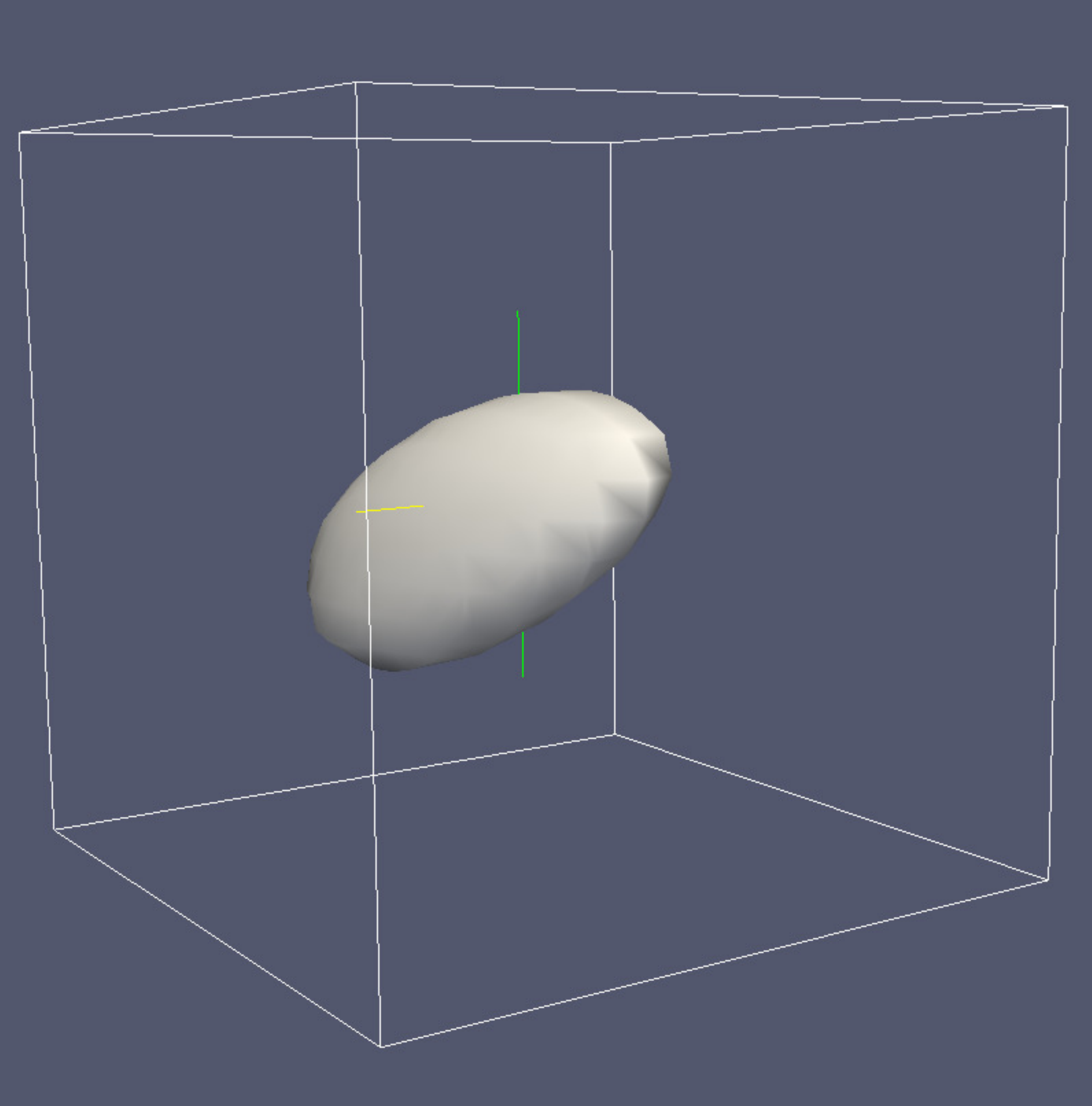}
        %{\bf (a)}
    \end{minipage}
    \begin{minipage}[b]{0.19\linewidth}
        \centering
        \includegraphics[width=1\textwidth]{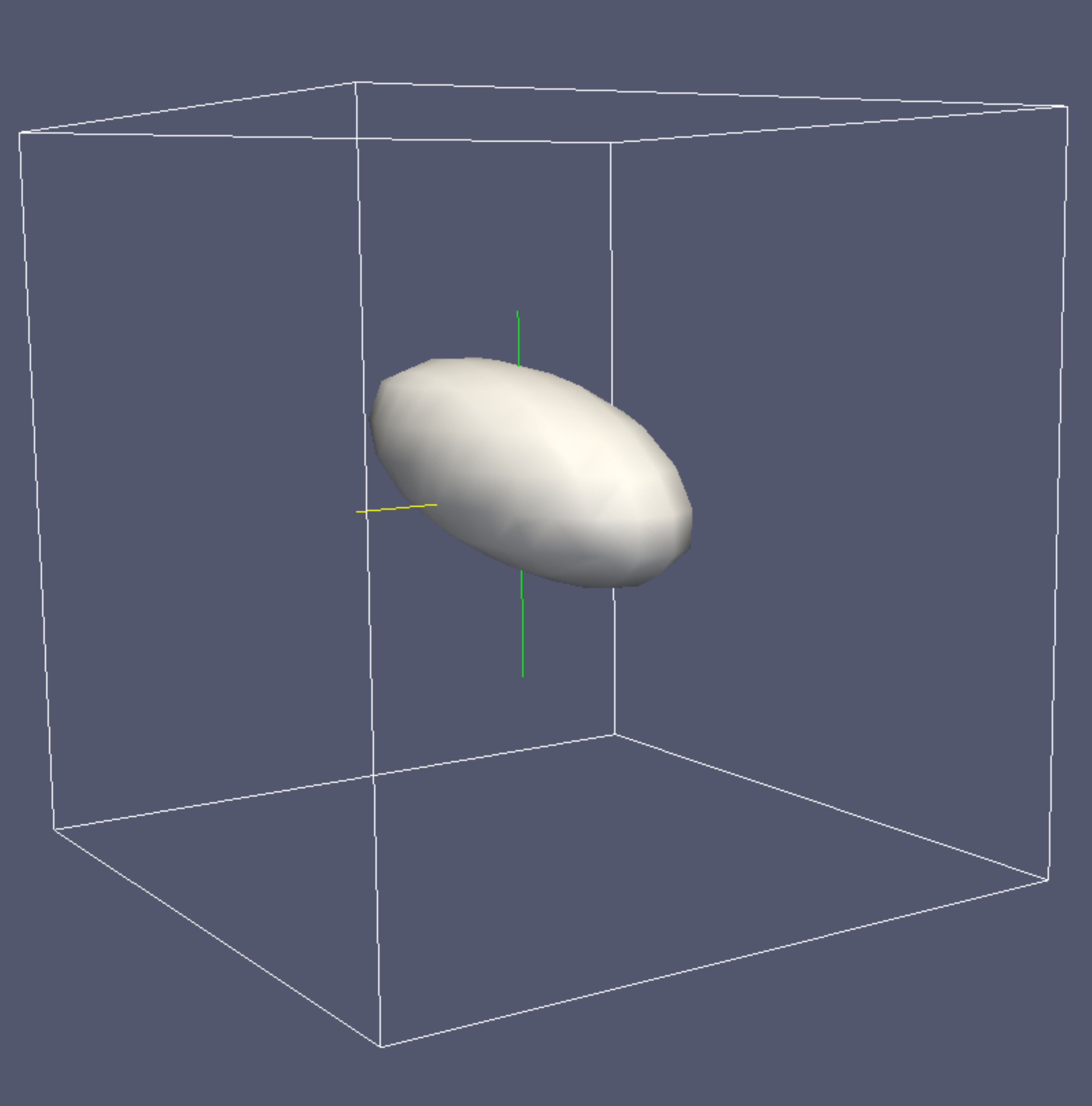}
        %{\bf (a)}
    \end{minipage}
    \begin{minipage}[b]{0.19\linewidth}
        \centering
        \includegraphics[width=1\textwidth]{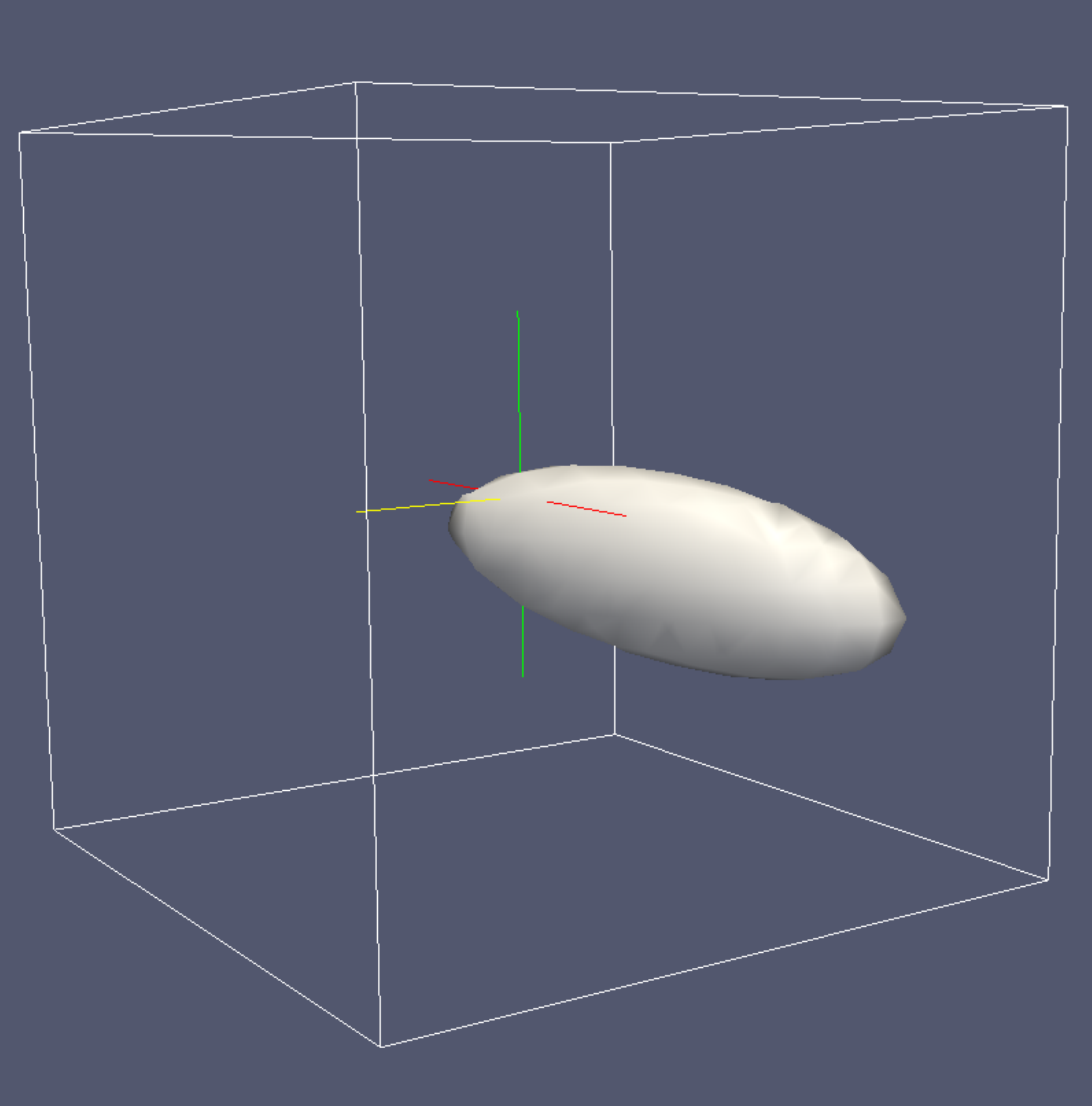}
        %{\bf (a)}
    \end{minipage}
    \begin{minipage}[b]{0.19\linewidth}
        \centering
        \includegraphics[width=1\textwidth]{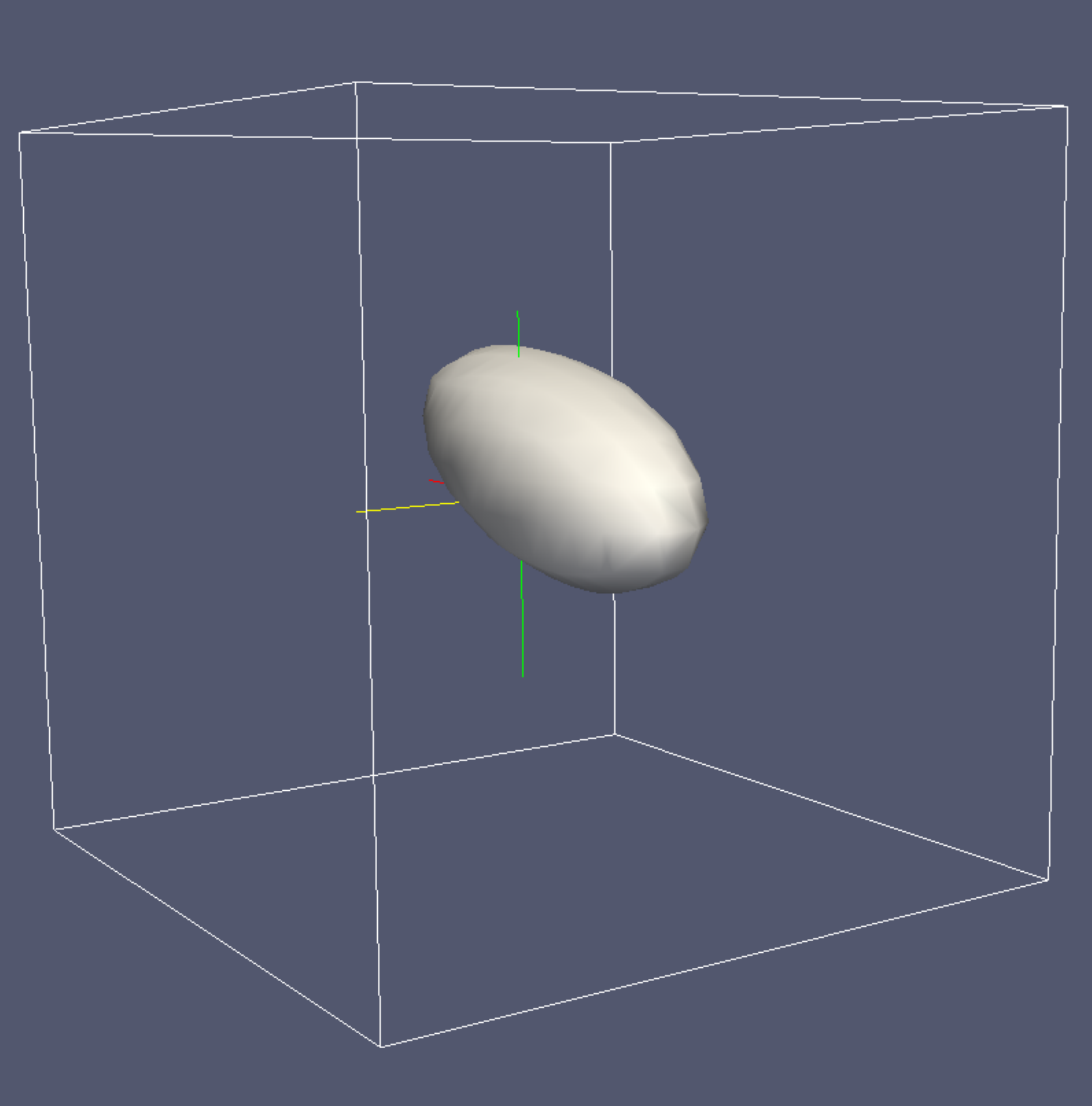}
        %{\bf (a)}
    \end{minipage}
    \begin{minipage}[b]{0.19\linewidth}
        \centering
        \includegraphics[width=1\textwidth]{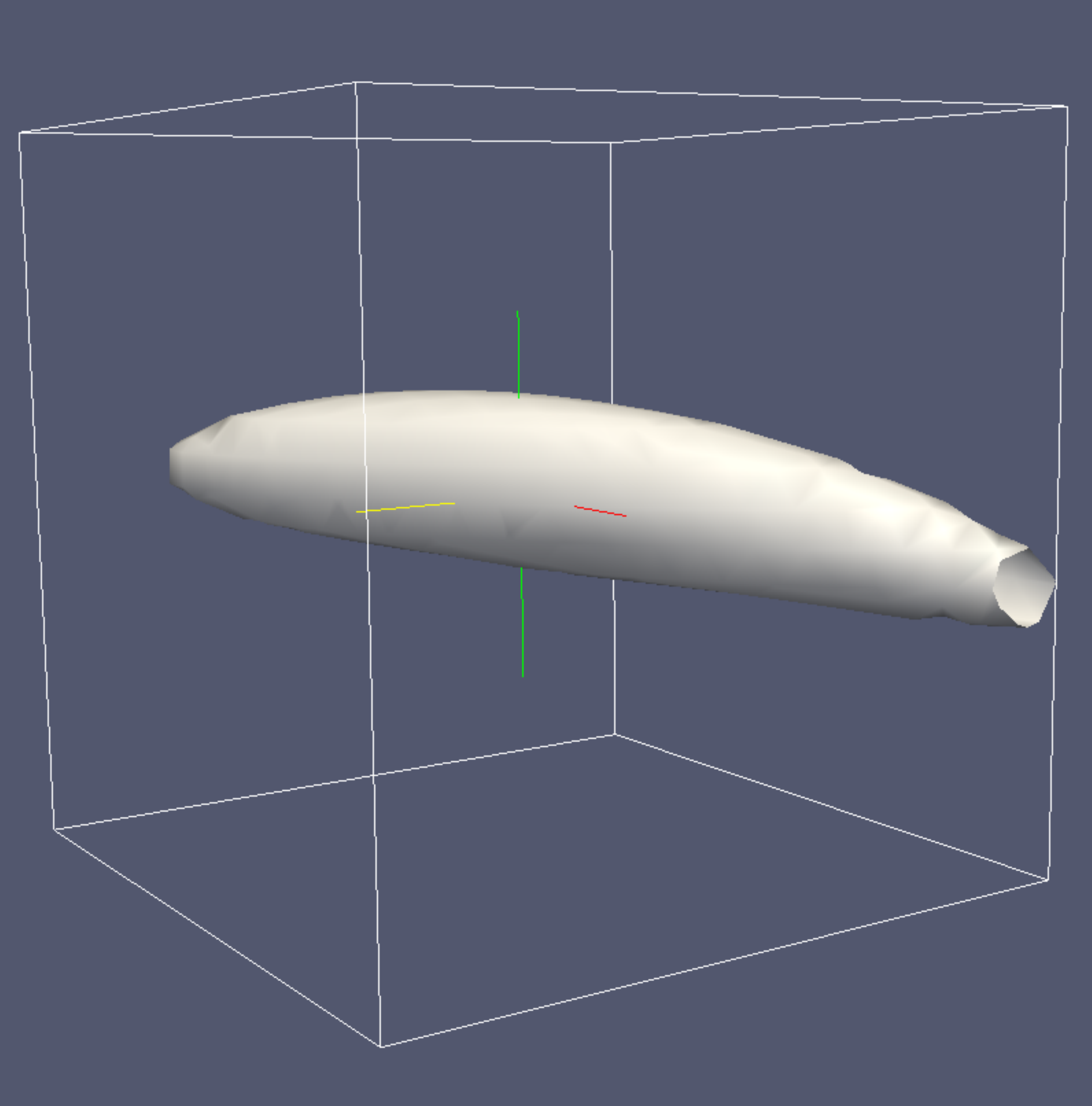}
        %{\bf (a)}
    \end{minipage}

    \begin{minipage}[b]{\linewidth}
        \centering
        {\it stable}
    \end{minipage}

    \hspace{2em}

    \begin{minipage}[b]{0.19\linewidth}
        \centering
        \includegraphics[width=1\textwidth]{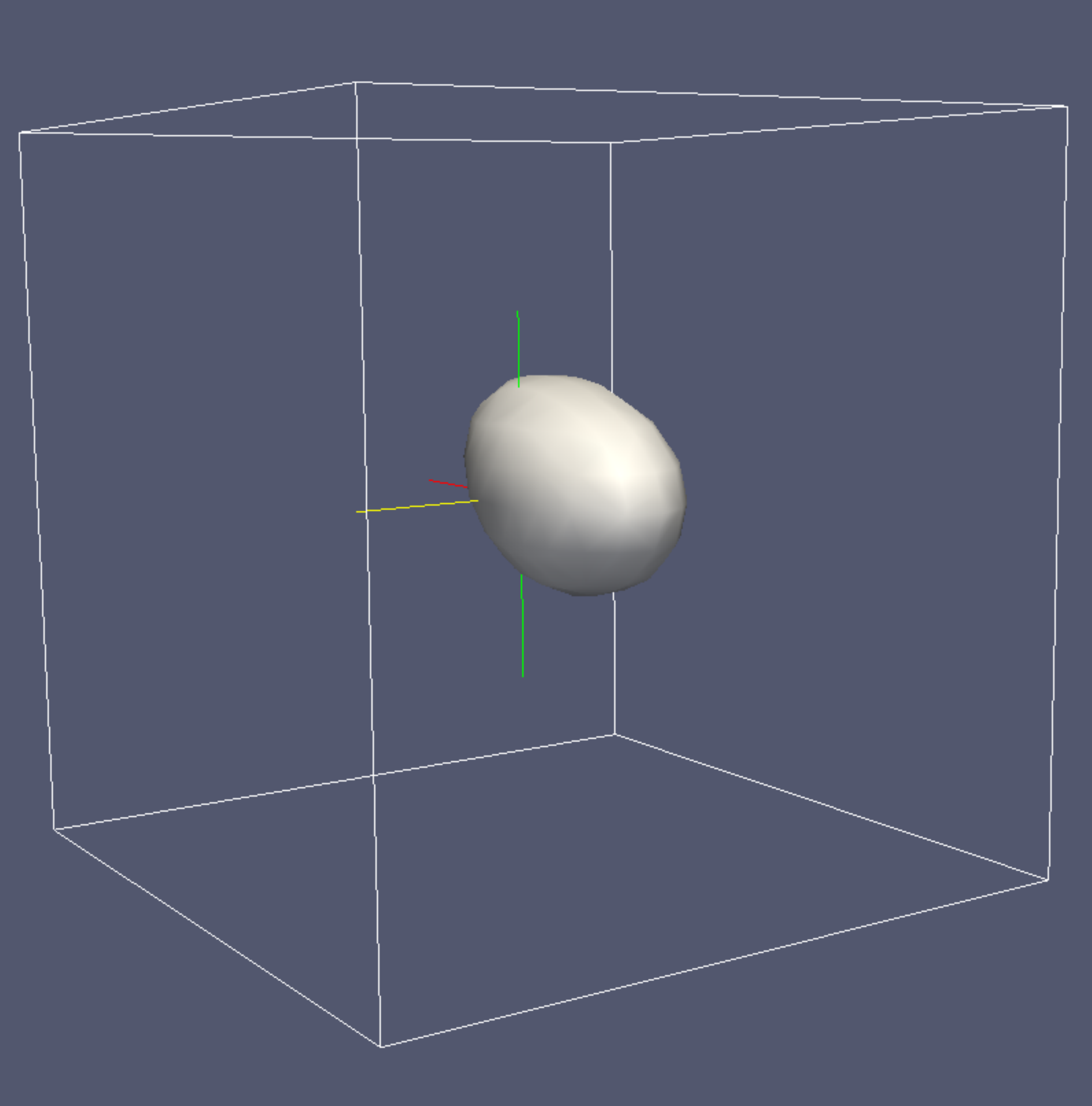}
        %{\bf (a)}
    \end{minipage}
    \begin{minipage}[b]{0.19\linewidth}
        \centering
        \includegraphics[width=1\textwidth]{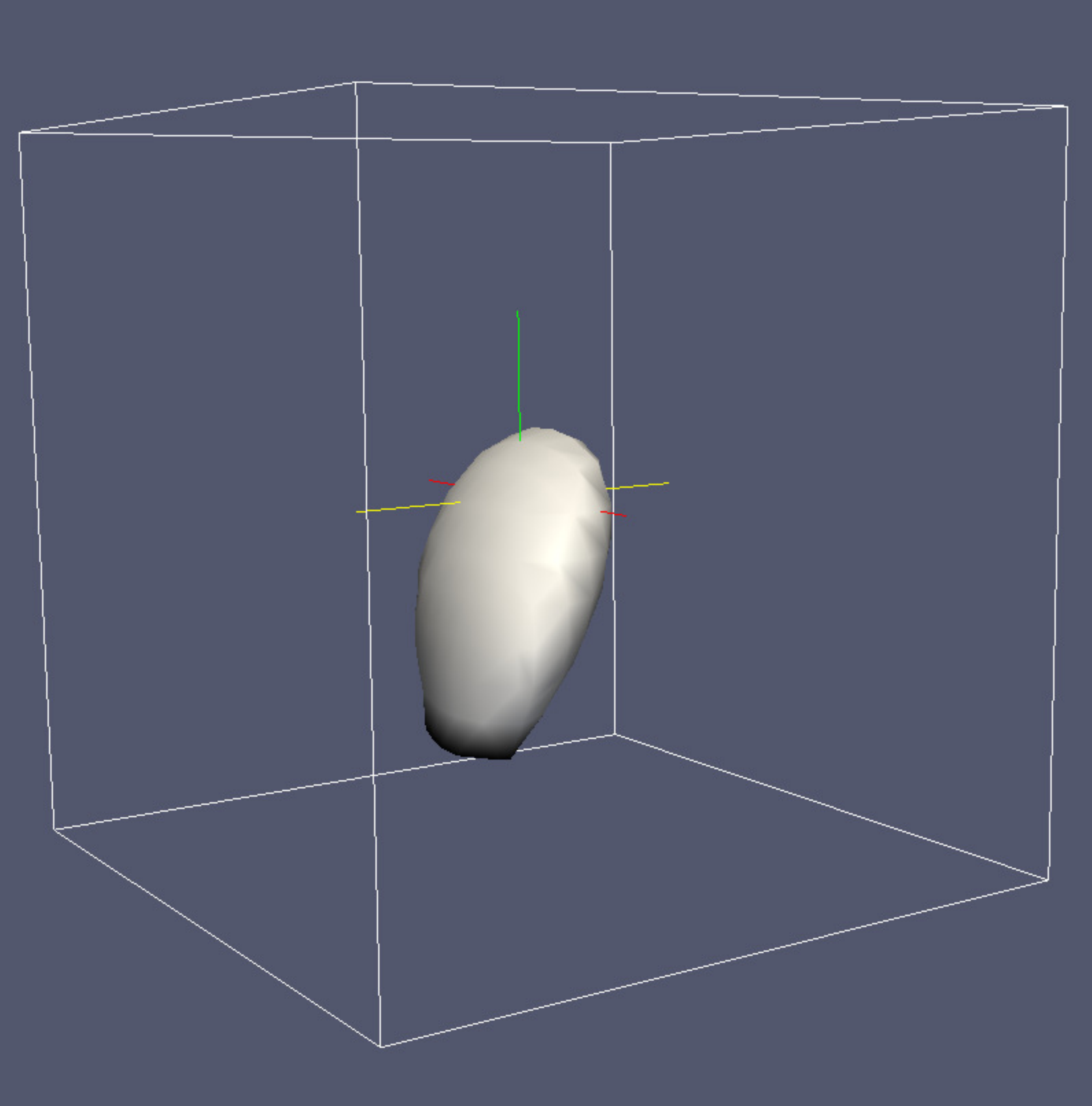}
        %{\bf (a)}
    \end{minipage}
    \begin{minipage}[b]{0.19\linewidth}
        \centering
        \includegraphics[width=1\textwidth]{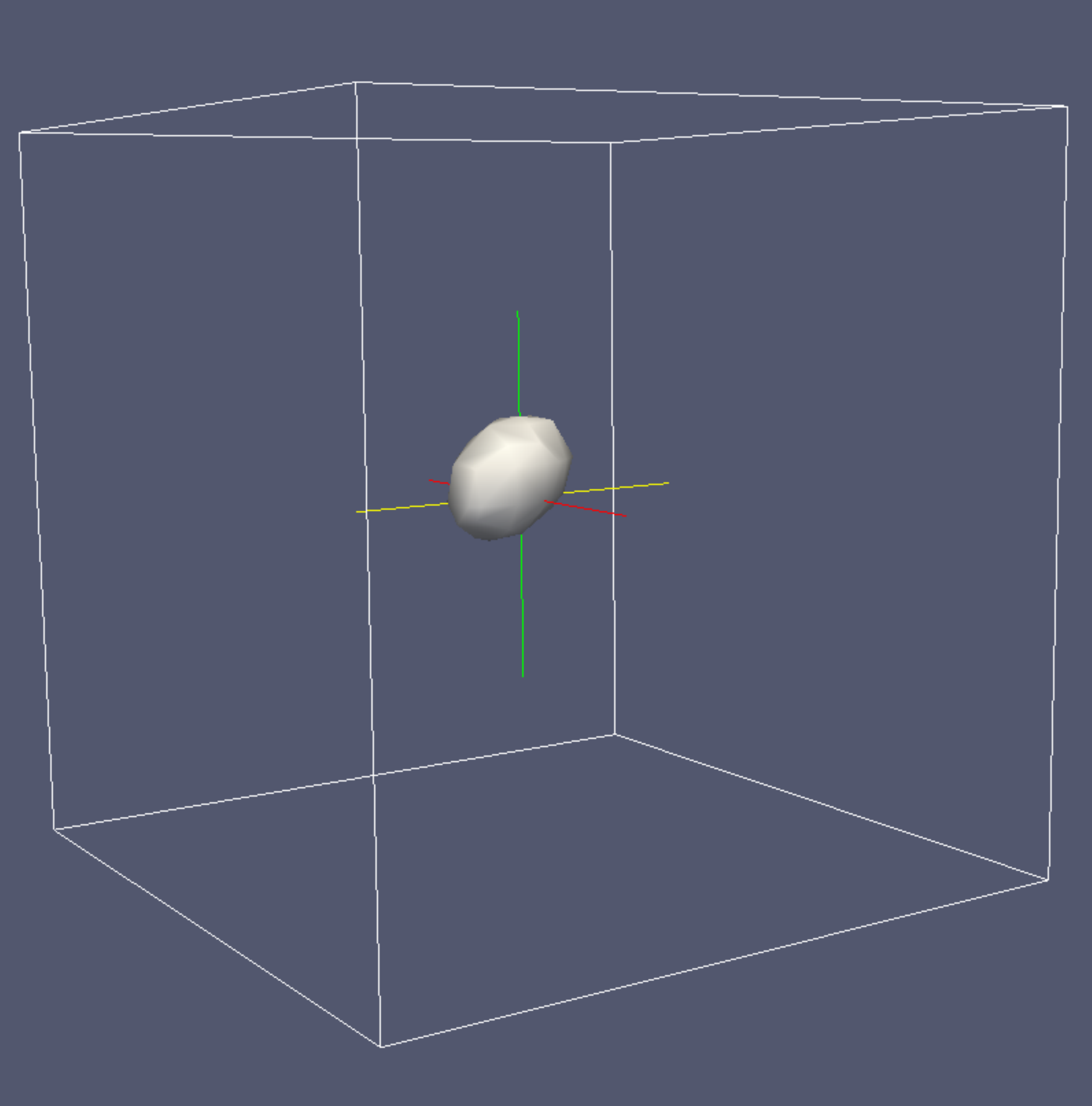}
        %{\bf (a)}
    \end{minipage}
    \begin{minipage}[b]{0.19\linewidth}
        \centering
        \includegraphics[width=1\textwidth]{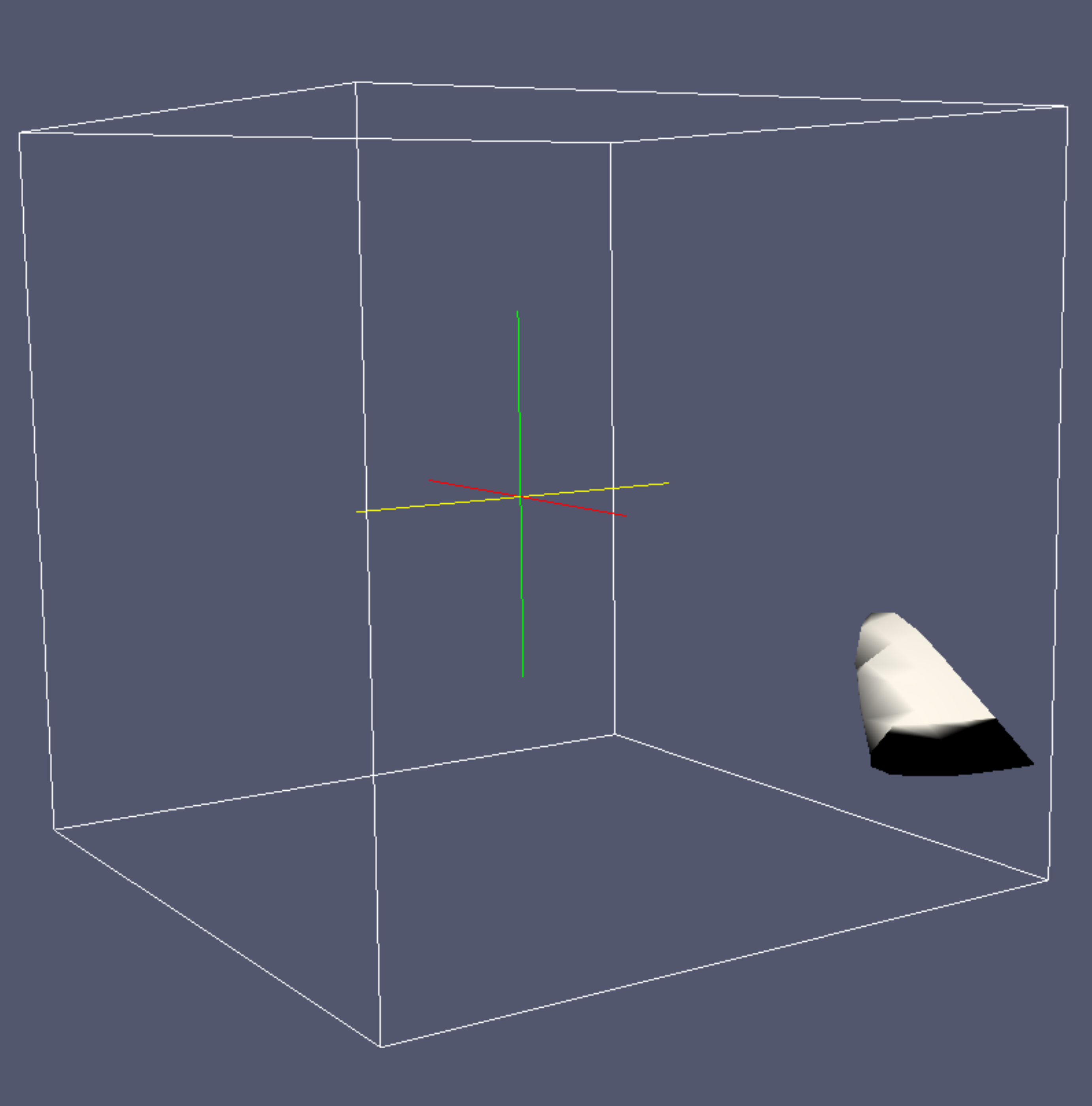}
        %{\bf (a)}
    \end{minipage}
    \begin{minipage}[b]{0.19\linewidth}
        \centering
        \includegraphics[width=1\textwidth]{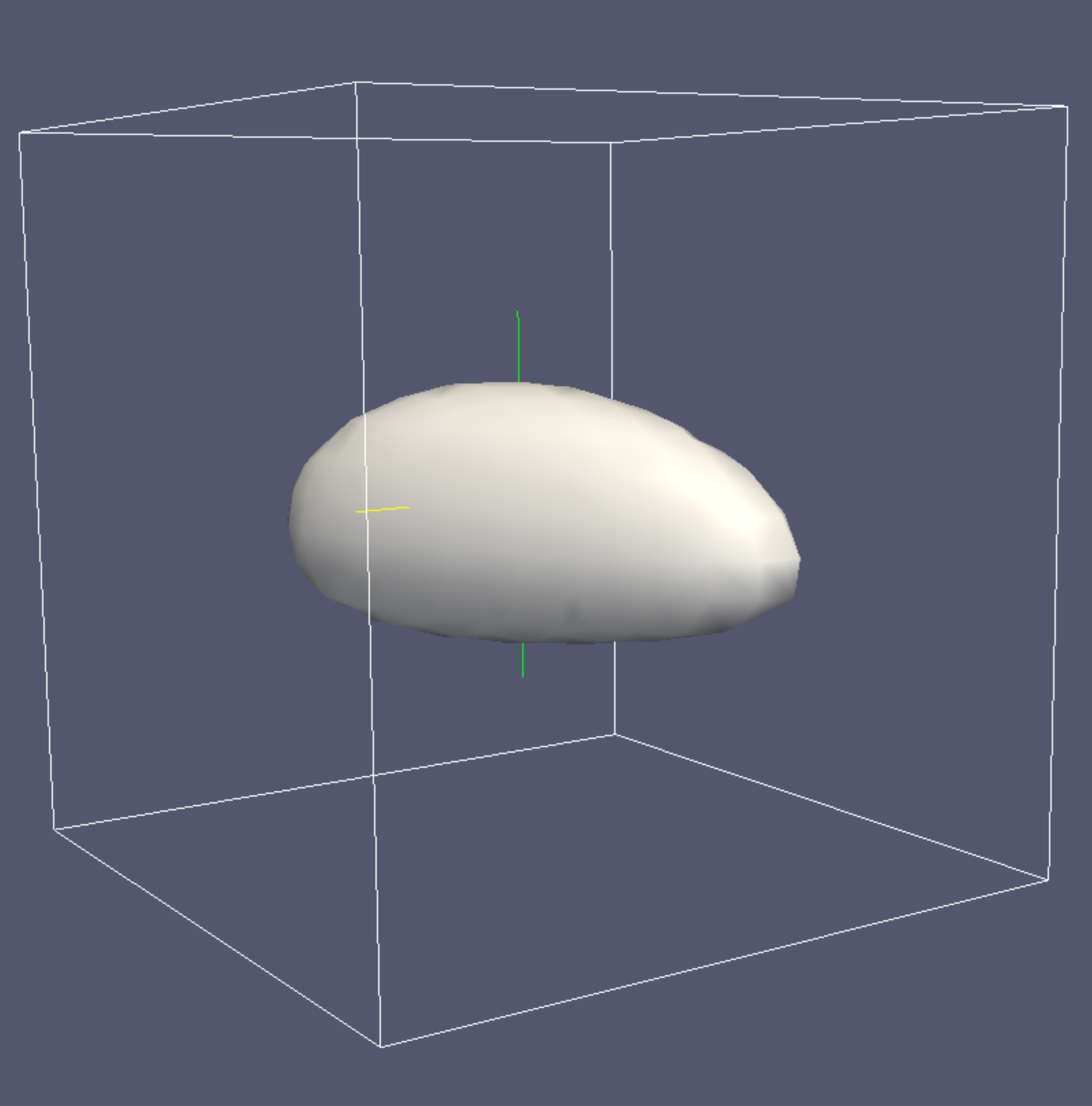}
        %{\bf (a)}
    \end{minipage}
    \begin{minipage}[b]{0.19\linewidth}
        \centering
        \includegraphics[width=1\textwidth]{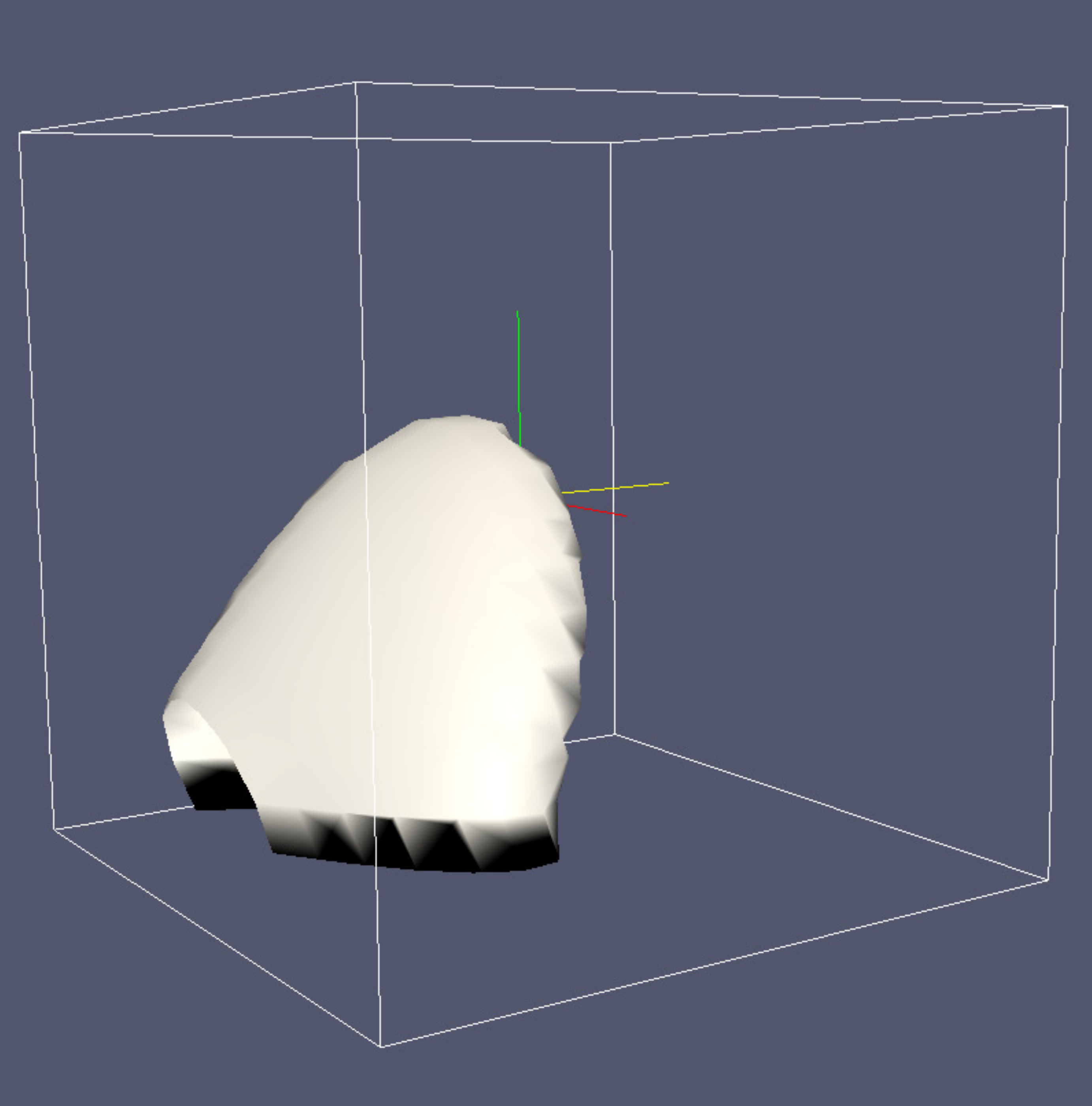}
        %{\bf (a)}
    \end{minipage}
    \begin{minipage}[b]{0.19\linewidth}
        \centering
        \includegraphics[width=1\textwidth]{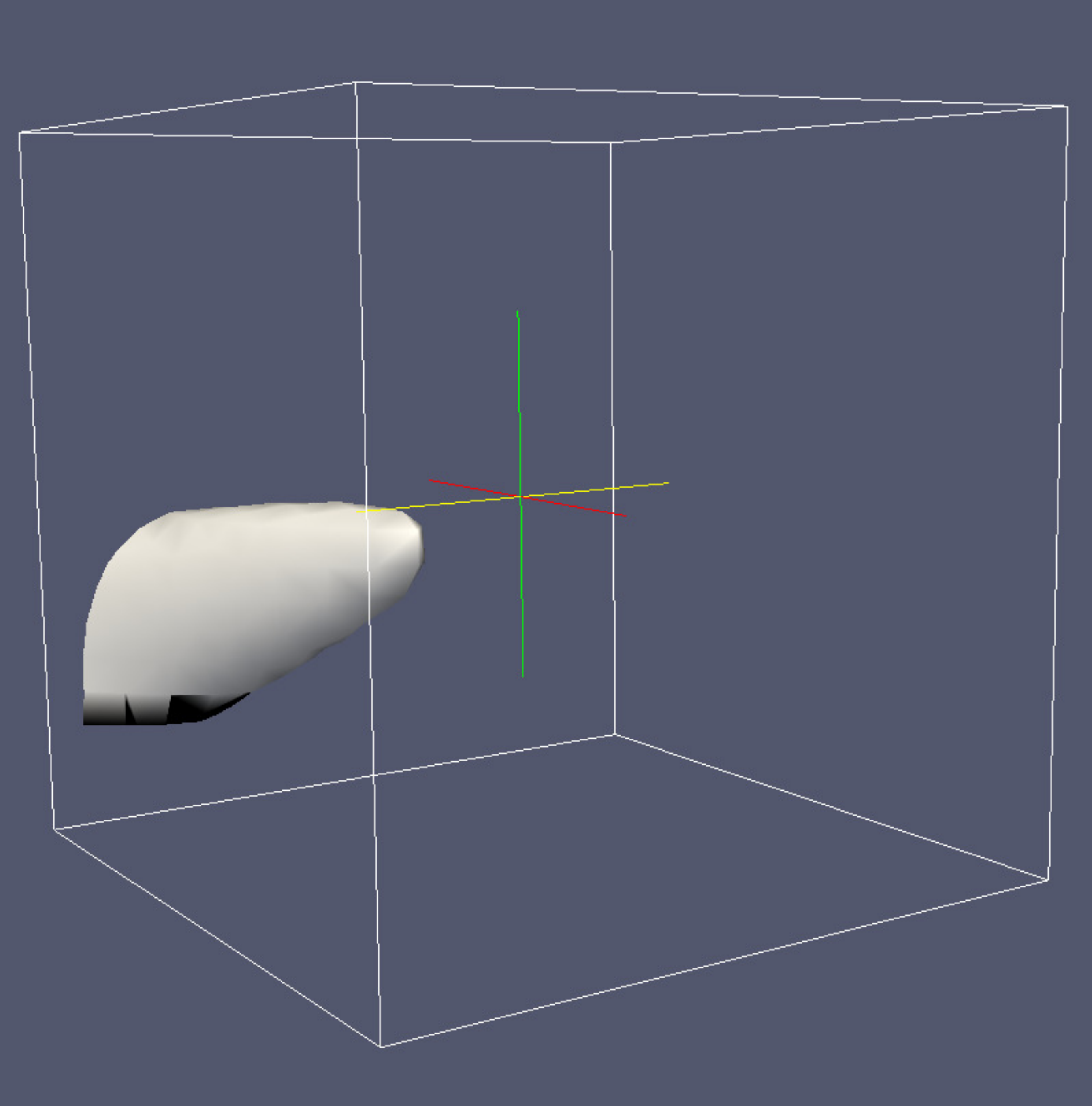}
        %{\bf (a)}
    \end{minipage}
    \begin{minipage}[b]{0.19\linewidth}
        \centering
        \includegraphics[width=1\textwidth]{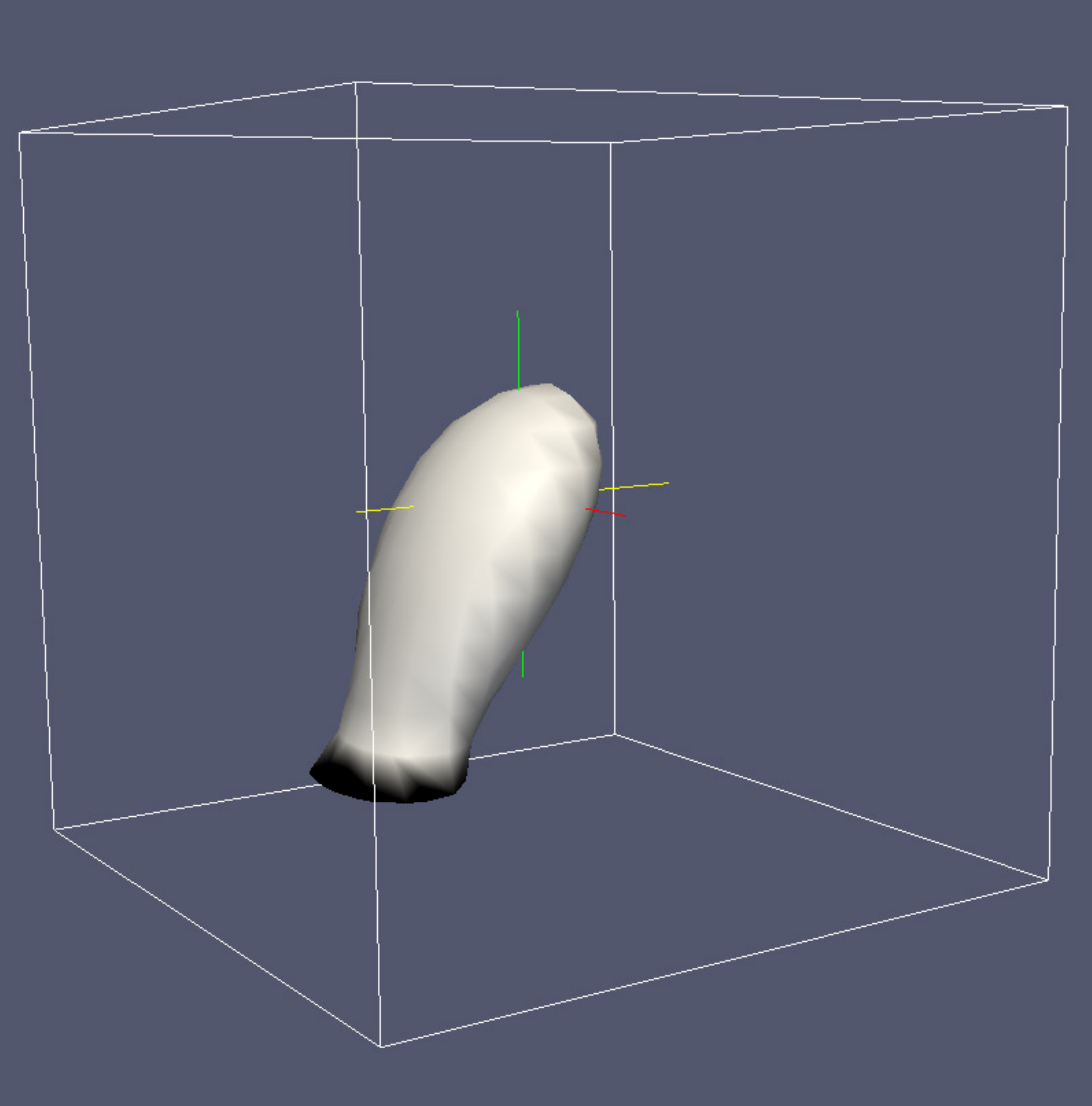}
        %{\bf (a)}
    \end{minipage}
    \begin{minipage}[b]{0.19\linewidth}
        \centering
        \includegraphics[width=1\textwidth]{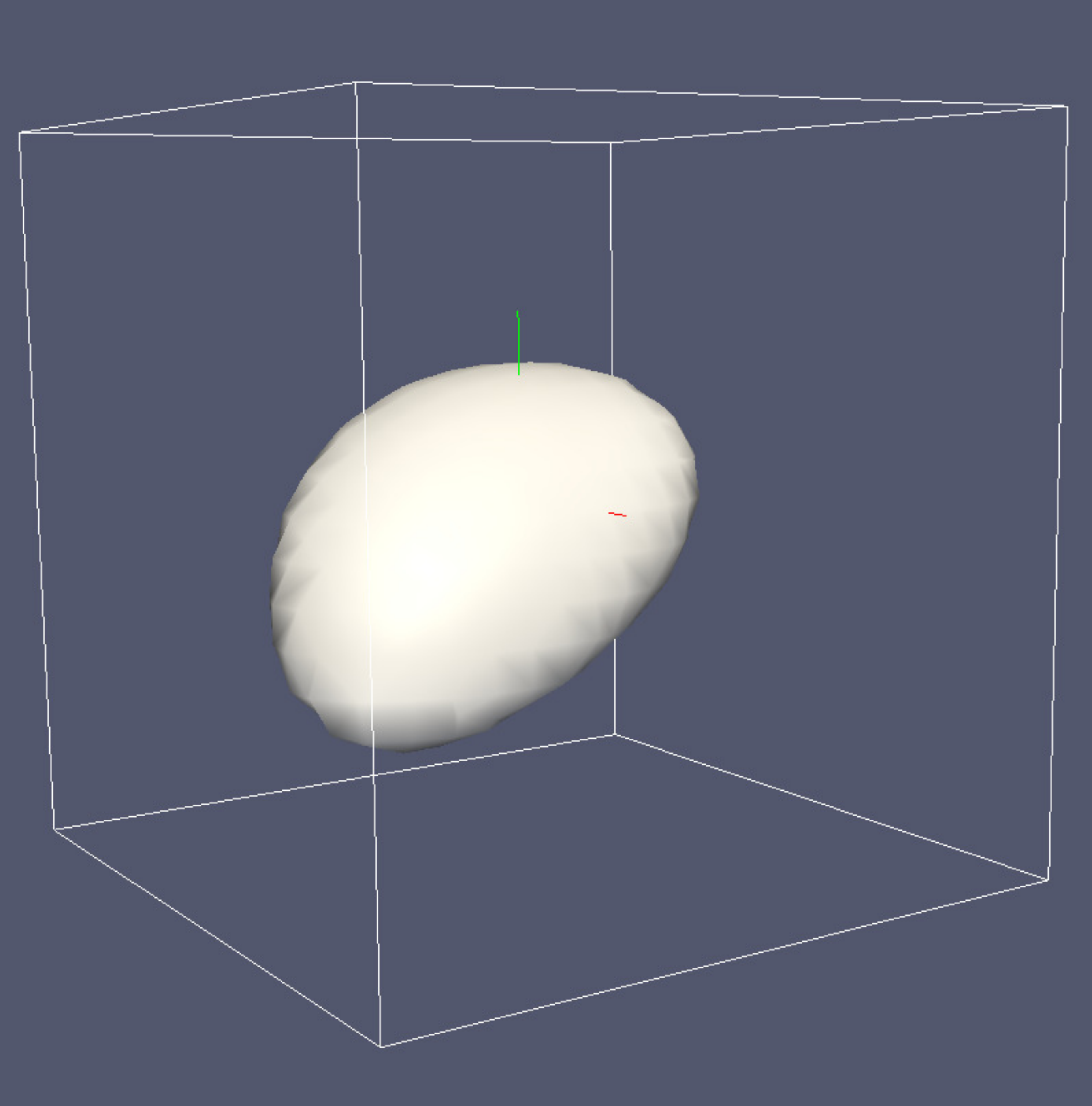}
        %{\bf (a)}
    \end{minipage}
    \begin{minipage}[b]{0.19\linewidth}
        \centering
        \includegraphics[width=1\textwidth]{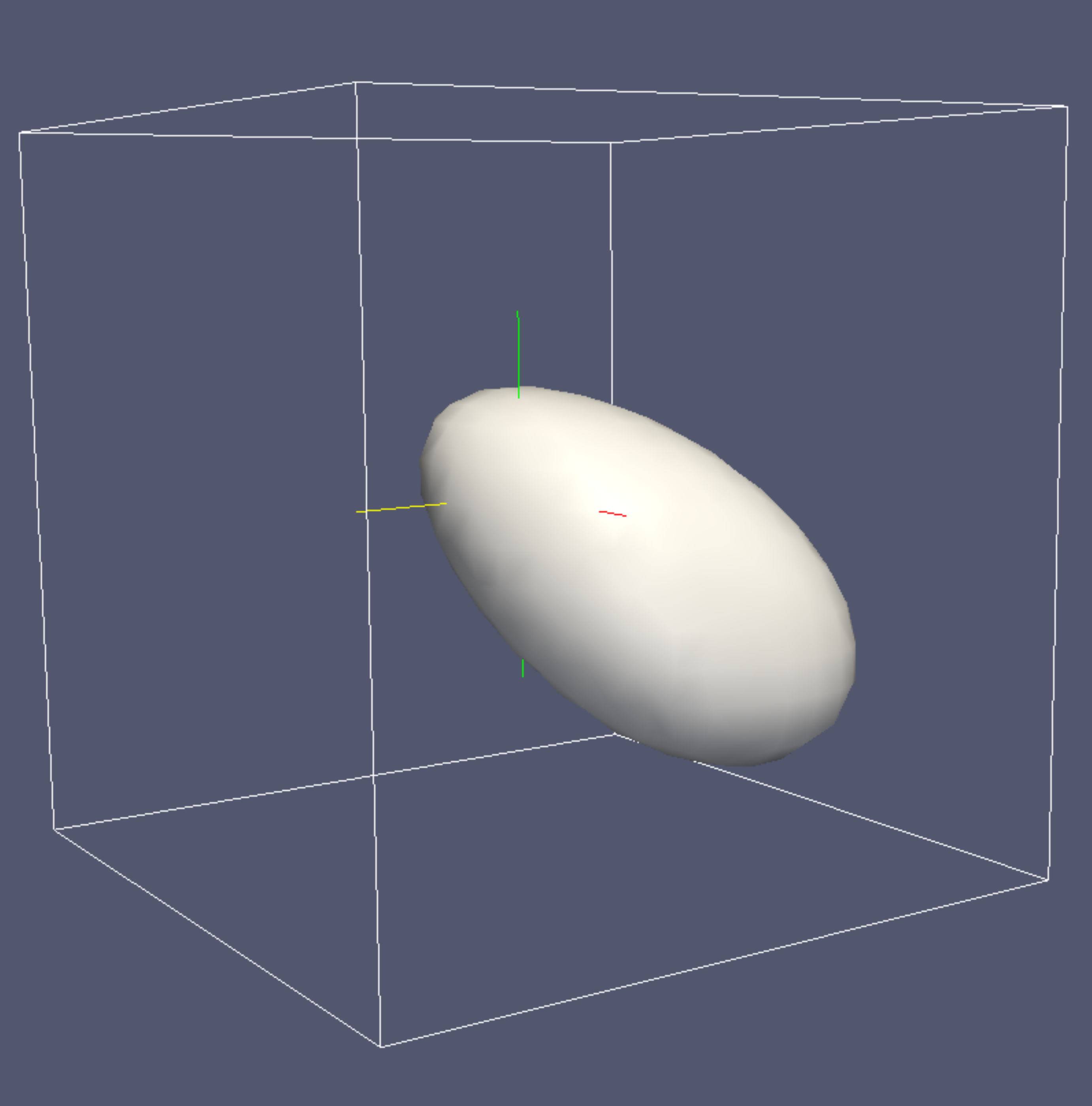}
        %{\bf (a)}
    \end{minipage}

    \begin{minipage}[b]{\linewidth}
        \centering
        {\it unstable}
    \end{minipage}
\end{center}
\caption{
    %
    %
 %   Illustration of the surrounding of candidates.
 %   %
 %   $25\times25\times25$ sample cubes are extracted around various extrema.
 %   %
 %   Using the marching cubes algorithm, iso-surfaces are computed for unstable
 %   extrema (left) and for stable extrema (right).
 %   %
    %
    Illustration of the DoG scale-space around detected keypoints.
    DoG Iso-surfaces are computed from a dense scale-space.
    %around keypoints for stable (top) and unstable
    %(below) extrema.
    %
    We observe a variety of configuration from isotropic shapes to elongated structures. 
    Furthermore, there seems to be no obvious connection between the local
    structures and the keypoint's stability level.
    %
    %
    %Unstable keypoints correspond in fact to elongated structures.
    %
 %   For such local configuration, certain sampling configuration can actually
 %   lead to the detection of \3d discrete extrema and eventually to
 %   interpolated extrema.
    %
    %
}
\label{fig:iso:surfaces}
\end{figure}

In an attempt to understand why the rudimentary detection and filtering procedures fail
to avoid spurious detections, we examined visually some of the detected scale-space local structures.
Figure~\ref{fig:iso:surfaces} shows  the DoG iso-surface computed around
several stable and unstable keypoints from a very dense scale-space.
Some detections are associated to isotropic shapes while others stem from
elongated structures.
There is no obvious link between how isotropic a structure is and its
overall stability. As shown in the figure, some elongated structures produce stable detections.
It seems therefore that a local analysis of the scale-space structure is not
sufficient to characterize unstable detections.

%\FloatBarrier
\subsection{The influence of extrema interpolation on stability, precision and invariance}
\label{sub:section:influence:extrema:interpolation}

The refinement of the discrete extrema position proposed in SIFT has two main
purposes.
First, it allows to locate the extrema to subpixel accuracy thanks to a local
continuous model of the DoG scale-space.
%
%But also, it allows to discard unreliable discrete extrema that cannot be precisely located.
%
But this refinement procedure also detects and discards unstable discrete
extrema.

In this section, we analyze the impact of the refinement procedure. To that
aim, we considered an input image and a series of transformations simulating
small displacements of the camera.
Although the analysis was restricted for a sake of simplicity to the case of
translations and scale changes, it could be easily generalized to more complex
image transformations such as perspective projections.

We examined the influence of the two main parameters in the refinement
procedure (see Section~\ref{sec:siftoverview}): the maximal number of allowed
interpolations $N_\text{interp}$, and the maximum offset $M_\text{offset}$
authorized for the extremum at each refinement iteration.
% that an extrema can be displaced.

\begin{figure}[htp]
\begin{center}
    \begin{minipage}[b]{0.48\linewidth}
        \centering
        
        \includegraphics[width=1\textwidth]{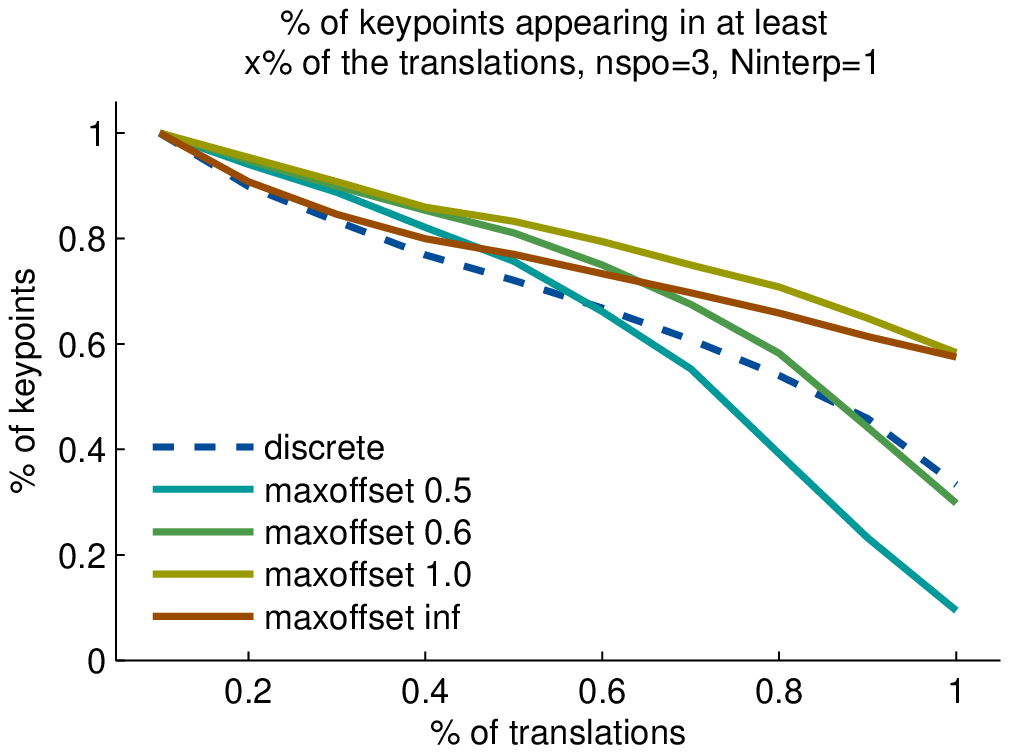} \vspace{-.5em}
        
        \includegraphics[width=1\textwidth]{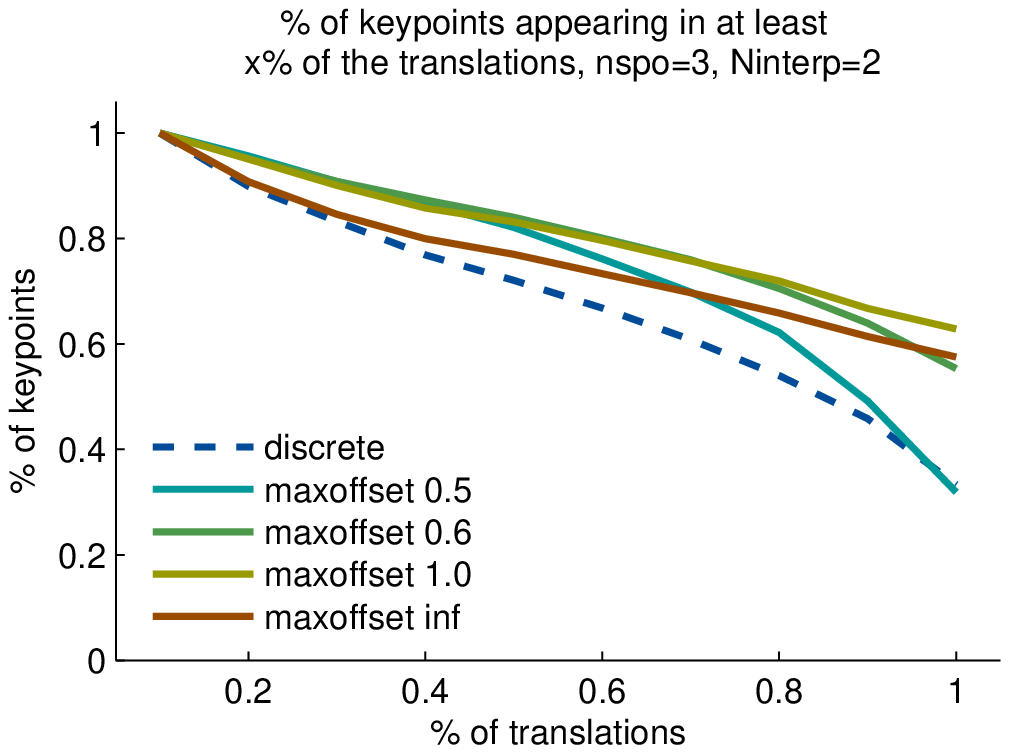} \vspace{-.5em}
        
        \includegraphics[width=1\textwidth]{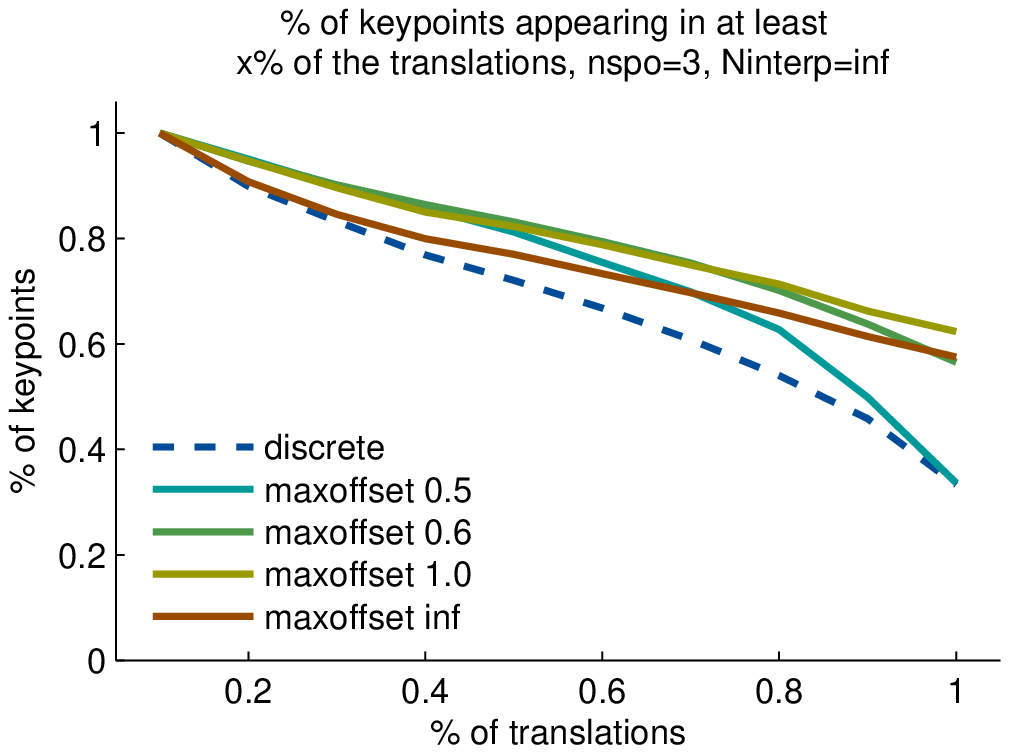} \vspace{-.5em}

           {\bf (a)}    $\nspo = 3$
           
    \end{minipage}
    \begin{minipage}[b]{0.48\linewidth}
        \centering
        
        \includegraphics[width=1\textwidth]{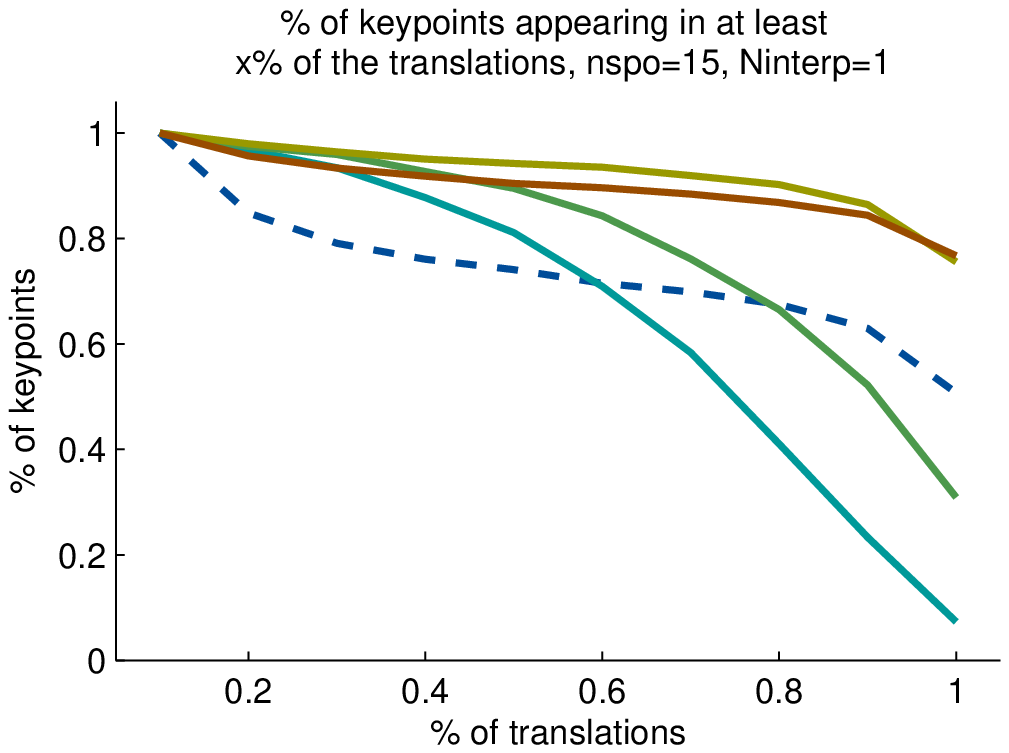} \vspace{-.5em}

        \includegraphics[width=1\textwidth]{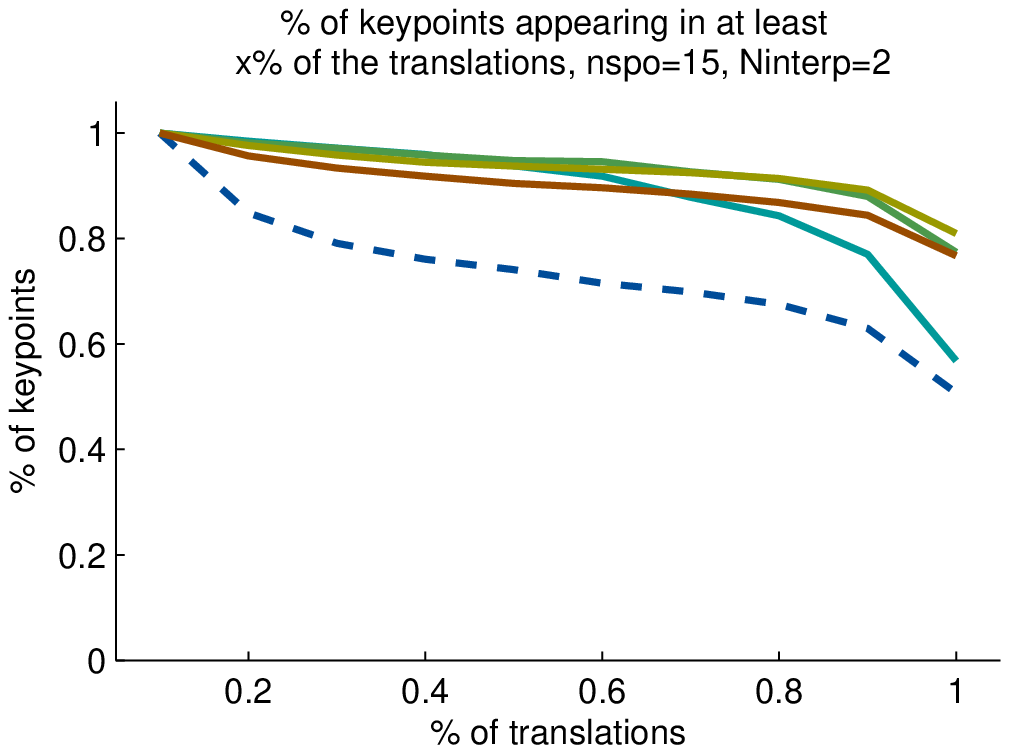} \vspace{-.5em}

        \includegraphics[width=1\textwidth]{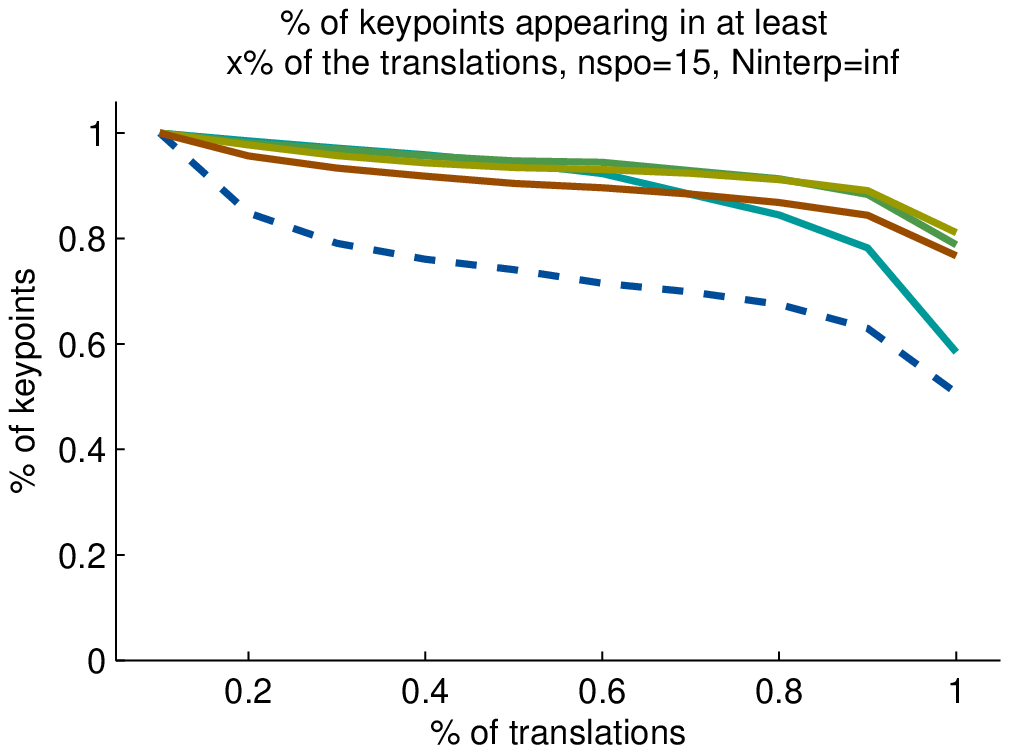} \vspace{-.5em}

         {\bf (b)}  $\nspo=15$
        
    \end{minipage}
    
    \vspace{1.2em}
    
     \begin{minipage}[b]{0.55\linewidth}
        \centering
        \includegraphics[width=.8\textwidth]{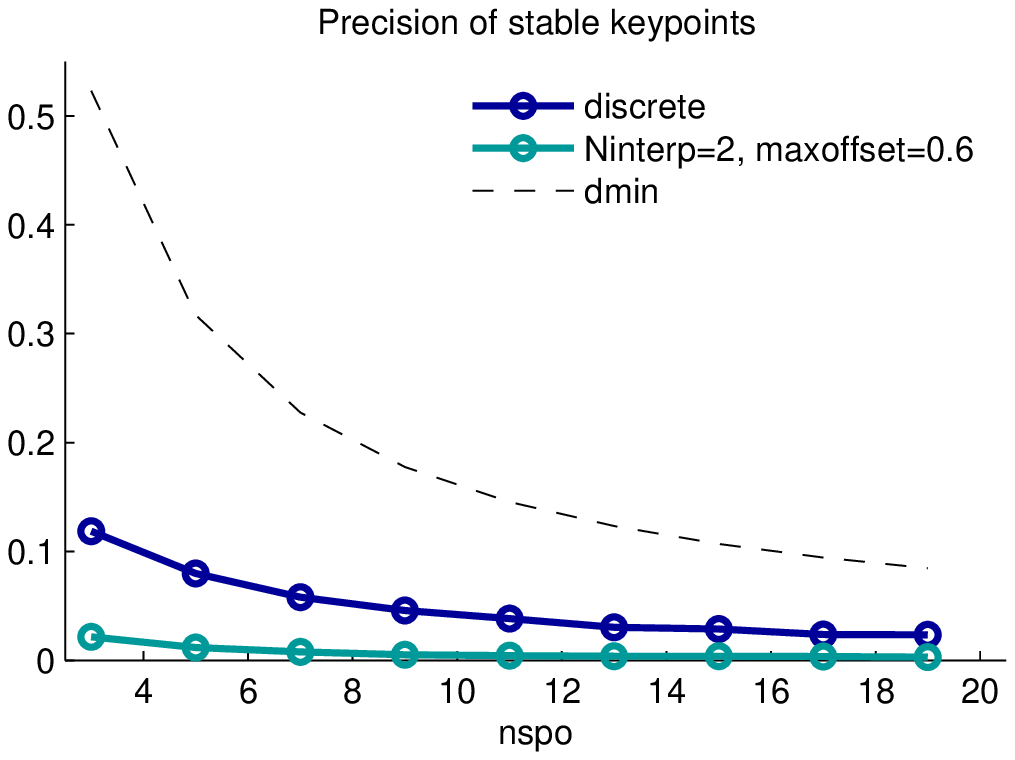}
        
        {\bf (c)}
    \end{minipage}

\end{center}
\caption{
    Influence of extrema refinement  parameters $M_\text{offset}$ and
    $N_\text{interp}$ on the detection stability/precision.
    A set of translated images was simulated and the keypoints  extracted.
    Each curve shows the percentage of unique keypoints appearing in at least a
    certain percentage of the simulated image translations for different values
    of $M_\text{offset} = 0.5, 0.6, 1.0, \infty$.
    The plots in the first, second and third row were generated considering a
    maximum number of interpolations $N_\text{interp}=1,2 \text{ and } \infty$
    respectively.
    The left block of plots {\bf (a)} was generated by sampling the scale-space
    with $\nspo=3$ (and the corresponding $\dmin$), while the right block {\bf
    (b)} was generated using $\nspo=15$.
    Allowing two iterations ($N_\text{interp}=2$) and a maximal offset of
    $M_\text{offset}=0.6$ gives the best performance in terms of stability of
    detected keypoints.
    Allowing for more interpolations attempts did not increase the performance,
    as can be seen by comparing the third row to the second row.
    {\bf (c)} shows the influence of the extrema refinement on the precision of
    the stable set of keypoints (appearing in at least $50\%$ of the simulated
    images).
    In this pure translation scenario, it appears that the precision of the
    detected extrema significantly increases  when using  extrema interpolation
    and when  sampling finely the scale-space (e.g., $\nspo > 3$).
   }
\label{fig:influ:interp:trans}
\end{figure}

Our performance measure was the \emph{stability}, measured by considering
the number of keypoints that appear in at least a certain percentage of the
simulated image transformations.
A perfectly stable keypoint would be one that appears in all the simulated images,
while a perfectly unstable keypoint would be one that only appears in one of the
images.
We also measured the \emph{precision} by computing the average standard
deviation of the location of the stable keypoints, where keypoints were
considered stable if they appeared in at least $50\%$ of the simulated
transformations.
Figure~\ref{fig:influ:interp:trans}~{\bf (a,b)} shows the percentage of unique
keypoints that appear in at least a given percentage of the translations for
different values of $M_\text{offset}$.
Each figure corresponds to a given sampling rate ($\nspo = 3 \text{ and } 15$)
and a given maximal number of interpolations ($N_\text{interp}=1,2,\infty$).
Ideally, one would like to have a large proportion of stable detections, which
would correspond to a flat curve.
The percentage of detections for the SIFT sampling rate ($\nspo=3$) decreases quickly  when considering only the more stable ones, present in a
large percentage of the simulated transformations.
On the other hand, $\nspo=15$ leads to flatter curves, which implies more
stable detections, and demonstrates that increasing the scale-space sampling
improves stability.
The refinement of the extrema helps discard the unstable ones.

The fact that the results with $N_\text{interp}=2$ and $N_\text{interp}=\infty$
are identical (second and third row of Figure~\ref{fig:influ:interp:trans}),
implies that there is no extra benefit in allowing more than two iterations.
The present analysis indicates that allowing a maximum of two interpolations
($N_\text{interp}=2$) in combination with a maximum displacement of
$M_\text{offset} = 0.6$ produce on average keypoints that are more stable.
This conclusion is independent of the considered $\nspo$.
Therefore, for the remainder of the article, we consider the refinement step
with these two values.
%
%\jmC{
%    %
%    Je ne comprends pas cette conclusion qui semble contredite par la figure 6,
%    en ce qui concerne l'offset. En effet c'est la courbe marron qui est
%    toujours au dessus des autres, qui est la plus plate, et qui donc la
%    meilleure.
%    %
%    Donc ces courbes impliquent qu'il faille ne pas imposer de seuil
%    sup\'erieur sur le d\'ecalage (offset) autoris\'e (ce qui est quand même très
%    surprenant). Y aurait-il une erreur dans le code de couleur?
%    %
%    Par ailleurs les figures  montrent le nombre d'extrema discrets en
%    pointill\'es. Il me semble qu'il faudrait expliquer pourquoi cette courbe
%    est presque tout le temps au dessus des autres courbes.
%    %
%    Je suppose que c'est explicable par le fait que le raffinement supprime des
%    extrema instables.
%}
%\ivC{
%    %
%    FIXED : Show the percentage of unique points as function of the percentage of
%    (least) translations.
%    %
%    Maybe the curve for $\infty$ offset will go down at the end, showing that
%    overall, the keypoint discarded with the offset are unstable.
%    %
%}

Increasing the scale-space sampling rate in conjunction with extrema
interpolation has a tremendous impact on the detection precision.
Figure~\ref{fig:influ:interp:trans} shows for both, discrete and interpolated
detections, the mean of the precision of stable keypoints (appearing in at
least $50\%$ of translations) as a function of the scale-space sampling rate.

\begin{figure}[htp]
\begin{center}
    \begin{minipage}[b]{0.48\linewidth}
        \centering
        \includegraphics[width=1\textwidth]{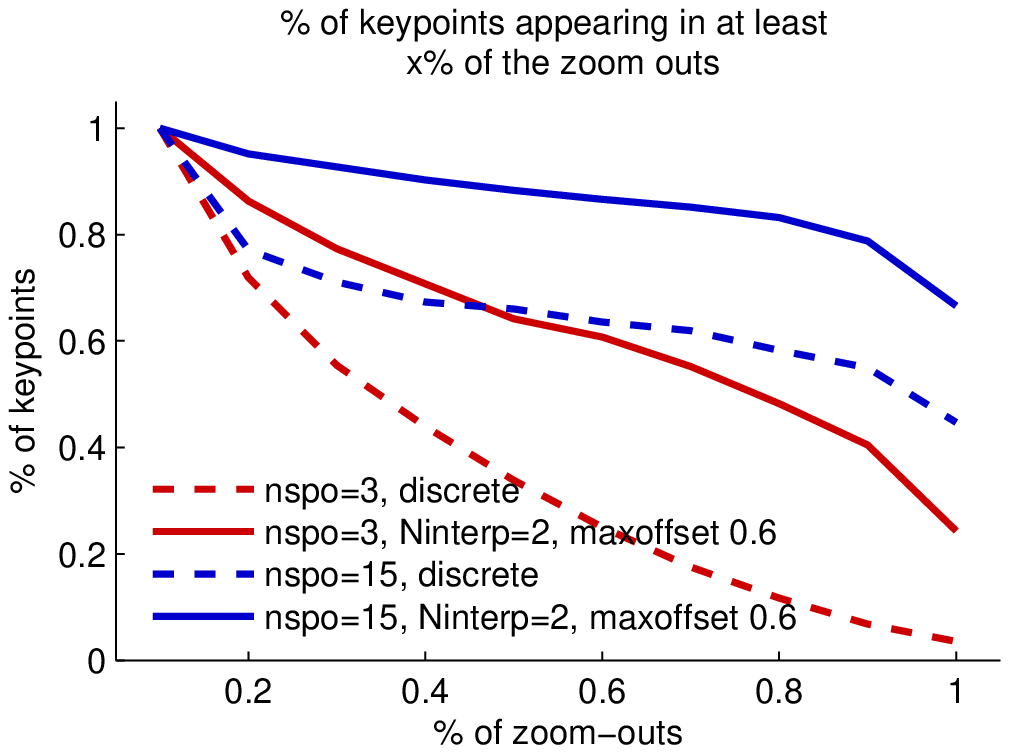}
        {\bf (a)}
    \end{minipage}
    \begin{minipage}[b]{0.48\linewidth}
        \centering
        \includegraphics[width=1\textwidth]{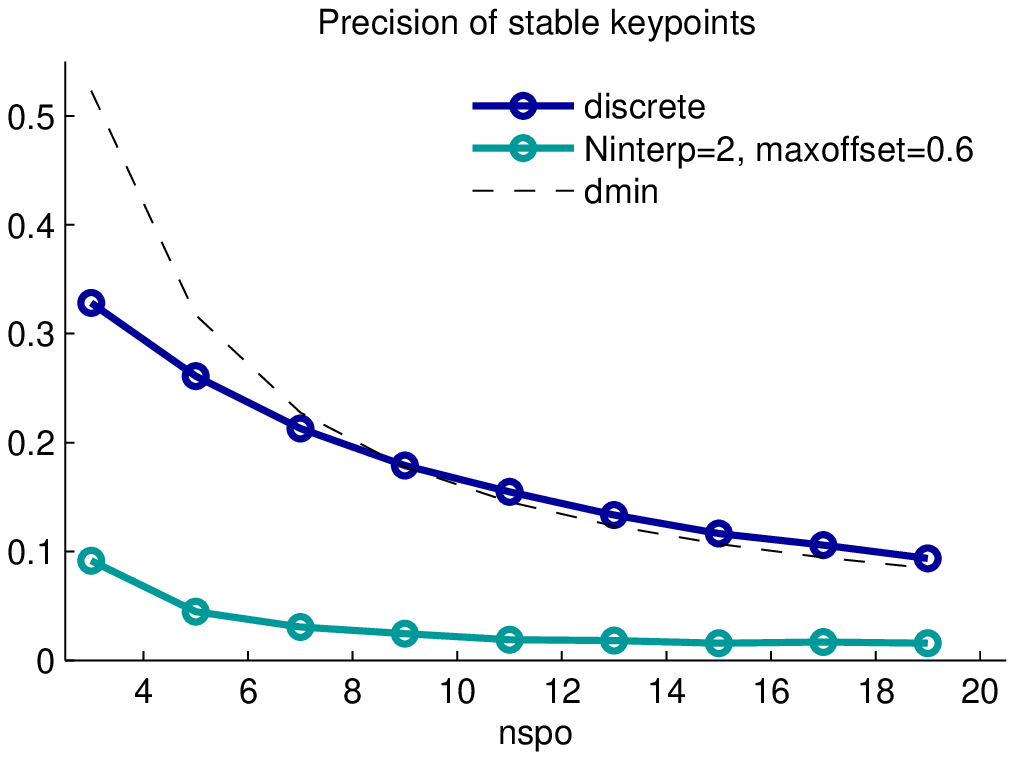}
        {\bf (b)}
    \end{minipage}
\end{center}
\caption{
    %
%    \ivC{Replacing Figure~\ref{fig:influ:interp:zoom:OLD}}
  %  \jmC{
  %      %
  %      Il me semble qu'il faut pr\'eciser ici que l'on ne regarde la r\'ep\'etition que
  %      pour les keypoints qui sont suppos\'es r\'eapparaître, c'est à dire dont
  %      l'\'echelle est sup\'erieure à l'\'echelle de zoom-out?
  %      %
  %  }
  %  \ivC{Done}
  %  \jmC{
  %      %
  %      Par ailleurs il me
  %      semble qu'il faut expliquer pourquoi la courbe discrète pour $\nspo=3$
  %      est largement au dessus de la courbe pour $\nspo=15$: il semble que cela
  %      indique qu'il y  a \'enorm\'ement plus de ``faux'' extrema discrets pour
  %      $\nspo=3$  que pour $\nspo=15$ (presque le triple), mais pourquoi ne
  %      voyait-on pas le même effet, du moins dans la même proportion, dans
  %      l'exprience avec les translations?
  %  }
  %  %
  %  \ivC{Percentages are less confusing - TODO}
    %
    Influence of scale-space sampling and extrema refinement on the invariance
    to zoom-outs.
    %
    %% A set of zoomed-out images was simulated and the keypoints extracted.
    %
    A set of zoomed-out images was simulated, scale-space were computed and the
    keypoints extracted and those which were detected outside the commonly
    covered scale range were discarded.
    {\bf (a)} The percentage of unique detections appearing in at least a
    certain percentage of the simulated images for different scale-space
    sampling and refinements.
    The best performance is obtained by significantly oversampling the
    scale-space, with $\nspo=15$, and by refining the extrema with the local
    interpolation.
    In this case, most of the detected keypoints are present in all the
    simulated images. On the other hand, the original SIFT sampling $\nspo=3$
    leads to low stability even with the extrema refinement step.
    {\bf (b)} Mean precision of stable keypoints location (appearing in at
    least $50\%$ of the zoom-outs) plotted as a function of the sampling rate
    $\nspo$.
    The local refinement of the extrema position significantly increases the
    precision of the extrema detection.
    Also, using a finer grid than the one proposed in SIFT (e.g.,
    $\nspo>3$) allows to better localize the extrema.
   }
\label{fig:influ:interp:zoom}
\end{figure}

We repeated the same experiment but different camera zoom-outs were simulated. 
The results are very similar to the pure camera translation case  (see
Figure~\ref{fig:influ:interp:zoom}).
In general, sampling the scale-space finer than what is proposed in SIFT
(e.g., $\nspo>3$) allows to better localize the DoG extrema.
In addition, the local refinement of the extrema position increases the extrema
precision.
We repeated the experiments with different rotations and reached the same
conclusions.

%\FloatBarrier
\subsection{Influence of $\kappa$}

The DoG scale-space is formed by computing the difference of Gaussians operator
at scales $\kappa\sigma$ and $\sigma$.
To analyze the influence of the DoG parameter $\kappa$, we computed the extrema
of different DoG scale-spaces produced with $\kappa =
2^{\nicefrac{1}{30}},2^{\nicefrac{1}{29}},\ldots,2^{\nicefrac{1}{2}}$.
In order to minimize sampling related instability, the scale-spaces were
sampled at $\nspo=15$ and the respective $\dmin$.

\begin{figure}[h!]
    \begin{center}
    \begin{minipage}[b]{0.48\linewidth}
        \centering
        \includegraphics[width=1\textwidth]{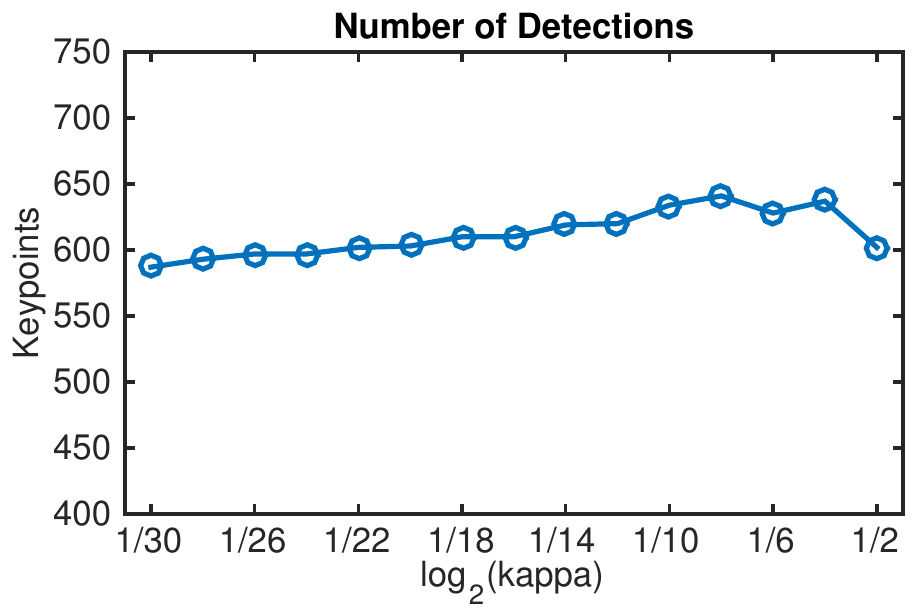}
        {\bf (a)}
    \end{minipage}
        \vspace{.5em}

        \begin{minipage}[b]{0.80\linewidth}
            \centering
            \includegraphics[width=\textwidth]{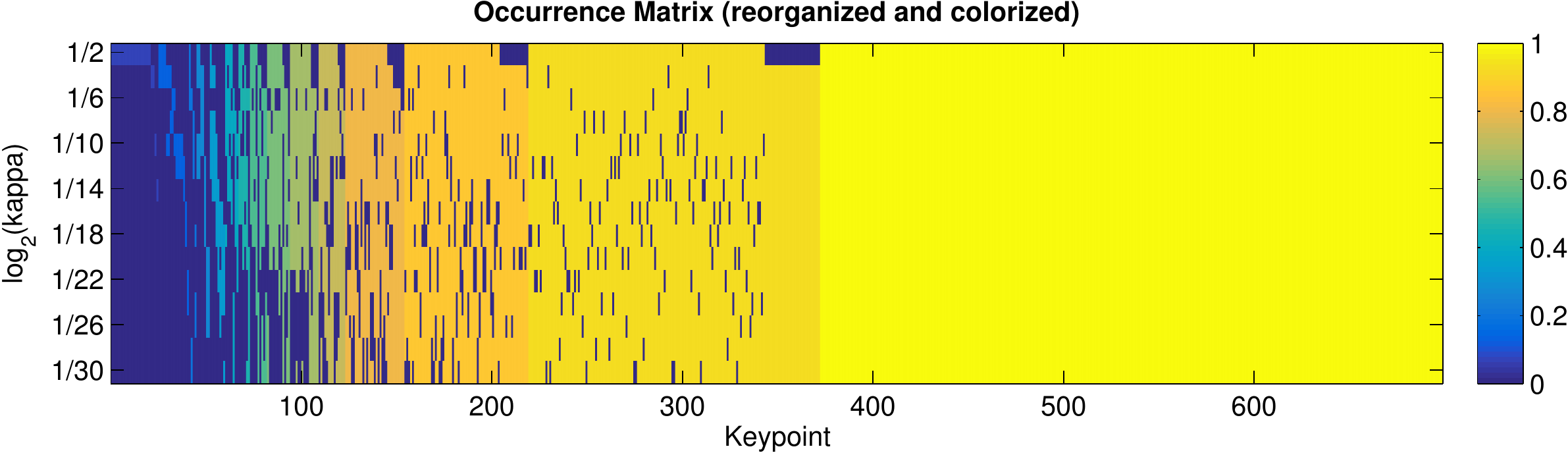} % xticks were wrong ayway
            {\bf (b)}
        \end{minipage}
    \end{center}
    \caption{
        %
  %  \ivC{The xticks in this figure are incorrect! The matrix is interpolated,
  %  the yticks positions are converted correctly, the xticks aren't.}
        %
        Influence of the DoG parameter $\kappa$. 
        The number of detected keypoints is roughly constant for different
        values of $\kappa$  {\bf (a)} . 
        The occurrence matrix for the set of unique normalized keypoints
        detected in the different DoG scale-spaces {\bf (b)}.
        A large majority of the keypoints are detected in most 
        simulated scale-spaces when changing the value of $\kappa$.
        %
       % \ivC{FIXED - corrected xticks after nn interpolation}
    }
    \label{fig:kappa}
\end{figure}

The number of detected extrema is more or less constant for different values of
$\kappa$ (Figure~\ref{fig:kappa}~{\bf(a)})
Depending on the $\kappa$ value, the same structure is detected at a different scale.
%
%In section~\ref{sub:section:modification:implementation}} we showed that a
%Gaussian blob of standard deviation $\sigma$  produces an extrema of the DoG at
%scale $\sigma / \sqrt{\kappa}$.
%
As pointed out in Section~\ref{sub:section:modification:implementation}, a
Gaussian blob of standard deviation $\sigma$  produces an extrema of the DoG at
scale $\sigma / \sqrt{\kappa}$.
Thus, we have normalized the detections scale by $\sigma_\text{normalized} =
\sigma \sqrt{\kappa}$. To compare the keypoints detected with different
$\kappa$ values, we also restricted the analysis to those lying on the common
scale range, that is,
$ \smin \sqrt{2^{1/2}} \le \sigma \le 2 \smin \sqrt{2^{1/30}}.$ 

We proceeded similarly as before by gathering all the detections from the different DoG 
scale-spaces and computed a set of unique detections. Then, we proceeded to create the occurrence matrix.
The occurrence matrix in Figure~\ref{fig:kappa}~{\bf(b)} shows that the
different $\kappa$'s lead for the most part to identical detections. Almost half
the keypoints are detected in every DoG scale-space and a large percentage 
of the keypoints is  detected in most simulated scale-spaces.

%\ivC{bluff}

\section{Impact of deviations from the perfect camera model}\label{sec:deviation}

In order to achieve perfect invariance, SIFT formally requires that the image
is acquired in perfect conditions.
This means that the input image should be noiseless, well-sampled
(according to the Nyquist-Shannon sampling theorem) and with an {\it a priori}
known level of Gaussian blur $c$.
These ideal conditions justify the construction of the image scale-space.
%
%However, in a practical scenario, this is hardly the case.
%
In this section, we evaluate what happens when there are deviations from these
ideal requirements.

%%
%In this section, we investigate how aliasing artifacts impact keypoint
%detection. We also investigate the consequences of computing the DoG
%scale-space while assuming an incorrect camera blur.
%

%\FloatBarrier
\subsection{Image aliasing}
Let us assume that the input image was generated with a camera having a Gaussian point-spread-function of standard deviation $c$.
If $c$ is low (i.e., $c\le0.7$) the acquired image will be subject to aliasing artifacts.
We shall assume first that this camera blur $c$ is known beforehand, so that the SIFT method can be applied consistently.

To evaluate the SIFT performance in this aliasing situation, we simulated random translations of the digital camera.
Then, we computed the extrema of the DoG scale-spaces generated with each translated image and compared the extrema.
All scale-space consisted of one octave computed with $\nspo = 15$, $\smin=1.1$ and the interpolation parameters were set to $N_\text{interp} = 2$ and
$M_\text{offset}=0.6$.

\begin{figure}[htpb]
\begin{center}
    \begin{minipage}[b]{0.48\linewidth}
        \centering
        \includegraphics[width=1\textwidth]{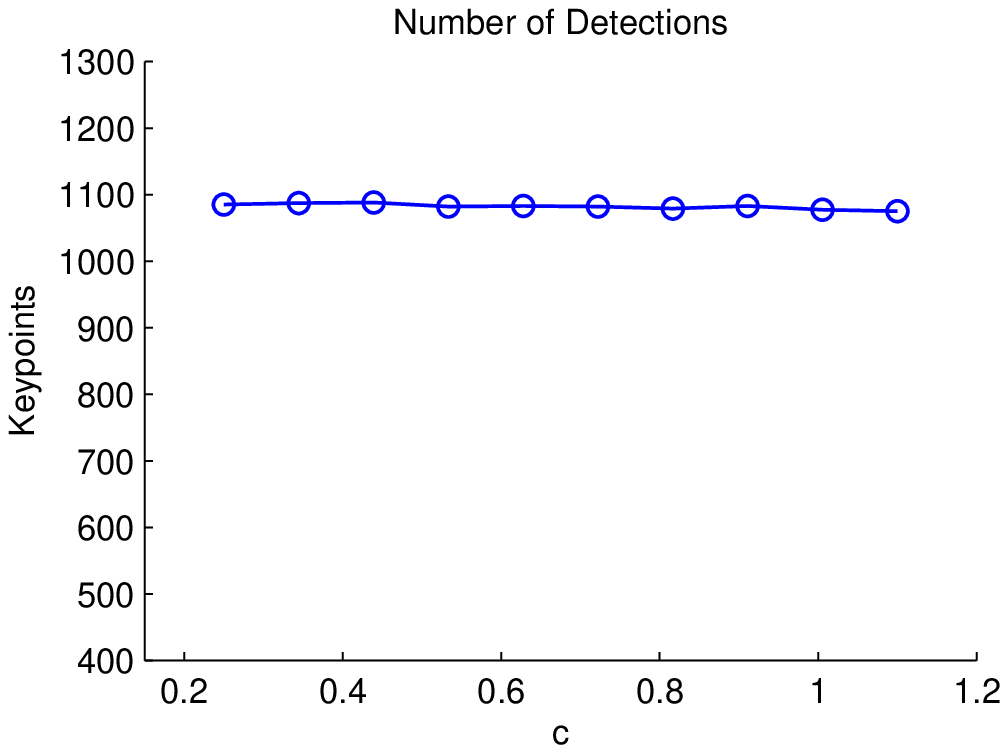}
        {\bf (a)}
    \end{minipage}
    \begin{minipage}[b]{0.48\linewidth}
        \centering
        \includegraphics[width=1\textwidth]{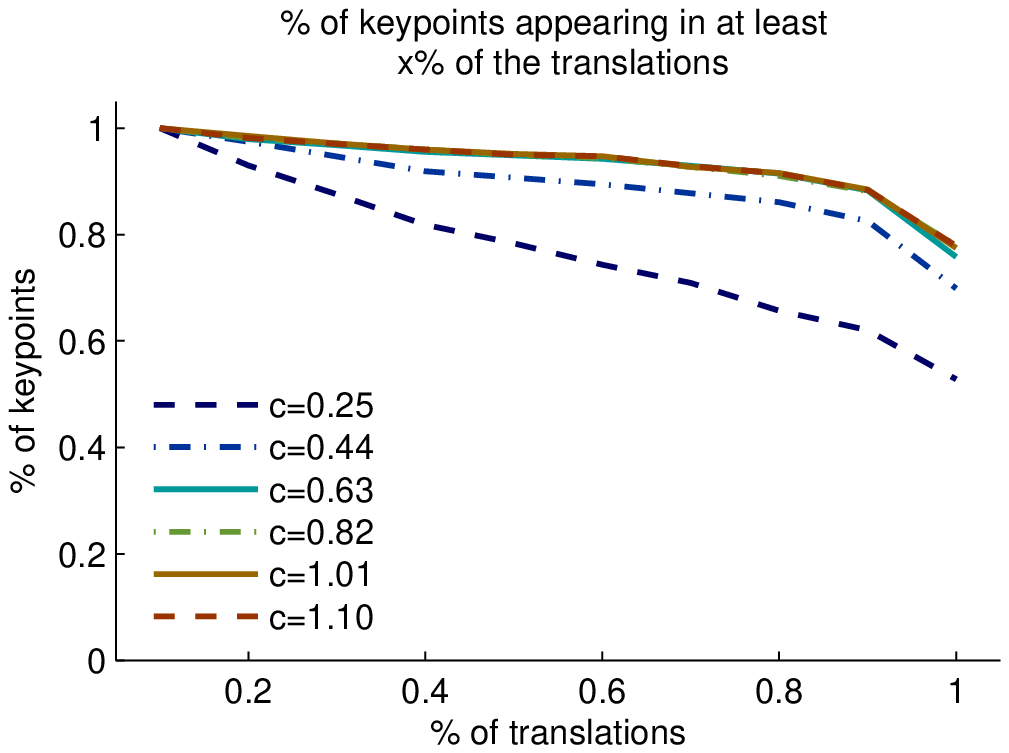}
        {\bf (b)}
    \end{minipage}
\end{center}
\caption{
    Impact of image aliasing.
    For various camera blurs, $0.25 \leq c \leq 1.1$, a set of translated
    images were simulated and the DoG keypoints extracted ($\nspo = 15$,
    $\smin=1.1$).
    Aliasing does not affect the number of detections  {\bf (a)}.
    In {\bf (b)} we show the percentage of unique keypoints appearing in at least a
    certain percentage of the simulated translations.
    Detections are less stable for severely aliased images ($c=0.25$), while
    for $c>0.6$, the impact of aliasing is negligible.
}
\label{fig:aliasing}
\end{figure}

%The considered camera blur range from $c=0.25\text{--}1.1$. All input
%images are simulate from \texttt{plants} ($S=4$).
%%
%All scale-space consist of one octave computed with $\nspo = 15$, $\smin=1.1$
%and the interpolation parameters are set to $\texttt{Ninterp} = 2$ and
%$M_\text{offset}=0.6$.
%%
%
Figure~\ref{fig:aliasing}~{\bf (a)} shows the average number of keypoints
detected as a function of the camera blur $c$.
The number of detections is independent of the camera blur. Indeed, a sharper
shot does not increase the number of keypoints.

In Figure~\ref{fig:aliasing}~{\bf (b)} we show the percentage of unique keypoints
that appear in at least a certain percentage of the translated images.
Keypoints detected from well sampled images (e.g., $c>0.6$) are stable to
translation (the curves are almost flat) while those from severely undersampled
images ($c\approx0.3$) are very sensitive to the position of the sampling grid,
as expected. 
%

%\begin{figure}[htpb]
%\begin{center}
%    %
%    %
%    %
%    \begin{minipage}[b]{0.48\linewidth}
%        \centering
%        \includegraphics[width=1\textwidth]{./figs/fig_aliasing_nkeys.eps}
%        {\bf (a)}
%    \end{minipage}
%    %
%    %
%    \begin{minipage}[b]{0.48\linewidth}
%        \centering
%        \includegraphics[width=1\textwidth]{./figs/fig_interp_cumhist_aliasing.eps}
%        {\bf (b)}
%    \end{minipage}
%    %
%    %
%\end{center}
%%
%\caption{
%    %
%    \ivC{Replace with percentage}
%    %
%    Impact of image aliasing.
%    %
%    For various camera blurs, $0.25 \leq c \leq 1.1$, a set of translated
%    images were simulated and the DoG keypoints extracted ($\nspo = 15$,
%    $\smin=1.1$).
%    %
%    Aliasing does not affect the number of detections  {\bf (a)}.
%    %
%    In {\bf (b)} we show the number of unique keypoints appearing in at least a
%    certain percentage of the simulated translations.
%    %
%    Detections are less stable for severely aliased images ($c=0.25$), while
%    for $c>0.6$, the impact of aliasing is negligible.
%    %
%    %
%    %
%}
%\label{fig:aliasing:OLD}
%\end{figure}

%\FloatBarrier
\subsection{Unknown input image blur level}

A more realistic scenario is the case where the level of blur of the input
image $c$ is unknown.
SIFT requires this value to create the scale-space starting at a known level of
image blur $\smin$.
A wrong assumption of the input camera blur affects the range of simulated
scales simulated in the Gaussian scale-space.
%

%Figure~\ref{fig:error:blur}~{\bf (a)} shows the number of keypoints detected
%assuming that the input image level of blur is $c=0.5$ when it was actually $c_\text{real}$, with $0.5 \leq c_\text{real}
%\leq 1.1$.
%%
%Note that $c_\text{real}= 1.1$ leads to the largest number of detections
%despite the fact that more details have been washed away as a result of an
%underestimated camera blur.

To demonstrate to what extent the wrong knowledge of the input camera blur
produces unrelated keypoints, we compared the keypoints extracted assuming an
image blur of $c=0.7$ from a set of images having actual random blur
$c_\text{real}$ uniformly picked from $[c -\Delta c, c+ \Delta c]$.

\begin{figure}[htpb]
\begin{center}
    \begin{minipage}[b]{0.48\linewidth}
        \centering
        \includegraphics[width=1\textwidth]{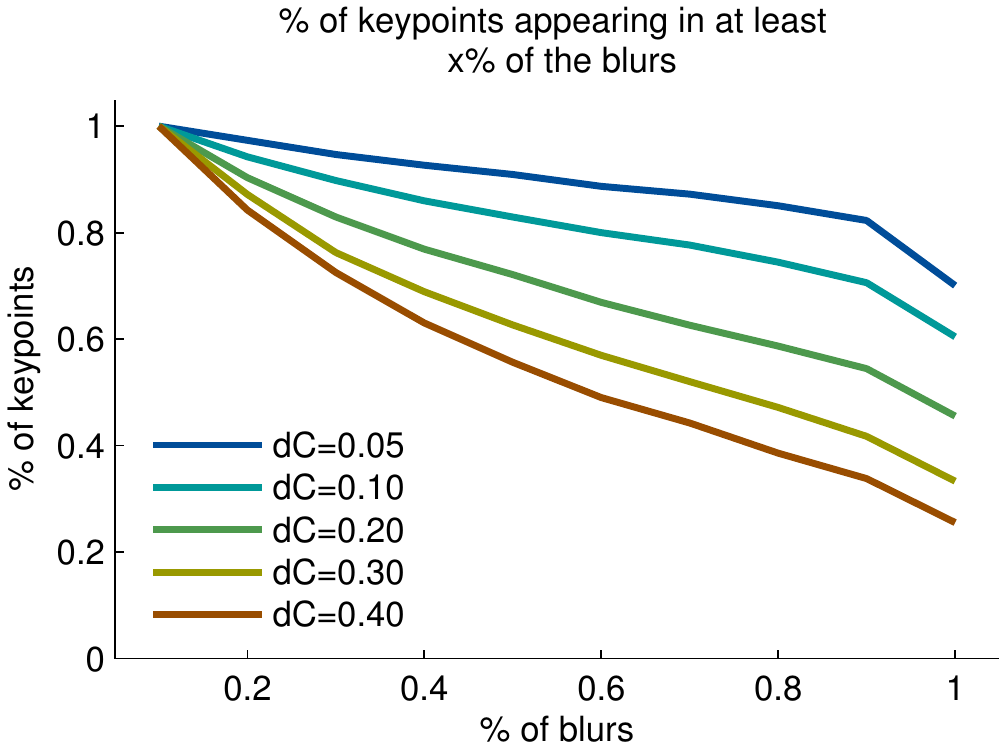}
        {\bf (a)}
    \end{minipage}
    \begin{minipage}[b]{0.48\linewidth}
        \centering
        \includegraphics[width=1\textwidth]{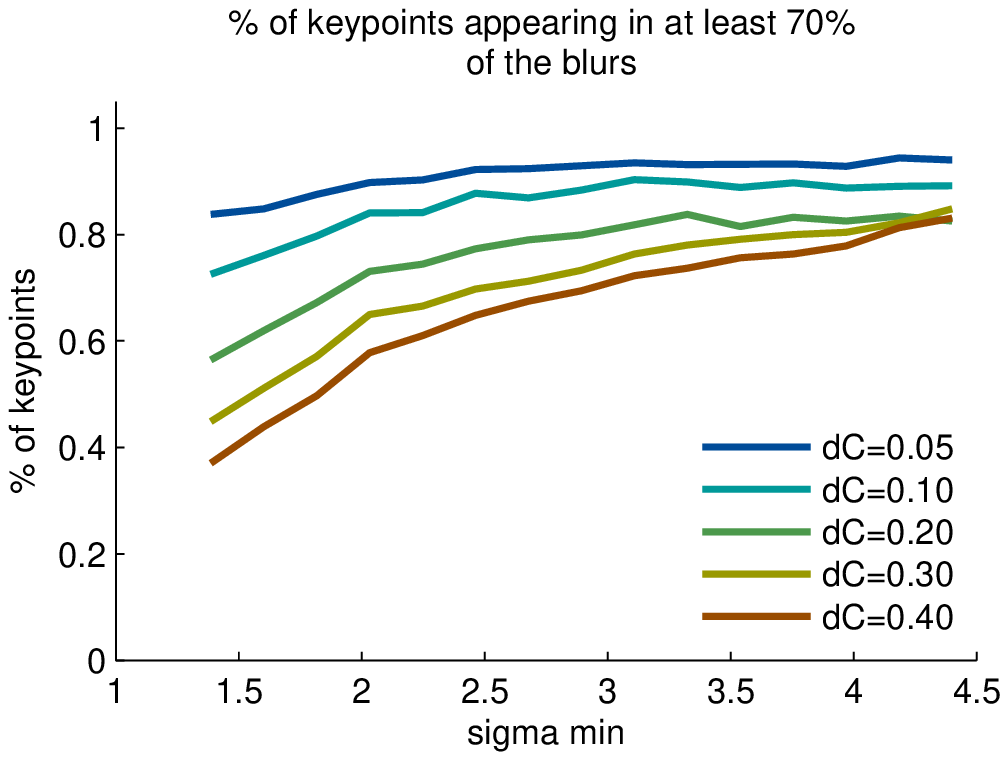}
        {\bf (b)}
    \end{minipage}
\end{center}
\caption{
    %
%    \ivC{Proposed replacement for Figure~\ref{fig:error:blur:OLD}}
    %
    The impact of a wrong assumption on the camera blur.
    Comparison of the keypoints extracted assuming $c=0.7$ when the real camera
    blur was picked randomly in $[c -\Delta c,  c + \Delta c ]$.
    {\bf (a)} The percentage of unique keypoints that appear in at least a
    certain percentage of the simulated images is plotted for different levels
    of uncertainty on camera blur ($\Delta c = 0.05-0.4$).
    {\bf (b)} Influence of scale on stability to wrong blur assumption.
    For keypoints detected at scales ranging from $\sigma_\text{min}$ and
    $2\sigma_\text{min}$, the proportion of unique keypoints that appear in at
    least 70\% of the simulated images is shown as a function of scale
    $\sigma_\text{min}$.
    %
    %The impact of noise slowly as we consider detections at larger scales.
    %
    %Low $\sigma_\text{min}$ values (\approx1.5) show  the stability to error
    %in blur assumptions of detections in the
    %first octave, while values for $\sigma_\text{min} = 2.2$ show to the
    %stability in the second octave.
    %
    The impact of a wrong blur level assumption decreases as we consider detections at
    larger scale (i.e., large $\sigma_\text{min}$).
}
\label{fig:error:blur}
\end{figure}

Figure~\ref{fig:error:blur} shows the number of unique keypoints that appear in
at least a certain percentage of the simulated images. This was evaluated for
different ranges of uncertainty (i.e., $\Delta c= 0.05-0.4$).
The larger the range of uncertainty $\Delta c$, the more unrelated the extrema
are (the curve decreases very fast, indicating the presence of many unique
keypoints appearing in only a few of the simulated images). 
Figure~\ref{fig:error:blur}{\bf (b)} explores the influence of detection
scale on stability to wrong blur assumption.
The percentage of unique keypoints appearing in at least 70\% of the simulated
images is shown as a function of scale.
The influence of a wrong assumption decreases with detection scales.

%
%\begin{figure}[htpb]
%\begin{center}
%    %
%    \begin{minipage}[b]{0.48\linewidth}
%        \centering
%        \includegraphics[width=1\textwidth]{./figs/fig_interp_cumhist_error_blur_variance.eps}
%        %{\bf (b)}
%    \end{minipage}
%    %
%\end{center}
%%
%\caption{
%    \ivC{Replaced with Figure~\ref{fig:error:blur}}
%    %
%    \ivC{Consider replacing it with the percentage of unique keypoints}
%    %
%    The impact of a wrong assumption on the camera blur.
%    %
%    Comparison of the keypoints extracted assuming $c=0.7$ when the real camera
%    blur was picked randomly in $[c -\Delta c,  c + \Delta c ]$.
%    %
%    The number of keypoints that appear in at least a certain percentage of
%    the simulated images is plotted for different levels of uncertainty on
%    camera blur ($\Delta c = 0.05-0.4$).
%    %
%    %
%}
%\label{fig:error:blur:OLD}
%\end{figure}
%

%

%\FloatBarrier
\subsection{Image noise}

The digital image acquisition is always affected by noise that undermines the
performance of SIFT.
To evaluate the impact of image noise we simulated different image acquisition,
by adding random white Gaussian noise to the input image.
Then, we proceeded to compute the keypoints that are detected in a certain
percentage of the simulated images.
%
%Figure~\ref{fig:noise} is self-explanatory and demonstrates the strong impact
%of noise level on keypoint stability.
%
%Figure~\ref{fig:noise}, for increasing levels of noise, we simulate a set of
%noisy images.

\begin{figure}[htpb]
\begin{center}
    \begin{minipage}[b]{0.48\linewidth}
        \centering
        \includegraphics[width=1\textwidth]{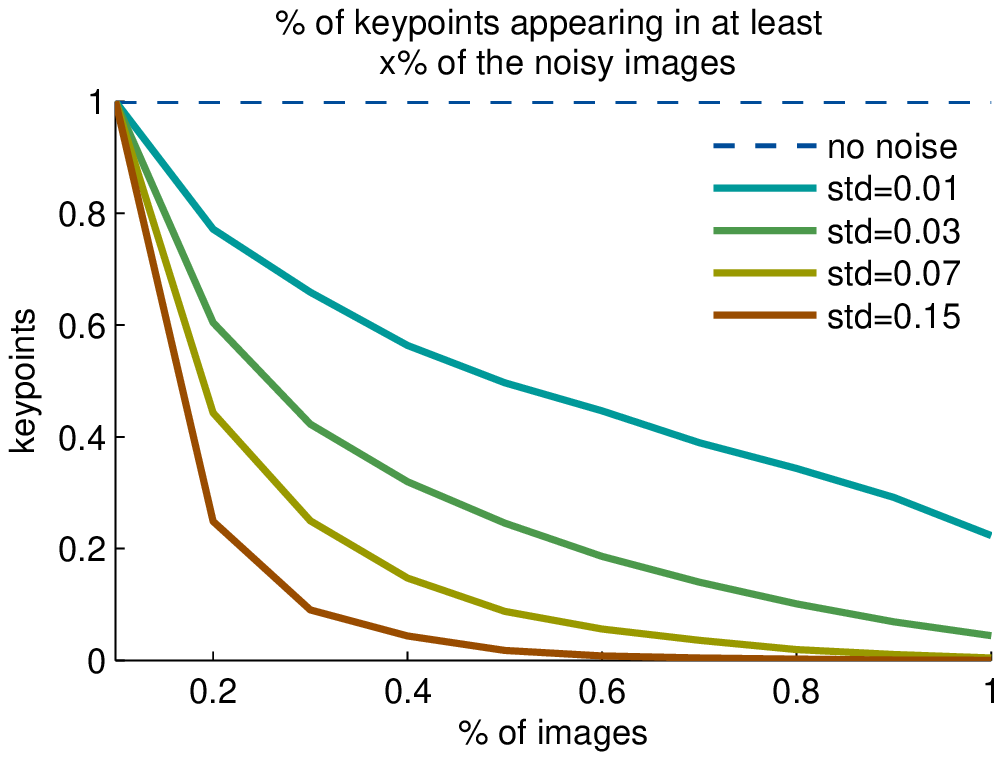}
        {\bf (a)}
    \end{minipage}
    \vspace{2em}
    \begin{minipage}[b]{0.4\linewidth}
    %\begin{minipage}[b]{0.38\linewidth}
        \centering
        \includegraphics[width=0.44\textwidth]{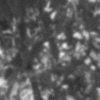}
        \includegraphics[width=0.44\textwidth]{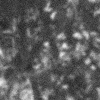}
        \includegraphics[width=0.44\textwidth]{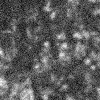}
        \includegraphics[width=0.44\textwidth]{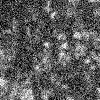}

        {\bf (b)}
    \end{minipage}
    \vspace{1em}
    \begin{minipage}[b]{0.48\linewidth}
        \centering
        \includegraphics[width=1\textwidth]{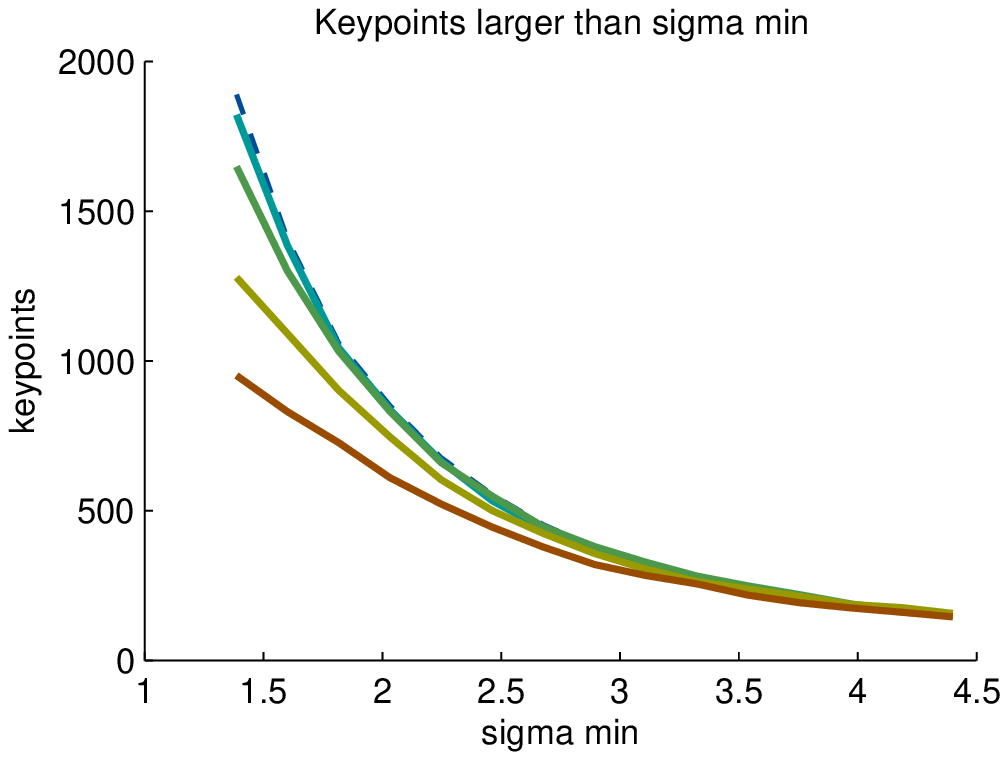}
        {\bf (c)}
    \end{minipage}
    \begin{minipage}[b]{0.48\linewidth}
        \centering
        \includegraphics[width=1\textwidth]{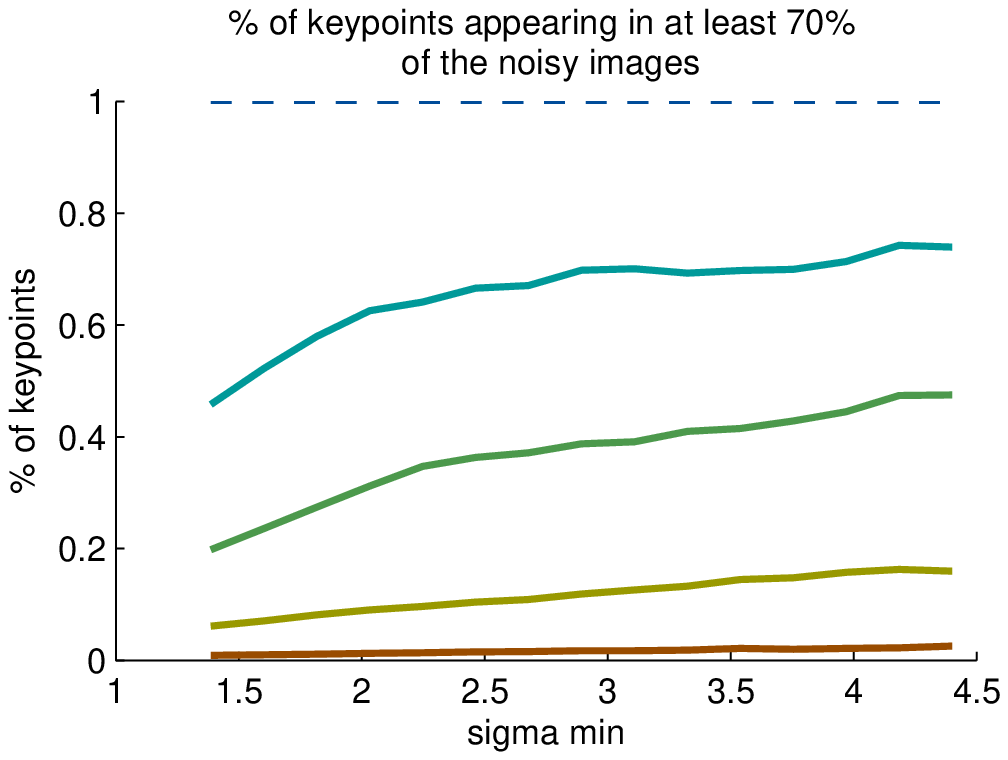}
        {\bf (d)}
    \end{minipage}
\end{center}
\caption{
    Impact of image noise. %
    {\bf (a)} The proportion of unique keypoints that appear in at least a
    certain proportion of the simulated images is plotted for different levels
    of image noise.
    Noise has a significant impact on the DoG extrema detection.
    {\bf (b)} Crops of the input images simulated with $c=0.8$ and added
    Gaussian white noise of standard deviation $\sigma_\text{noise} = 0.01,
    0.03, 0.07 \text{ and } 0.15$.
    {\bf (c)} Number of keypoints detected at a scale larger than $\smin$ as a
    function of $\smin$.
    The number of detections decreases as the level of noise increases.
    {\bf (d)} Influence of scale on stability to noise.
    For keypoints detected at scales ranging from $\sigma_\text{min}$ to
    $2\sigma_\text{min}$, the proportion of unique
    keypoints that appear in at least 70\% of the simulated images is shown a
    function of scale $\sigma_\text{min}$.
    %
 %   \mdC{Where is the change from first octave to second octave,
    %   $\sigma_\text{min}?$ I assume that you mean in the default
    %   $\sigma_\text{min}=0.8?$. I don't see how can you conclude what is
    %   next.}
 %   %
 %   Values on the left side of the curve show therefore the stability to noise
 %   of detections in the first octave, while values for $\sigma_\text{min} =
 %   2.2$ correspond to the stability in the second octave.
    %
    Unsurprisingly we observe that, for a given level of noise, the stability
    in the second octave is comparable to the stability achieved in the first
    octave with half the level of noise.
   }
\label{fig:noise}
\end{figure}

Figure~\ref{fig:noise} shows results when considering set of input images with increasing level of noise.

Specifically, Figure~\ref{fig:noise}  {\bf (a)} shows the percentage of unique keypoints that appear in at
least a certain percentage of the simulated images.

It demonstrates the strong impact of noise level on keypoint stability.
%
%Noise has a strong impact on the stability of the detected keypoints.
%
Such impact however is mitigated for detections at larger scales. 
In a Gaussian scale-space, the level of noise decreases as the scale increases.
In fact, the noise standard deviation observed in a given octave is half
the one observed in the previous octave.
This is confirmed in Figure~\ref{fig:noise} {\bf (d)}, which shows, for
keypoints detected in a range of scale $[\sigma_\text{min},
2\sigma_\text{min}]$, the proportion of unique keypoints that appear in at
least 70\% of the simulated noisy image as a function of scale
$\sigma_\text{min}$.

%\begin{figure}[htpb]
%\begin{center}
%    %
%    %
%    \begin{minipage}[b]{0.48\linewidth}
%        \centering
%        \includegraphics[width=1\textwidth]{./figs/fig_interp_cumhist_noise_variance-eps-converted-to.pdf}
%        {\bf (a)}
%    \end{minipage}
%    %
%    \begin{minipage}[b]{0.4\linewidth}
%    %\begin{minipage}[b]{0.38\linewidth}
%        \centering
%        \includegraphics[width=0.44\textwidth]{./figs/crop_noise_std_0_01.png}
%        \includegraphics[width=0.44\textwidth]{./figs/crop_noise_std_0_03.png}
%        \includegraphics[width=0.44\textwidth]{./figs/crop_noise_std_0_07.png}
%        \includegraphics[width=0.44\textwidth]{./figs/crop_noise_std_0_15.png}
%        {\bf (b)}
%    \end{minipage}
%    %
%\end{center}
%\caption{
%    %
%    \ivC{See replacement proposal - with percentage and the influence of scale.}
%    %
%    Impact of image noise.
%    %
%    {\bf (a)} The number of unique keypoints that appear in at  least a certain proportion of the
%    simulated images is plotted for different levels of image noise. 
%    Noise has a significant impact on the DoG extrema detection.
%    %
%    {\bf (b)} Crops of the input images simulated with $c=0.8$ and added
%    Gaussian white noise of standard deviation $\sigma_\text{noise} = 0.01,
%    0.03, 0.07 \text{ and } 0.15$.
%%
%}
%\label{fig:noise}
%\end{figure}

\section{Concluding remarks}\label{sec:conclusion}

We presented a systematic analysis of the main steps involved in the detection
of keypoints in the SIFT algorithm.
One of the main conclusions is that the original parameter choice in SIFT is
not sufficient to ensure a theoretical and practical scale (and even
translation) invariance, which was the main claim of the SIFT method.
In addition, we showed that the SIFT invariance claim is strongly affected if the
assumption on the level of blur in the input image is wrong.

Specifically, we showed that increasing the scale-space sampling from $\nspo=3$
to $\nspo=15$ (and respectively the space sampling rate $\dmin$) improves the
stability of the detected keypoints.
This implies that if a series of image transformations (e.g., translations,
zoom-outs) are applied to an image, the keypoints detected in one of them will
be detected with high probability in all the others.
This stability property is fundamental for fulfilling the scale invariance
claim.
The extrema refinement was shown to improve both the precision and the
stability of the detected keypoints.
We showed that the largest number of stable keypoints is achieved with
parameters $M_\text{offset} = 0.6$ and $N_\text{interp}=2$ (while SIFT
recommends $N_\text{interp}=5$).
We also demonstrated that the DoG threshold fails to filter out unstable
keypoints, and that the different definitions of the DoG scale-space (parameter
$\kappa$) lead for the most part to identical detections up to a normalization
of the scale.
Finally, we showed how the presence of aliasing and noise in the
acquired image deteriorate detections stability.

\section*{Acknowledgements}
Work  partially supported by  Centre National d'Etudes Spatiales
(CNES, MISS Project),  European Research Council (Advanced Grant Twelve
Labours),  Office of Naval Research (Grant N00014-97-1-0839), Direction
G\'{e}n\'{e}rale de l'Armement (DGA), Fondation Math\'{e}matique Jacques
Hadamard and Agence Nationale de la Recherche (Stereo project).

\bibliographystyle{IEEEtran}
\bibliography{jmiv_sampling_sift_v5-4_arxiv}

\end{document}